%% file: main.tex
\documentclass{article}

\usepackage{iclr2027_conference,times}

\input{latex/shared/preamble.tex}
\input{latex/shared/metadata.tex}
\title{\PaperTitle}
\input{latex/format/iclr_authors.tex}

\iclrfinalcopy

\newcommand{\ProjectPageFootnote}{\footnote{Project page: \url{https://reefmenaged.github.io/Agentic_Automata_Learning}.}}

\begin{document}
\maketitle
% Override the camera-ready wording inserted by the ICLR style.
\lhead{Preprint. Under review at ICLR 2027}

\begin{abstract}
\input{latex/shared/abstract.tex}
\end{abstract}

\input{latex/shared/main_text.tex}
% Public version includes acknowledgments and the ICLR AI-use statement.
\input{latex/sections/Acknowledgments.tex}
\input{latex/iclr_ai_use_statement.tex}

% Bibliography is pre-generated and inlined for arXiv compatibility.

\appendix
\clearpage
\input{latex/shared/appendix.tex}

\end{document}

%% file: latex/shared/preamble.tex
\usepackage{amssymb}
\usepackage{placeins}
\usepackage{tikz}
\usetikzlibrary{positioning, backgrounds, calc}
\usepackage{latexsym}
\usepackage{hyperref}
\usepackage{amsmath}
\usepackage{subcaption}
\usepackage{xcolor}
\definecolor{taborange}{HTML}{FF7F0E}
\definecolor{tabblue}{HTML}{1F77B4}
\definecolor{tabgreen}{HTML}{2CA02C}
\definecolor{tabred}{HTML}{D62728}
\definecolor{tabpink}{HTML}{E377C2}
\definecolor{gold}{HTML}{FFD700}
\usepackage{pgfplots}
\pgfplotsset{compat=1.18}
\usepackage{booktabs}
\usepackage[table]{xcolor}
\usepackage{array}
\usetikzlibrary{patterns}
\usepackage{fvextra}
\usepackage{mdframed}

\definecolor{rfomodelwin}{HTML}{DCE8C8}
\definecolor{rfobothfailed}{HTML}{C7C7C7}
\definecolor{rfotie}{HTML}{9FCFD0}
\definecolor{rfopassivewin}{HTML}{908BC4}

\pgfplotsset{
    rfoalgaxis/.style={
        xbar stacked,
        width=\linewidth,
        height=3.9cm,
        xmin=0,
        xmax=100,
        xtick={0,20,40,60,80,100},
        xlabel={Rate (\%)},
        symbolic y coords={
            Gemini 3.1 Pro,
            DeepSeek V4 Pro,
            Gemini 3.1 Flash Lite,
            Gemini 3 Flash,
            GPT-5.4,
            Llama-3.3-70B
        },
        ytick=data,
        yticklabels={,,,,,},
        y dir=reverse,
        bar width=7pt,
        enlarge y limits=0.10,
        grid=major,
        grid style={gray!18},
        tick label style={font=\tiny},
        label style={font=\tiny},
        title style={font=\scriptsize\bfseries},
        axis line style={black!65},
        tick style={black!65},
        clip=false,
    }
}

\usepackage[T1]{fontenc}
\usepackage[utf8]{inputenc}
\usepackage{microtype}
\usepackage{inconsolata}
\usepackage{graphicx}
\usepackage{cuted}
\usepackage{caption}
\usepackage{bbm}

%% file: latex/shared/metadata.tex
\newcommand{\PaperTitle}{Can Agents Infer Environment from Interaction? 
Evidence from Agentic Automata Learning}

%% file: latex/format/iclr_authors.tex
\author{
\hspace*{-0.2cm}
\textbf{Reef Menaged}$^{1}$ \hspace{0.8em}
\textbf{Gili Lior}$^{1}$ \hspace{0.8em}
\textbf{Shauli Ravfogel}$^{2}$ \hspace{0.8em}
\textbf{Roee Aharoni}$^{3}$ \hspace{0.8em}
\textbf{Gabriel Stanovsky}$^{1}$
\\[0.5em]
\texttt{reef.menaged@mail.huji.ac.il}
\\[0.8em]
\begin{tabular}{l}
$^{1}$\textit{The Hebrew University of Jerusalem}\\
$^{2}$\textit{New York University}\\
$^{3}$\textit{Google Research}
\end{tabular}
}

%% file: latex/shared/abstract.tex
We propose \emph{agentic automata learning} to evaluate the extent to which tool-calling LLM agents can uncover hidden environments through interaction, a capability increasingly required in agentic tasks (e.g., reproducing an executable without access to its source code by interacting with it). In our setup, an agent should uncover a hidden deterministic finite automaton (DFA) by interacting with an oracle through (1) membership queries ("Does this string belong to the target language?") and (2) equivalence queries ("Is this the target DFA?"). Agentic automata learning yields a scalable testbed with controlled task complexity, measurable interaction efficiency, and strong algorithms to compare against from the classic automata-learning literature. Evaluating state-of-the-art LLMs with a multi-turn agent scaffold, we find that while they are able to recover simple DFAs, their performance drops sharply as DFA size increases. Results improve with a more elaborate ReAct-style state-tracking scaffold, yet strong models still struggle with complex instances. Trajectory analyses reveal recurring failures in query planning, evidence integration, and hypothesis construction. These failures occur even though the models mention classic automata learning algorithms in their reasoning, and can implement and execute them when given access to a coding environment. Our results suggest that for state-of-the-art LLMs, identifying a solution to the problem is insufficient, and reliably executing the resulting plan still poses a distinct challenge.\ProjectPageFootnote

%% file: latex/shared/main_text.tex
% Main paper text shared by both formats.

\input{latex/sections/01_introduction.tex}
\input{latex/sections/02_background.tex}

\input{latex/sections/03_task_formulation.tex}
\input{latex/sections/04_evaluation_framework.tex}
\input{latex/sections/05_results.tex}

\input{latex/sections/07_related_work.tex}

\input{latex/sections/08_conclusion.tex}

\input{latex/sections/limitations.tex}

%% file: latex/sections/01_introduction.tex
\section{Introduction}

LLMs are increasingly used as interactive agents across environments of varying complexity, from settings like games with fixed rules to dynamic environments such as navigation in open worlds~\citep{park2026orak, DBLP:journals/corr/abs-2505-11942}. Despite their strong performance, it remains unclear whether LLMs can infer a correct representation of the environment they interact with, or whether they rely on local patterns and shallow heuristics~\citep{vafa2024worldmodel}. Addressing this question requires controlled tasks where the complexity of the hidden environment can be controlled systematically, and the agent's discovery process can be observed throughout the interaction.

\input{latex/figures/framework_diagram}

The classic theoretical problem of \emph{active automata learning} offers a well-studied formal framework for learning hidden structure via interaction~\citep{isberner2015foundations}. In this formulation, 
the hidden rules governing the environment are represented with a deterministic finite automaton (DFA), which a learner attempts to identify  through a series of interactions, represented by queries to an oracle. \emph{Membership Queries} ask whether a word belongs to the target language, while \emph{Equivalence Queries} submit a hypothesis DFA, which the oracle either accepts or refutes with a counterexample, i.e., a word on which the hypothesis DFA and the hidden DFA disagree. 

In this work, we recast active automata learning as \emph{agentic automata learning}, as exemplified in Figure~\ref{fig:framework_diagram}. To do so, we build an oracle with a target DFA and equip an LLM agent with tools to perform membership and equivalence queries against the oracle. This formulation turns automata learning into a controlled task for testing whether LLM agents can infer latent structure through interaction. It provides formal measures for task complexity (e.g., the number of states in the DFA) as well as for discovery efficiency (e.g.,  the number of queries to the oracle), and enables direct comparison against strong deterministic solvers for the task, including the $L^*$ and TTT algorithms~\citep{angluin1987learning,angluin1988queries,isberner2014ttt}.

% Agentic automata learning studies an environment simpler than real-world environments, as a necessary step before considering more complex environments, while also using an agent without an advanced harness, in order to identify limitations in its parametric capabilities that do not rely on external tools. The environment is represented as a small DFA, while the oracle provides informative feedback, making the task solvable within the query budget and enabling formal analysis of the sources of failure. In addition, a current trend in agent evaluation benchmarks is to provide agents with increasingly powerful harnesses that allow them to delegate reasoning to external tools. Here, we choose to limit the harness in order to isolate the agent's parametric capabilities and identify their limitations.

Our results show that agentic automata learning is challenging for state-of-the-art LLMs. LLMs can reliably discover DFAs with few states, but their performance quickly deteriorates as the DFA grows: for DFAs with 8 or 9 states, the highest success rate is only 30.0\%. Adding a ReAct-style state-tracking scaffold~\citep{yao2022react, rozanov-rei-2025-stateact}  generally improves performance, raising the highest success rate in this setting to 57.5\%, but still leaves substantial room for improvement. 
% The gap also extends to interaction efficiency: on successful 8--9-state runs, Gemini 3.1 Pro uses 30.1\% more tool calls than TTT under Stateless and 23.5\% more under Stateful.

% Beyond task performance, our analyses reveal substantial gaps in models’ ability to turn knowledge into coherent execution. 
% \gabis{We we find that all tested LLMs know the classic algorithms, which they sometimes mention in their reasoning tokens, and which correctly reproduce in code when given access to a python interpreter and a terminal.}
% \gabisrep{Even when models can identify the appropriate strategy,}{Despite this knowledge,} they often struggle to plan informative queries, consistently integrate accumulated evidence, and use prior observations to construct and refine hypotheses. 
% These findings suggest that for frontier LLMs, knowing how to solve a task is not sufficient, and that carrying out a strategy remains a distinct challenge.

Beyond task performance, our analyses reveal substantial gaps in models’ ability to turn knowledge into coherent execution. We find that all tested LLMs have knowledge of the classic automata learning algorithms encoded in their parameters. Evidently, they sometimes refer to these algorithms in their reasoning traces and can reproduce them correctly when given access to a Python interpreter and a terminal. Despite knowing a path to the solution, they often struggle with planning informative queries, integrating accumulated evidence consistently, and using prior observations to construct and refine hypotheses.
These findings suggest that, even for frontier LLMs, knowing how to solve a task is not sufficient and that carrying out a strategy remains a distinct challenge.

To summarize, our main contributions are as follows:
(1) We introduce \emph{agentic automata learning}, a controlled framework for evaluating whether tool-calling LLM agents can infer latent structure through interaction; (2) show that advanced models can perform non-trivial interactive discovery, but degrade sharply as automata complexity increases; (3) analyze interaction trajectories and identify recurring limitations in planning, reasoning, and consistent use of accumulated information; and (4) show that identifying a solution is insufficient, and that reliably executing the resulting plan poses a distinct challenge.

%% file: latex/figures/framework_diagram.tex
\begin{figure}[t!]
\centering
\includegraphics[width=0.9\columnwidth]{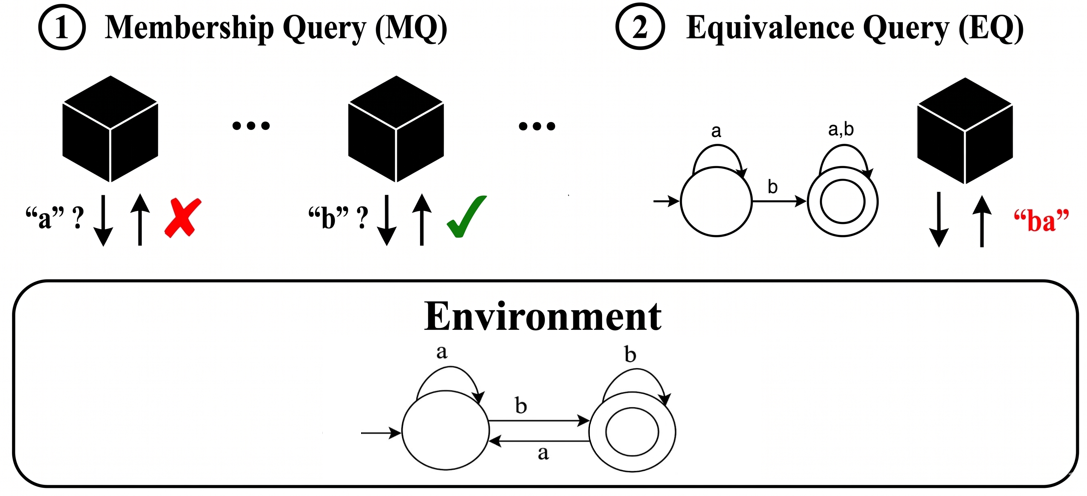}
\caption{\textbf{Agentic automata learning.} An agent is tasked to identify a hidden DFA through successive interactions with the environment. \emph{Membership Queries} ask whether a word is accepted by the hidden DFA (left), while \emph{Equivalence Queries} submit a hypothesized DFA and ask whether it matches the hidden target. The environment either confirms equivalence or returns a counterexample on which the two DFAs differ (right).
\label{fig:framework_diagram}}
\end{figure}

%% file: latex/sections/02_background.tex
\section{Background: Automata Learning}

Automata learning tests the extent to which learning algorithms are capable of discovering the structure of a hidden DFA from observations of which words it accepts or rejects. Automata learning is commonly divided into two paradigms: \emph{active automata learning}, where the learner can interact with an oracle through queries, and \emph{passive automata learning}, where the learner infers a DFA from a fixed set of labeled examples~\citep{delahiguera2010grammatical}.

\subsection{Active Automata Learning}
In this setting, a learner interacts with an oracle through adaptively chosen queries in order to identify a hidden DFA~\citep{isberner2015foundations}. The oracle has access to the hidden DFA and answers queries about the language it recognizes. Traditionally, there are two types of queries~\citep{angluin1988queries, angluin1990negative}. A Membership Query allows the learner to ask whether a given word is accepted by the hidden DFA, while an Equivalence Query allows the learner to propose a hypothesis DFA and test whether it is equivalent to the hidden DFA, which the oracle can either accept or refute
with a counterexample -- a word on which the hypothesis DFA and the hidden DFA disagree.

Prior work has established formal complexity and evaluation measures, as well as efficient deterministic active automata learning algorithms~\citep{angluin1987learning, rivest1993inference}. Task complexity is typically characterized by the number of states in the minimal DFA accepting the language of the hidden DFA, while learning efficiency is measured by the number of queries required to identify the hidden DFA~\citep{isberner2015foundations}.

As can be seen on the right-hand side of Table~\ref{tab:example-trajectories}, classic learners such as L* and TTT solve this task through an iterative hypothesis-refinement process: they consider a hypothesis which matches all given observations using EQ, and when a counterexample is returned, use a polynomial number of MQs to construct a new hypothesis that again accounts for all given information while adding at least one previously unidentified state in the hidden DFA. These algorithms were proven to converge on the hidden DFA in polynomial time~\citep{angluin1987learning, isberner2014ttt}. This exemplifies several desired agentic real-world traits, including maintaining a memory of observed behavior, reasoning over it, choosing actions based on previous interactions, and adapting to new information.

\input{latex/figures/interaction_table}

\subsection{Passive Automata Learning}
\label{sec:passive_learning}
In passive learning, the learner does not query an oracle. Instead, it receives a fixed labeled sample of accepted and rejected words and must infer a DFA consistent with that sample~\citep{murphy1996passively}. Classic passive algorithms such as RPNI, EDSM, and Blue-Fringe learn automata in this way~\citep{oncina1992rpni, lang1998edsm}. 

While we focus on active automata learning, we use passive learning algorithms to assess the information gathered by LLM agents during active interaction: given the labeled words collected by an agent via membership queries, we test whether a passive learner can already infer the hidden DFA.

%% file: latex/figures/interaction_table.tex
\begin{table*}[b!]
\centering
\renewcommand{\arraystretch}{0.60}
\begin{tabular}{c cc cc cc cc}
\toprule
& \multicolumn{4}{c}{\textbf{LLM-based learners}}
& \multicolumn{4}{c}{\textbf{Classic algorithms}} \\
\cmidrule(lr){2-5}
\cmidrule(lr){6-9}
\# step
& \multicolumn{2}{c}{Gemini 3.1 Pro}
& \multicolumn{2}{c}{DeepSeek-v4-Pro}
& \multicolumn{2}{c}{$L^\star$}
& \multicolumn{2}{c}{TTT} \\
\cmidrule(lr){2-3}
\cmidrule(lr){4-5}
\cmidrule(lr){6-7}
\cmidrule(lr){8-9}
& \multicolumn{1}{c}{Tool Call} & \multicolumn{1}{c}{Resp.}
& \multicolumn{1}{c}{Tool Call} & \multicolumn{1}{c||}{Resp.}
& \multicolumn{1}{c}{Tool Call} & \multicolumn{1}{c|}{Resp.}
& \multicolumn{1}{c}{Tool Call} & Resp. \\
\midrule
1  & \multicolumn{1}{|c}{EQ($\emptyset$)} & \multicolumn{1}{c|}{$bba$}
   & MQ($\epsilon$) & \multicolumn{1}{c||}{0}
   & MQ($b$) & \multicolumn{1}{c|}{1}
   & MQ($\epsilon$) & 0 \\

2  & \multicolumn{1}{|c}{EQ($\Sigma^*$)} & \multicolumn{1}{c|}{$aaa$}
   & MQ($a$) & \multicolumn{1}{c||}{0}
   & MQ($a$) & \multicolumn{1}{c|}{0}
   & EQ($\emptyset$) & $bba$ \\

3  & \multicolumn{1}{|c}{EQ($b \in w$)} & \multicolumn{1}{c|}{$\top$}
   & MQ($b$) & \multicolumn{1}{c||}{1}
   & MQ($\epsilon$) & \multicolumn{1}{c|}{0}
   & MQ($bba$) & 1 \\

4  & \multicolumn{2}{|c|}{---}
   & MQ($aa$) & \multicolumn{1}{c||}{0}
   & MQ($bb$) & \multicolumn{1}{c|}{1}
   & MQ($ba$) & 1 \\

5  & \multicolumn{2}{|c|}{---}
   & MQ($ab$) & \multicolumn{1}{c||}{1}
   & MQ($ba$) & \multicolumn{1}{c|}{1}
   & MQ($a$) & 0 \\

6  & \multicolumn{2}{|c|}{---}
   & MQ($ba$) & \multicolumn{1}{c||}{1}
   & EQ($b \in w$) & \multicolumn{1}{c|}{$\top$}
   & MQ($b$) & 1 \\

7  & \multicolumn{2}{|c|}{---}
   & MQ($bb$) & \multicolumn{1}{c||}{1}
   & \multicolumn{2}{c|}{---}
   & MQ($aa$) & 0 \\

8  & \multicolumn{2}{|c|}{---}
   & MQ($aba$) & \multicolumn{1}{c||}{1}
   & \multicolumn{2}{c|}{---}
   & MQ($baa$) & 1 \\

9  & \multicolumn{2}{|c|}{---}
   & MQ($aaa$) & \multicolumn{1}{c||}{0}
   & \multicolumn{2}{c|}{---}
   & EQ($b \in w$) & $\top$ \\

10 & \multicolumn{2}{|c|}{---}
   & EQ($b \in w$) & \multicolumn{1}{c||}{$\top$}
   & \multicolumn{2}{c|}{---}
   & \multicolumn{2}{c}{---} \\
\bottomrule
\end{tabular}
\caption{
\textbf{Example interactions with an oracle} (\emph{Resp.} column), for the hidden DFA accepting all words containing at least one $b$.
--- indicates that the learner has already solved the problem.
Gemini 3.1 Pro solves the task using only equivalence queries, whereas DeepSeek-v4-Pro, $L^\star$, and TTT use both membership and equivalence queries.
In EQ notation, $\emptyset$ denotes the empty language, $\Sigma^*$ denotes all words, and $b \in w$ denotes the language of words containing $b$.
}
\label{tab:example-trajectories}
\end{table*}

%% file: latex/sections/03_task_formulation.tex
\section{Agentic Automata Learning}
In this section we formally define the task of \emph{agentic automata learning}, which recasts the classic interactive automata learning for agentic tool-calling LLMs~(\S\ref{sec:task-def}). A representative example of such an interaction is shown on the left-hand side of Table~\ref{tab:example-trajectories}. We then describe how task instances can be generated automatically at arbitrary scale by sampling DFAs while controlling for their complexity, and discuss how the task requires important agentic capabilities, such as adaptive planning, memory organization, and efficient tool use.

\paragraph{Task definition.}
\label{sec:task-def}
Let $\Sigma$ be a finite alphabet, 
$\mathcal{D}_\Sigma$ be the set of all DFAs over $\Sigma$, and
$A^\star \in \mathcal{D}_\Sigma$ be a minimal DFA inducing the target language $L^\star \subseteq \Sigma^*$. 
An agent interacts with an oracle  
$\mathcal{O}_{A^\star}$ through a sequence of tool calls, each termed a query. 

At each time step 
$t = 1,\ldots,T$, the agent chooses one of two tool calls:
\emph{(1) Membership Query}, $\mathrm{MQ}:\Sigma^* \mapsto \{0,1\}$ for some word 
    $w \in \Sigma^*$, to which the oracle returns 
    $\mathbbm{1}_{L^\star}\{w\}$.
 \emph{(2) Equivalence Query}, $\mathrm{EQ}:\mathcal{D}_\Sigma  \mapsto \{\top\} \cup \Sigma^*$ for some hypothesis DFA 
    $A$, to which the oracle returns either $\top$ if 
    $L(A) = L^\star$, or a counterexample 
    $w \in L(A) \triangle L^\star$ if they differ.

The agent's prompt at time $t$ contains the entire interaction history:
\[
h_{t-1} = ((a_1,o_1),\ldots,(a_{t-1},o_{t-1})),
\]
where $a_i$ is the query and $o_i$ the oracle response at step $i$. See prompt examples in Appendix~\ref{sec:interaction_example}.

The goal of the agent is to identify the target language by issuing an 
equivalence query $\mathrm{EQ}(\hat{A})$ such that:
\[
L(\hat{A}) = L^\star.
\]
An interaction is successful if the oracle returns $\top$ for an equivalence query 
within a query budget $T$, while its cost is the number of queries issued up to and including the accepted equivalence query.

The complexity of an interaction is defined by the number of states in $A^\star$. This is motivated by the classic active automata learning problem, where the number of states is a central factor affecting the efficiency of algorithms such as L* and TTT.

\paragraph{Automatic generation of instances.}
A key advantage of agentic automata learning is that task instances can be generated procedurally and at arbitrary scale, while controlling for their complexity through properties of the target DFA, such as the number of states.  In particular, we use the Boltzmann sampling method, which samples uniformly from the space of accessible DFAs over a binary alphabet,\footnote{A DFA is accessible if every state can be reached from the initial state by some input word.} and then applies rejection sampling to filter out non-minimal DFAs~\citep{BassinoNicaud2007}. While in our experiments we keep the alphabet fixed, this sampling procedure itself does not inherently restrict the alphabet size. This procedural construction allows us to create arbitrarily many environments over a chosen alphabet, rather than relying on a fixed pool of predefined tasks or examples requiring expensive human annotation.

\paragraph{Agentic automata learning provides a controlled, simple, yet challenging environment for evaluating core agentic capabilities.}
We intentionally simplify the environment, scaffold, and harness to isolate the model's inherent limitations while retaining the core demands of interactive problem solving. To discover the latent structure defined by a hidden DFA, an agent must combine adaptive planning, reasoning from observations, memory organization, and tool use. Because each query produces new feedback, the agent must select subsequent queries that reduce uncertainty about the target DFA, infer structural properties from the accumulated history of accepted and rejected words, and maintain consistency with prior observations by avoiding contradictory hypotheses and redundant queries. Although DFAs are less complex than many real-world environments, this controlled setting makes failures in these capabilities easier to identify and analyze. Moreover, the minimal harness and toolset limit the extent to which agents can compensate for weaknesses in parametric reasoning through external support.

%% file: latex/sections/04_evaluation_framework.tex
\section{Experimental Setup}
In this section, we describe the different design choices and evaluation criteria used to assess LLMs on the agentic automata learning task. Additional details on the experimental setup are provided in Appendix~\ref{app:experimental_setup}.

\paragraph{Stateless vs. stateful agent scaffolds.}
We evaluate a multi-turn agent under two scaffolds, which we refer to as \emph{stateless} and \emph{stateful} scaffolds. In the \emph{stateless scaffold}, at each step, the model receives the full history of previous actions and environment feedback and is equipped with membership and equivalence query tools for obtaining new information from the environment. The \emph{stateful scaffold} adds a ReAct-style state-tracking mechanism~\citep{yao2022react,rozanov-rei-2025-stateact}. As illustrated in Figure~\ref{fig:stateful_scaffold_mechanism}, its output format requires the model at every step to explicitly maintain its current goal, a compact working-memory representation of the accumulated state, and the reasoning used to select the next tool call. These fields are carried forward and updated across interaction steps, allowing intermediate structure to be preserved rather than repeatedly recovered from the interaction history.

\input{latex/figures/p1_p2_figure}

\paragraph{Models and metrics.}
We evaluate three advanced models, with their reasoning configuration set to high: DeepSeek-V4-Pro~\citep{deepseek_v4_pro}, a 1.6T-parameter model, Gemini 3.1 Pro Preview~\citep{gemini_31_pro}, and Gemini 3.8 Flash~\citep{gemini_3_8_flash}. We selected these models to cover frontier-level reasoning capabilities while keeping the full evaluation computationally feasible across our experimental setup. We did not include OpenAI or Claude models because evaluating their high-reasoning configurations at the same scale would have substantially increased the total API cost.

We evaluate two aspects of model performance. First, \textit{Success Rate} measures the rate at which models successfully identify a hidden DFA: 
\begin{equation}
\small
Success Rate = \frac{1}{N}\sum_{i=1}^{N} Succ(M, x_i),
\end{equation}

where \(Succ(M, x_i)\) is a binary indicator of whether model \(M\) successfully infers the hidden DFA \(x_i\), and \(N\) is the number of sampled automata.

Second, we measure \emph{interaction efficiency} by the difference between the number of tool calls made by a model $M$ when it finds the hidden DFA and the minimum number of tool calls made by the deterministic automata-learning algorithms $L^*$ and TTT~\cite{angluin1987learning, isberner2014ttt}:

\begin{equation}
\small
\Delta \text{\emph{Tool Calls from Deterministic Algorithms}} =
\frac{1}{|S_{\mathrm{succ}}|}
\sum_{i\in S_{\mathrm{succ}}}
\left(
T_M(x_i)-
\min\left\{
T_{L^*}(x_i),
T_{\mathrm{TTT}}(x_i)
\right\}
\right),
\end{equation}

where \(S_{\mathrm{succ}}\) denotes successful runs, and \(T_A(x_i)\) denotes the number of tool calls used by algorithm \(A\) to identify the hidden DFA \(x_i\).

\paragraph{Query budget.}
The number of queries allowed in each instance is defined to be at most twice the number of queries required by the better of \(L^*\) and TTT for the corresponding DFA, both of which have provable upper bounds on the number of queries. 
In particular, L* uses \(O(|\Sigma| n^2 m)\) queries~\citep{angluin1987learning}, while TTT uses \(O(|\Sigma| n^2 + n \log m)\) queries~\citep{isberner2014ttt}, where \(n\) is the number of states in the minimal hidden DFA, \(|\Sigma|\) is the alphabet size, and \(m\) is the maximum counterexample length.
Defining the budget as twice the number of queries guarantees that a correct solution exists within the allocated query budget. Throughout the evaluation, we discard any malformed tool call and do not count it against the query budget.

\paragraph{Dataset construction.}
We sample an evaluation dataset consisting of 160 task instances evenly distributed into 4 complexity levels, defined by the number of states in the minimal DFA: 2--3, 4--5, 6--7, and 8--9. Although additional instances can be generated automatically, evaluation remains expensive in runtime and monetary cost. We decide to evaluate our models on 160 instances to base our conclusions on a larger set of tasks and reduce dependence on the outcomes of individual instances, while keeping the evaluation feasible. This scale is also consistent with other benchmarks for long-horizon agent tasks, such as AgencyBench, which contains 138 instances~\citep{li2601agencybench}. To quantify uncertainty around the performance estimates and assess the stability of the observed trends with respect to variation in the sampled instances, we report 95\% confidence intervals computed using bootstrap with 100,000 resamples~\citep{hesterberg2015teachers}.

\paragraph{Oracle counterexample selection strategy.}
For each equivalence query, if the proposed DFA is not equivalent to the hidden DFA, the oracle must return a counterexample (a word on which the two DFAs disagree), yet  the classic task formulation does not define \emph{which} counterexample it should return. We return a counterexample designed to balance two considerations. On one hand, the length of the minimal counterexample directly affects the runtime of the classic learning algorithms and, consequently, the query budget allocated to the model and the overall evaluation cost. On the other hand, a minimal counterexample provides additional implicit information. Intuitively, if the oracle consistently returns a minimal counterexample of length $n$, then the two DFAs agree on all words of length $< n$, allowing the agent to infer their membership without explicitly querying them. We therefore sample a short counterexample at random.

\input{latex/figures/combined_success_tools_figure}

%% file: latex/figures/p1_p2_figure.tex
% TikZ recreation of the supplied scaffold figure.
% Add to the document preamble:
%   \usepackage{tikz}
%   \usepackage{xcolor}
%   \usepackage{graphicx}
%
% This version is vector-based (no external PNGs are required).

\begin{figure*}[b!]
    \centering
    \resizebox{\textwidth}{!}{%
    \begin{tikzpicture}[x=1cm,y=1cm]
        \definecolor{scaffoldpeach}{HTML}{F2B07D}
        \definecolor{scaffoldyellow}{HTML}{FDE49A}

        % Global typography for the figure.
        \tikzset{
            figtext/.style={font=\sffamily},
            box/.style={draw=black, line width=0.9pt},
            centertext/.style={figtext, align=center},
            lefttext/.style={figtext, align=left},
            sidecat/.style={centertext, font=\sffamily\bfseries\fontsize{8.2}{9.4}\selectfont},
            mainboxtext/.style={lefttext, font=\sffamily\fontsize{6.9}{8.2}\selectfont}
        }

        % ---------------------------
        % Column headings
        % ---------------------------
        \node[figtext, font=\sffamily\bfseries\fontsize{12.3}{13.8}\selectfont, anchor=north] at (2.03,9.96)
            {\underline{Stateless Scaffold}};
        \node[figtext, font=\sffamily\bfseries\fontsize{12.3}{13.8}\selectfont, anchor=north] at (9.03,9.96)
            {\underline{Output Example}};
        \node[figtext, font=\sffamily\bfseries\fontsize{12.3}{13.8}\selectfont, anchor=north] at (16.08,9.96)
            {\underline{Stateful Scaffold}};

        % ---------------------------
        % Left: stateless scaffold
        % ---------------------------
        \draw[box, fill=scaffoldyellow] (0.00,3.00) rectangle (4.05,3.45);
        \node[sidecat] at (2.025,3.225) {Tool Action};

        \draw[box, fill=scaffoldyellow] (0.00,2.15) rectangle (4.05,2.85);
        \node[sidecat] at (2.025,2.50) {Oracle Feedback};

        % ---------------------------
        % Middle: output example
        % ---------------------------
        % Goal / instruction box
        \draw[box] (4.28,8.47) rectangle (13.76,9.17);
        \node[mainboxtext, anchor=north west, text width=9.13cm, inner sep=0pt] at (4.43,9.02) {%
            Reconstruct the hidden target DFA exactly by efficiently calling ... using a\\
            minimal number of calls within the budget of 138.
        };

        % Working-memory / observation-table box
        \draw[box] (4.28,4.49) rectangle (13.76,8.32);
        \node[mainboxtext, anchor=north west, text width=9.13cm, inner sep=0pt] at (4.43,8.16) {%
            L* observation table: $S=\{\varepsilon, a, aa, aaa, aaab, aab, aaabb\}$, $E=\{\varepsilon,b,a\}$.\\
            Rows:\\
            r0($\varepsilon$): T,T,T\\
            r1(a): T,T,F\\
            r2(aa/aaabb): F,F,T\\
            ...\\
            Transitions resolved:\\
            r0: a$\to$r1, b$\to$r0\\
            r1: a$\to$r2, b$\to$r3\\
            ...\\
            To ensure closure, need full rows for $S\!\cdot\!A$ entries aaabba (r2$\cdot$a) ...\\
            Query ``aaabbab'' (aaabba$\cdot$b) next.
        };

        % Thought box
        \draw[box] (4.28,3.60) rectangle (13.76,4.34);
        \node[mainboxtext, anchor=north west, text width=9.13cm, inner sep=0pt] at (4.43,4.18) {%
            Query ``aaabbab'' to fill the row for aaabba and verify it matches an\\
            existing state.
        };

        % Tool action box
        \draw[box] (4.28,3.00) rectangle (13.76,3.45);
        \node[mainboxtext, anchor=west, text width=9.13cm, inner sep=0pt] at (4.43,3.225) {%
            \detokenize{{"tool_name": "is_word_in_language", "input": {"word": "aaabbab"}}}
        };

        % Oracle feedback box
        \draw[box] (4.28,2.15) rectangle (13.76,2.85);
        \node[mainboxtext, anchor=west, text width=9.13cm, inner sep=0pt] at (4.43,2.50) {%
            \detokenize{{"tool_outputs": [{"tool_name": "is_word_in_language", "call_count": 30, "error": null, "output": {"word": "aaabbab", "accepted": false}}]}}
        };

        % ---------------------------
        % Right: stateful scaffold
        % ---------------------------
        \draw[box, fill=scaffoldpeach] (14.00,8.47) rectangle (18.08,9.17);
        \node[sidecat] at (16.04,8.82) {Goal};

        \draw[box, fill=scaffoldpeach] (14.00,4.49) rectangle (18.08,8.32);
        \node[sidecat] at (16.04,6.405) {Working Memory};

        \draw[box, fill=scaffoldpeach] (14.00,3.60) rectangle (18.08,4.34);
        \node[sidecat] at (16.04,3.97) {Thought};

        \draw[box, fill=scaffoldyellow] (14.00,3.00) rectangle (18.08,3.45);
        \node[sidecat] at (16.04,3.225) {Tool Action};

        \draw[box, fill=scaffoldyellow] (14.00,2.15) rectangle (18.08,2.85);
        \node[sidecat] at (16.04,2.50) {Oracle Feedback};

        % Thin border around the entire figure content.
        \draw[black, line width=0.5pt] (-0.15,1.95) rectangle (18.23,10.15);
    \end{tikzpicture}%
    }

    \caption{\textbf{Comparison of the stateless (left) and stateful (right) scaffolds along with an output example (middle).}
At each step, the stateless scaffold outputs the next tool action directly from the interaction history, whereas the stateful scaffold additionally exposes an explicit goal, working memory, and thought state.
Orange components are unique to the stateful scaffold, while yellow components are shared by both scaffold types.
The center panel presents an excerpt from a stateful DeepSeek run, in which the model maintains an $L^\ast$ observation table in working memory and uses it to select the next query.}
    \label{fig:stateful_scaffold_mechanism}
\end{figure*}

%% file: latex/figures/combined_success_tools_figure.tex
% Add these in the preamble if not already present:
% \usepackage{pgfplots}
% \pgfplotsset{compat=1.18}
% \usepackage{subcaption}
% \usepackage{xcolor}
% \usetikzlibrary{patterns}

\definecolor{figtaborange}{HTML}{FF7F0E}
\definecolor{figtabblue}{HTML}{1F77B4}
\definecolor{figgold}{HTML}{FFD700}

% Vertical confidence/error bar at x=#1 with an x-shift #2.
\newcommand{\FigureErrorBar}[4]{%
    \draw[black, line width=0.8pt]
        ([xshift=#2]axis cs:#1,#3) -- ([xshift=#2]axis cs:#1,#4);
    \draw[black, line width=0.8pt]
        ([xshift=\dimexpr#2-2.5pt\relax]axis cs:#1,#3) --
        ([xshift=\dimexpr#2+2.5pt\relax]axis cs:#1,#3);
    \draw[black, line width=0.8pt]
        ([xshift=\dimexpr#2-2.5pt\relax]axis cs:#1,#4) --
        ([xshift=\dimexpr#2+2.5pt\relax]axis cs:#1,#4);
}

\begin{figure*}[t!]

\centering

% Shared legend: color = model.
\begin{minipage}{0.96\textwidth}
\centering
\begin{tabular}{ccc}
\tikz[baseline=-0.45ex]{\filldraw[fill=figtaborange,draw=black] (0,0) rectangle (0.28,0.13);}~\scriptsize Gemini 3.1 Pro Preview (high) &
\tikz[baseline=-0.45ex]{\filldraw[fill=figtabblue,draw=black] (0,0) rectangle (0.28,0.13);}~\scriptsize Gemini 3.8 Flash (high) &
\tikz[baseline=-0.45ex]{\filldraw[fill=figgold,draw=black] (0,0) rectangle (0.28,0.13);}~\scriptsize DeepSeek-v4-Pro (high)
\end{tabular}
\end{minipage}

\vspace{0.55em}

% -----------------------------------------------------------------------------
% LEFT: Model performance -- grouped by DFA-state bucket.
% Bar order is fixed by descending performance in the 8--9-state bucket:
% G3.1 StateAct, G3.8 StateAct, DeepSeek StateAct,
% G3.8 Multi-turn, G3.1 Multi-turn, DeepSeek Multi-turn.
% -----------------------------------------------------------------------------
\begin{subfigure}[t]{0.60\textwidth}
\centering
\begin{tikzpicture}[trim axis left, trim axis right, baseline=(current bounding box.north)]
\begin{axis}[
    ybar,
    width=0.82\linewidth,
    height=4.9cm,
    scale only axis,
    clip=false,
    ymin=0,
    ymax=110,
    ytick={0,20,40,60,80,100},
    ylabel={Success rate (\%) $\rightarrow$},
    xlabel={Minimal DFA States},
    xmin=0.45,
    xmax=4.55,
    xtick={1,2,3,4},
    xticklabels={2--3,4--5,6--7,8--9},
    x tick label style={font=\scriptsize},
    bar width=5.0pt,
    enlarge x limits=false,
    grid=major,
    grid style={gray!18},
    tick label style={font=\scriptsize},
    label style={font=\small},
]

% 1) Gemini 3.1 Pro Preview (high) - StateAct
\addplot+[
    fill=figtaborange, draw=black, bar shift=-13.75pt,
    postaction={pattern=north east lines,pattern color=black}
] coordinates {(1,100) (2,85) (3,85) (4,57.5)};

% 2) Gemini 3.8 Flash (high) - StateAct
\addplot+[
    fill=figtabblue, draw=black, bar shift=-8.25pt,
    postaction={pattern=north east lines,pattern color=black}
] coordinates {(1,92.5) (2,72.5) (3,57.5) (4,40)};

% 3) DeepSeek-v4-Pro (high) - StateAct
\addplot+[
    fill=figgold, draw=black, bar shift=-2.75pt,
    postaction={pattern=north east lines,pattern color=black}
] coordinates {(1,100) (2,97.5) (3,100) (4,37.5)};

% 4) Gemini 3.8 Flash (high) - Multi-turn
\addplot+[fill=figtabblue, draw=black, bar shift=2.75pt]
coordinates {(1,95) (2,85) (3,70) (4,30)};

% 5) Gemini 3.1 Pro Preview (high) - Multi-turn
\addplot+[fill=figtaborange, draw=black, bar shift=8.25pt]
coordinates {(1,100) (2,80) (3,75) (4,27.5)};

% 6) DeepSeek-v4-Pro (high) - Multi-turn
\addplot+[fill=figgold, draw=black, bar shift=13.75pt]
coordinates {(1,90) (2,65) (3,52.5) (4,15)};

% Bootstrapped 95% confidence intervals aligned to each bar.
% Gemini 3.1 Pro - StateAct
\FigureErrorBar{1}{-13.75pt}{100}{100}
\FigureErrorBar{2}{-13.75pt}{72.5}{95}
\FigureErrorBar{3}{-13.75pt}{72.5}{95}
\FigureErrorBar{4}{-13.75pt}{42.5}{72.5}

% Gemini 3.8 Flash - StateAct
\FigureErrorBar{1}{-8.25pt}{82.5}{100}
\FigureErrorBar{2}{-8.25pt}{57.5}{85}
\FigureErrorBar{3}{-8.25pt}{42.5}{72.5}
\FigureErrorBar{4}{-8.25pt}{25}{55}

% DeepSeek-v4 - StateAct
\FigureErrorBar{1}{-2.75pt}{100}{100}
\FigureErrorBar{2}{-2.75pt}{92.5}{100}
\FigureErrorBar{3}{-2.75pt}{100}{100}
\FigureErrorBar{4}{-2.75pt}{22.5}{52.5}

% Gemini 3.8 Flash - Multi-turn
\FigureErrorBar{1}{2.75pt}{87.5}{100}
\FigureErrorBar{2}{2.75pt}{72.5}{95}
\FigureErrorBar{3}{2.75pt}{55}{82.5}
\FigureErrorBar{4}{2.75pt}{17.5}{45}

% Gemini 3.1 Pro - Multi-turn
\FigureErrorBar{1}{8.25pt}{100}{100}
\FigureErrorBar{2}{8.25pt}{67.5}{92.5}
\FigureErrorBar{3}{8.25pt}{60}{87.5}
\FigureErrorBar{4}{8.25pt}{15}{42.5}

% DeepSeek-v4 - Multi-turn
\FigureErrorBar{1}{13.75pt}{80}{97.5}
\FigureErrorBar{2}{13.75pt}{50}{80}
\FigureErrorBar{3}{13.75pt}{37.5}{67.5}
\FigureErrorBar{4}{13.75pt}{5}{27.5}

% Compact manual legend: same visual style as before, with tighter spacing.
\node[
    anchor=north east,
    fill=white,
    fill opacity=0.92,
    text opacity=1,
    draw=black!25,
    inner xsep=2.5pt,
    inner ysep=1.5pt
] at (rel axis cs:0.965,0.905) {%
\begingroup
\tiny
\setlength{\tabcolsep}{1.5pt}%
\renewcommand{\arraystretch}{0.90}%
\begin{tabular}{@{}cl@{}}
\raisebox{-0.15ex}{\tikz{\node[draw=black,fill=white,pattern=north east lines,pattern color=black,line width=0.55pt,minimum width=0.38cm,minimum height=0.18cm,inner sep=0pt] {};}} & Stateful \\
\raisebox{-0.15ex}{\tikz{\node[draw=black,fill=white,line width=0.55pt,minimum width=0.38cm,minimum height=0.18cm,inner sep=0pt] {};}} & Stateless
\end{tabular}%
\endgroup
};

\node[font=\scriptsize, text=gray!70, anchor=north] at (rel axis cs:0.50,0.985) {Error bars: bootstrapped 95\% CI, 100,000 resamples};
\end{axis}
\end{tikzpicture}
\end{subfigure}%
\hspace{-0.045\textwidth}%
% -----------------------------------------------------------------------------
% RIGHT: Delta tool calls from the minimum of L* and TTT
% -----------------------------------------------------------------------------
\begin{subfigure}[t]{0.305\textwidth}
\centering
\vspace*{0.10cm}
\begin{tikzpicture}[trim axis left, trim axis right, baseline=(current bounding box.north)]
\begin{axis}[
    width=0.98\linewidth,
    height=4.9cm,
    scale only axis,
    ymin=-7,
    ymax=62,
    ytick={0,10,20,30,40,50,60},
    axis y line*=right,
    yticklabel pos=right,
    ylabel near ticks,
    ylabel={$\leftarrow$ $\Delta$ Tool Calls from deterministic algorithms},
    ylabel style={font=\scriptsize},
    xlabel={Minimal DFA States},
    xlabel style={yshift=-1pt},
    xtick={1,2,3,4},
    xticklabels={2--3,4--5,6--7,8--9},
    x tick label style={rotate=0,anchor=north,font=\scriptsize,yshift=-4pt},
    enlarge x limits=0.10,
    grid=major,
    grid style={gray!18},
    tick label style={font=\scriptsize},
]

% Shaded standard-deviation bands (\pm 1 SD) for the right subplot.

% G3.1 Pro - StateAct
\addplot[draw=none, fill=figtaborange, fill opacity=0.13]
coordinates {(1,3.90) (2,11.13) (3,23.48) (4,33.99)
             (4,-4.86) (3,-5.25) (2,-2.60) (1,-1.55)} -- cycle;

% DeepSeek-v4 - StateAct
\addplot[draw=none, fill=figgold, fill opacity=0.15]
coordinates {(1,6.89) (2,17.96) (3,27.28) (4,58.23)
             (4,12.57) (3,2.37) (2,2.55) (1,-0.74)} -- cycle;

% G3.8 Flash - StateAct
\addplot[draw=none, fill=figtabblue, fill opacity=0.13]
coordinates {(1,6.59) (2,21.32) (3,31.60) (4,47.86)
             (4,8.52) (3,7.61) (2,5.02) (1,0.01)} -- cycle;

% G3.1 Pro - Multi-turn
\addplot[draw=none, fill=figtaborange, fill opacity=0.24]
coordinates {(1,3.34) (2,11.01) (3,12.57) (4,34.99)
             (4,-0.44) (3,-4.70) (2,-2.70) (1,-1.34)} -- cycle;

% DeepSeek-v4 - Multi-turn
\addplot[draw=none, fill=figgold, fill opacity=0.24]
coordinates {(1,7.87) (2,25.24) (3,43.42) (4,51.53)
             (4,36.81) (3,13.34) (2,10.15) (1,0.80)} -- cycle;

% G3.8 Flash - Multi-turn
\addplot[draw=none, fill=figtabblue, fill opacity=0.24]
coordinates {(1,8.05) (2,24.91) (3,39.43) (4,47.68)
             (4,20.82) (3,16.07) (2,8.97) (1,-0.63)} -- cycle;

% StateAct: dashed lines + triangles.
\addplot+[
    color=figtaborange, dashed, line width=1.6pt,
    mark=triangle*, mark size=2.0pt,
    mark options={solid,fill=figtaborange,draw=figtaborange}
] coordinates {(1,1.18) (2,4.26) (3,9.12) (4,14.57)};

\addplot+[
    color=figgold, dashed, line width=1.6pt,
    mark=triangle*, mark size=2.0pt,
    mark options={solid,fill=figgold,draw=figgold}
] coordinates {(1,3.08) (2,10.26) (3,14.82) (4,35.40)};

\addplot+[
    color=figtabblue, dashed, line width=1.6pt,
    mark=triangle*, mark size=2.0pt,
    mark options={solid,fill=figtabblue,draw=figtabblue}
] coordinates {(1,3.30) (2,13.17) (3,19.61) (4,28.19)};

% Multi-turn: solid lines + circles.
\addplot+[
    color=figtaborange, solid, line width=1.6pt,
    mark=*, mark size=1.8pt,
    mark options={solid,fill=figtaborange,draw=figtaborange}
] coordinates {(1,1.00) (2,4.16) (3,3.93) (4,17.27)};

\addplot+[
    color=figgold, solid, line width=1.6pt,
    mark=*, mark size=1.8pt,
    mark options={solid,fill=figgold,draw=figgold}
] coordinates {(1,4.33) (2,17.69) (3,28.38) (4,44.17)};

\addplot+[
    color=figtabblue, solid, line width=1.6pt,
    mark=*, mark size=1.8pt,
    mark options={solid,fill=figtabblue,draw=figtabblue}
] coordinates {(1,3.71) (2,16.94) (3,27.75) (4,34.25)};

% Compact manual legend.
\node[
    anchor=north west,
    fill=white,
    fill opacity=0.92,
    text opacity=1,
    draw=black!25,
    inner xsep=2.5pt,
    inner ysep=1.5pt
] at (rel axis cs:0.04,0.905) {%
\begingroup
\tiny
\setlength{\tabcolsep}{1.5pt}%
\renewcommand{\arraystretch}{0.90}%
\begin{tabular}{@{}cl@{}}
\raisebox{0.35ex}{\tikz{\draw[gray!70,dashed,line width=1.15pt] (0,0) -- (0.22,0);}} & Stateful \\
\raisebox{0.35ex}{\tikz{\draw[gray!70,solid,line width=1.15pt] (0,0) -- (0.22,0);}} & Stateless
\end{tabular}%
\endgroup
};

\node[
    font=\tiny,
    text=gray!70,
    anchor=north
] at (rel axis cs:0.50,0.985) {Shaded bands show $\pm 1$ standard deviation};

% No explicit y=0 reference line is drawn.

% Draw the missing left frame edge explicitly while keeping the Y axis on the right.
\draw[black, line width=1.35pt] (rel axis cs:0,0) -- (rel axis cs:0,1);

\end{axis}
\end{tikzpicture}
\end{subfigure}

\caption{\textbf{Model performance and query efficiency as the number of states in the minimal DFA increases.}
Left: success rate for DFA-state buckets 2--3, 4--5, 6--7, and 8--9, with $n=40$ runs for each bucket.
Right: mean $\Delta$ Tool Calls relative to deterministic algorithms $(\mathrm{L^*},\mathrm{TTT})$. Positive values indicate more tool calls than the more query-efficient of classic strategies.}

\label{fig:success_and_tools_by_states}
\end{figure*}

%% file: latex/sections/05_results.tex
\section{Results}

Figure~\ref{fig:success_and_tools_by_states} reports \textit{Success Rate} which measures whether models identify the hidden DFA, and \textit{$\Delta$ Tool Calls from Deterministic Algorithms} which measures interaction efficiency. 
We complement this analysis by differentiating between  planning and reasoning errors and examining non-informative query rates in Figure~\ref{fig:error_analysis_combined}. Together, these results characterize task difficulty and reveal systematic differences both across models and across scaffolds. We first highlight the main findings, then turn to a detailed error and cost analysis.

\paragraph{\emph{Agentic automata learning} is challenging for state-of-the-art models in both stateless and stateful scaffolds}. Figure~\ref{fig:success_and_tools_by_states} shows that, under both stateless and stateful scaffolds, the performance of all models generally declines as the number of states in the minimal DFA increases. For automata with 8 or 9 states, the highest success rate is 57.5\% under the stateful scaffold and 30.0\% under the stateless scaffold. The gap relative to classic algorithms is also reflected in interaction efficiency: in the same 8--9-state range, when considering only successful runs, even the most efficient model-scaffold combination, Gemini 3.1 Pro with a stateful scaffold, uses approximately 23.5\% more tool calls than TTT.

\paragraph{Stateful scaffolding generally improves performance over stateless scaffolding on the agentic automata learning task and changes the relative ranking of models.}
Figure~\ref{fig:success_and_tools_by_states} shows that the stronger models, DeepSeek-v4-Pro and Gemini 3.1 Pro, benefit substantially from the stateful scaffold, both in success rate and interaction efficiency. One possible explanation is that the agent does not need to repeatedly re-integrate all accumulated information at every step. As illustrated in Figure~\ref{fig:stateful_scaffold_mechanism}, the model maintains an observation table similar to that used by the classic $L^*$ algorithm in its working memory. Based on this table, it chooses to query whether the word \emph{"aaabbab"} belongs to the target language in order to fill in a missing table entry. In contrast, the smaller Gemini 3.8 Flash model is harmed by the addition of state. This may be since the more complex structured output format makes it harder for the model to consistently satisfy the scaffold requirements. This interpretation is supported by the increase in its invalid tool-call rate from 0.2\% to 8.7\% when state tracking is added.

\paragraph{\emph{LLM agents identify $L^*$ as the appropriate solution, but struggle to execute it reliably over long sequences.}}
In a complementary evaluation, we give the models access to coding tools through the OpenHands harness~\citep{DBLP:conf/iclr/0001LSXTZPSLSTL25} without Internet access, and evaluate whether they can identify prior work on automata learning and reproduce a classic algorithm for the task. \emph{All examined models identify and write the code for the classic $L^*$ algorithm}. Thus, the lack in performance does not stem from a lack of knowledge of the classic algorithms or from failing to recognize them as an appropriate solution. However, this knowledge does not translate into reliable execution during interaction: since $L^*$ is deterministic, we can compare the sequence of queries generated by a model with the sequence generated by each algorithm on the same task instance. We find no task instance in which a model exactly reproduces the same query sequence. In addition, when solving the task using these algorithms, the number of equivalence queries is bounded by the number of states in the hidden DFA, and the size of hypotheses submitted through EQs increases monotonically throughout the learning process. Every model--scaffold combination violates at least one of these properties in a substantial fraction of runs. Even the combination closest to classic behavior, DeepSeek-v4-Pro with a stateful scaffold, satisfies both properties in only 31\% of runs. This conclusion is consistent with prior findings which showed that LLMs struggle to reliably execute algorithms over long sequences of interactive steps, even when the algorithm is provided to them~\citep{DBLP:journals/tmlr/YangZX25}.

\subsection{Error Analysis}
We conduct several analyses to better understand the performance of the models, shed light on which aspects of the task are challenging, and  outline avenues for future research.

\paragraph{Models fail both in planning which queries to make and in inferring structure from available evidence.}
We classify model failures into two mutually exclusive categories: planning failures and reasoning failures. A \emph{planning failure} occurs when the model fails and none of the classic passive learning algorithms, RPNI, EDSM, or Blue-Fringe, introduced in Section~\ref{sec:passive_learning}, can infer the hidden DFA from the observations accumulated by the model~\citep{oncina1992rpni,lang1998edsm}. In this case, we assume that the model did not collect sufficient information during interaction. In contrast, a \emph{reasoning failure} occurs when the model fails, but at least one passive learning algorithm can infer the hidden DFA from the accumulated observations. Thus, the necessary information was available, but the model failed to infer the correct DFA from it.
Figure~\ref{fig:error_analysis_combined} shows that all evaluated models suffer from both reasoning and planning failures. In general, failure rates are lower under the stateful scaffold, suggesting that explicit state tracking generally improves the models' ability to maintain progress through the interaction process.

\input{latex/figures/task_completion_outcomes_single_axis.tex}

\paragraph{All tested LLMs increasingly make non-informative queries as interactions progress, while stateful scaffolding reduces but does not eliminate this behavior.}

We measure how consistently LLMs use the information accumulated during interaction through the notion of \emph{non-informative queries}: tool calls whose answers can be implied by previous oracle responses and therefore cannot provide new evidence.
For membership queries, a query is non-informative if the same word has already been queried. For equivalence queries, a query is non-informative if it duplicates a previous hypothesis, or if the proposed DFA is already contradicted by known evidence, i.e., there exists a previously queried word that the hypothesis classifies differently from the oracle. 
% For each model, we compute the percentage of non-informative queries issued at each interaction step across all task instances.
Figure~\ref{fig:error_analysis_combined} shows that the rate of non-informative queries increases as interactions grow longer, suggesting that as evidence accumulates in context, LLM agents increasingly struggle to organize and use it consistently, while explicit state tracking helps mitigate this difficulty. 

% while the stateful scaffold reduces the prevalence of this phenomenon across all evaluated models without eliminating it. \gabist{After step 80, non-informative queries remain common even under Stateful, averaging approximately 10.9\%--19.8\% across models, compared with 16.4\%--58.7\% under Stateless. This pattern appears across all evaluated LLMs, whereas classic algorithms maintain 0\% non-informative queries by construction.} 

\input{latex/figures/budget_factor_two_subplots}

\subsection{Cost Analysis}
A central question in current LLM evaluation is the computational and monetary cost. This issue is especially important for agents, where each  instance may involve a long sequence of interactions. Figure~\ref{fig:cost_runtime_tokens_by_paired_states} shows runtime and cost increase with DFA size, reaching up to about 16 million tokens, 20 hours, and \$50 per run. Overall, the total monetary cost of the full experimental suite was \$3000. The Stateful scaffold reduces cost across models because it reduces the number of reasoning tokens required at each step: the model can preserve useful intermediate information instead of repeatedly reconstructing it from the full interaction history. The largest reduction is observed for Gemini 3.1 Pro Preview, whose average cost per run decreases from \$10 to \$3.25.

\textbf{Figure~\ref{fig:budget_factor_two_subplots} shows that evaluation cost increases with the query budget, while success rates saturate at different budgets across models.} We analyze the need for such a budget and find that additional budget is critical for solving DFAs that models cannot recover under smaller budgets. Specifically, DeepSeek-v4-Pro improves from 3.1\% at 100\% to 55.6\% at 200\%, with each increase in budget yielding a substantial improvement of similar magnitude in success rate. In contrast, Gemini 3.1 Pro Preview reaches 54.4\% success already at 125\% of the baseline budget, after which the gains taper, reaching 70.6\% at 200\%.

%% file: latex/figures/task_completion_outcomes_single_axis.tex
% Combined side-by-side figure:
% left panel = non-informative query rate by step
% right panel = task completion outcomes
%
% Required packages in the parent document:
%   \usepackage{xcolor}
%   \usepackage{tikz}
%   \usepackage{pgfplots}
%   \pgfplotsset{compat=1.18}
%   \usetikzlibrary{plotmarks}

\definecolor{figtaborange}{HTML}{FF7F0E}
\definecolor{figtabblue}{HTML}{1F77B4}
\definecolor{figgold}{HTML}{FFD700}

\definecolor{outcomesuccess}{HTML}{6FBF73}
\definecolor{outcomereasoning}{HTML}{F2B84B}
\definecolor{outcomeplanning}{HTML}{E57373}

\pgfplotstableread[row sep=\\, col sep=space]{
step gpro_stateful gpro_stateless deepseek_stateful deepseek_stateless flash_stateful flash_stateless\\
1 0.31 1.84 0.38 2.91 0.46 1.99\\
2 0.31 1.84 0.38 2.91 0.61 1.99\\
3 0.31 1.84 0.38 2.91 0.69 1.99\\
4 0.31 1.84 0.38 2.91 1.00 1.99\\
5 0.31 1.84 0.38 2.91 0.38 1.99\\
6 0.31 1.84 0.38 2.91 0.46 2.68\\
7 0.31 1.84 0.38 2.91 0.61 3.22\\
8 0.31 1.84 0.38 2.91 0.77 3.53\\
9 0.31 1.84 0.38 2.91 1.00 3.68\\
10 0.31 1.84 0.61 2.91 1.23 3.76\\
11 0.31 1.84 0.77 2.91 1.53 3.60\\
12 0.31 1.84 0.77 2.91 1.99 3.37\\
13 0.31 1.84 0.84 2.91 2.68 4.45\\
14 0.46 1.84 0.92 2.91 2.99 4.45\\
15 0.54 1.84 1.15 2.91 3.22 4.45\\
16 0.61 1.84 1.46 2.91 2.38 4.45\\
17 0.61 1.84 1.69 2.91 2.38 5.06\\
18 0.61 1.84 1.84 2.91 2.61 5.82\\
19 0.69 1.84 1.92 2.91 2.84 6.51\\
20 0.77 1.84 1.92 2.91 2.84 7.13\\
21 0.92 1.84 1.84 2.91 3.07 7.59\\
22 1.07 1.84 2.15 2.91 3.37 8.05\\
23 0.54 1.84 2.15 2.91 3.68 8.58\\
24 0.54 2.15 1.76 2.91 4.14 9.12\\
25 0.54 2.53 2.76 4.06 4.52 9.43\\
26 0.54 2.91 2.99 4.06 5.06 9.35\\
27 0.61 4.29 3.22 4.06 5.52 8.89\\
28 0.92 5.06 2.53 4.06 5.75 8.20\\
29 1.38 6.13 2.53 4.45 5.90 7.59\\
30 1.76 7.20 2.76 4.83 5.82 7.20\\
31 2.15 8.20 2.84 5.44 5.82 7.13\\
32 2.45 9.35 3.07 5.98 5.52 7.28\\
33 2.68 10.58 3.22 6.59 5.52 7.51\\
34 2.76 12.34 3.30 7.20 5.14 7.82\\
35 2.84 13.95 3.37 8.66 4.60 8.05\\
36 2.84 15.33 3.53 9.27 4.22 8.28\\
37 2.76 16.17 3.83 10.12 4.06 8.43\\
38 2.76 16.48 4.22 10.81 4.22 8.81\\
39 2.84 16.71 4.60 11.27 4.68 9.27\\
40 2.84 17.09 4.91 11.80 5.29 9.96\\
41 2.84 17.78 5.29 12.49 5.82 10.65\\
42 2.84 18.93 5.75 13.41 6.05 11.27\\
43 2.99 20.39 6.21 14.18 6.05 11.88\\
44 2.99 21.92 6.51 14.49 5.98 12.34\\
45 2.76 23.53 6.74 14.33 5.67 12.95\\
46 2.45 25.29 6.97 14.18 5.36 13.34\\
47 2.38 27.44 7.13 14.41 5.14 13.57\\
48 2.38 29.81 7.13 14.95 5.14 13.80\\
49 2.99 32.42 7.36 15.41 5.29 14.03\\
50 3.83 34.72 8.05 15.79 5.67 14.18\\
51 4.98 36.56 8.97 16.02 6.21 14.18\\
52 5.98 37.71 10.04 16.25 6.59 13.87\\
53 7.36 38.24 10.58 16.71 6.82 13.34\\
54 7.36 38.70 10.50 17.40 6.97 12.80\\
55 7.51 39.39 10.27 18.24 6.97 12.26\\
56 6.82 40.62 9.96 18.93 7.28 12.19\\
57 6.44 42.23 9.73 19.39 7.66 12.19\\
58 5.67 43.99 9.66 20.00 8.05 12.34\\
59 5.29 45.60 9.66 20.69 7.89 12.57\\
60 5.44 46.75 10.04 21.38 7.51 12.72\\
61 5.82 47.82 10.65 22.07 7.20 13.11\\
62 7.28 48.90 11.88 22.76 6.90 13.49\\
63 8.20 49.97 13.11 23.07 6.82 13.87\\
64 9.20 51.04 14.64 22.99 6.67 14.10\\
65 9.81 52.19 14.79 22.61 6.44 14.26\\
66 9.89 53.19 14.03 22.23 6.13 14.41\\
67 9.81 53.80 13.72 22.00 5.82 14.64\\
68 9.43 53.96 13.34 22.00 5.44 14.87\\
69 8.81 53.34 13.11 22.23 5.44 15.18\\
70 8.05 52.58 13.41 22.69 5.90 15.48\\
71 7.20 51.89 14.18 23.22 6.44 15.71\\
72 6.13 52.12 14.87 23.68 6.97 15.86\\
73 5.67 53.27 15.25 23.91 7.51 15.79\\
74 5.67 55.41 15.18 24.22 7.89 15.64\\
75 6.21 57.56 14.72 24.60 8.35 15.33\\
76 7.59 59.32 13.80 25.45 9.04 15.25\\
77 9.12 60.55 12.57 26.75 9.66 15.41\\
78 10.88 61.16 11.57 28.13 10.27 15.56\\
79 12.72 61.08 11.34 29.20 10.81 15.41\\
80 14.33 60.16 11.57 29.74 11.34 15.18\\
81 17.17 58.71 11.57 29.58 11.80 14.87\\
82 19.39 57.41 12.80 29.20 12.03 14.64\\
83 20.85 56.87 14.18 28.89 12.03 14.95\\
84 20.69 57.41 16.63 29.05 11.96 15.71\\
85 19.62 58.40 18.85 29.20 11.50 16.63\\
86 18.32 59.32 20.62 28.97 10.50 17.40\\
87 17.17 59.86 21.54 28.43 9.27 17.47\\
88 17.78 60.09 21.46 27.82 8.43 17.01\\
89 18.55 60.32 21.00 27.51 8.66 16.25\\
90 19.47 60.55 20.46 27.74 9.43 15.56\\
91 19.93 60.85 20.08 27.97 10.42 15.25\\
92 18.93 61.08 19.77 27.90 11.11 15.33\\
93 17.55 60.93 19.70 27.44 11.42 15.94\\
94 15.79 60.09 19.93 26.75 11.42 16.86\\
95 15.10 58.32 20.62 26.52 11.11 17.86\\
96 15.86 56.10 21.77 27.05 11.11 18.55\\
97 18.09 54.42 23.61 27.97 11.11 18.32\\
98 21.54 54.65 25.75 28.89 11.11 17.47\\
99 24.68 56.87 25.75 28.89 11.27 16.63\\
100 29.35 59.47 28.89 30.20 11.27 16.17\\
}\nonInformativeStepTable

\pgfplotsset{
    outcomeaxishalf/.style={
        xbar stacked,
        width=0.82\linewidth,
        height=2.95cm,
        xmin=0,
        xmax=100,
        xtick={0,20,40,60,80,100},
        ytick=data,
        y dir=reverse,
        bar width=6.5pt,
        grid=major,
        grid style={gray!16},
        tick label style={font=\scriptsize},
        label style={font=\scriptsize},
        axis line style={black!60},
        tick style={black!60},
        yticklabel style={
            font=\tiny,
            text width=1.95cm,
            align=right
        },
        enlarge y limits=0.26,
    }
}

\begin{figure*}[t!]
\centering

% ================================================================
% LEFT PANEL: 50%
% ================================================================
\begin{minipage}[t]{0.49\textwidth}
\vspace{0pt}
\centering

{\footnotesize\bfseries (a) Query-level failures\par}
\vspace{0.25em}

% Fixed-height content box.
\begin{minipage}[t]{\linewidth}
\centering

% Compact model legend -- all models on one row.
\begin{minipage}[c][0.58cm][c]{\linewidth}
\makebox[\linewidth][l]{%
\hspace{0.175\linewidth}%
\fontsize{5.2}{5.8}\selectfont
\begin{tabular}{@{}c@{\hspace{0.45em}}c@{\hspace{0.45em}}c@{}}
\tikz[baseline=-0.5ex]{\draw[figtaborange,line width=1.15pt] (0,0.055)--(0.20,0.055);}
~Gemini 3.1 Pro
&
\tikz[baseline=-0.5ex]{\draw[figtabblue,line width=1.15pt] (0,0.055)--(0.20,0.055);}
~Gemini 3.8 Flash
&
\tikz[baseline=-0.5ex]{\draw[figgold,line width=1.15pt] (0,0.055)--(0.20,0.055);}
~DeepSeek-v4-Pro
\end{tabular}%
}
\end{minipage}

\vspace{0.08em}

\begin{tikzpicture}
\begin{axis}[
    width=0.99\linewidth,
    height=5.45cm,
    xmin=1,
    xmax=100,
    ymin=0,
    ymax=105,
    xlabel={Step number},
    ylabel={Non-informative queries (\%)},
    grid=major,
    grid style={gray!18},
    tick label style={font=\scriptsize},
    label style={font=\scriptsize},
]

\addplot[color=figtaborange,line width=1.25pt,dashed]
table[x=step,y=gpro_stateful] {\nonInformativeStepTable};

\addplot[color=figtaborange,line width=1.25pt,solid]
table[x=step,y=gpro_stateless] {\nonInformativeStepTable};

\addplot[color=figgold,line width=1.25pt,dashed]
table[x=step,y=deepseek_stateful] {\nonInformativeStepTable};

\addplot[color=figgold,line width=1.25pt,solid]
table[x=step,y=deepseek_stateless] {\nonInformativeStepTable};

\addplot[color=figtabblue,line width=1.25pt,dashed]
table[x=step,y=flash_stateful] {\nonInformativeStepTable};

\addplot[color=figtabblue,line width=1.25pt,solid]
table[x=step,y=flash_stateless] {\nonInformativeStepTable};

% Compact in-plot scaffold legend: black dashed vs. black solid.
% The line samples are drawn manually so they never inherit a model color.
\node[
    anchor=north west,
    draw=black!35,
    line width=0.3pt,
    fill=white,
    text=black,
    inner xsep=2.0pt,
    inner ysep=1.2pt,
    font=\fontsize{6.2}{6.8}\selectfont,
] at (rel axis cs:0.025,0.975) {%
    \begin{tabular}{@{}c@{\hspace{0.32em}}l@{}}
    \tikz[baseline=-0.45ex]{\draw[draw=black,line width=0.8pt,dash pattern=on 2.5pt off 1.6pt] (0pt,0pt)--(22pt,0pt);} & Stateful \\[-0.15ex]
    \tikz[baseline=-0.45ex]{\draw[draw=black,line width=0.8pt,solid] (0pt,0pt)--(22pt,0pt);} & Stateless
    \end{tabular}%
};

\end{axis}
\end{tikzpicture}

\end{minipage}
\end{minipage}
\hspace{0.002\textwidth}
% ================================================================
% RIGHT PANEL: 50%
% ================================================================
\begin{minipage}[t]{0.49\textwidth}
\vspace{0pt}
\centering

{\footnotesize\bfseries (b) Outcome-level failures\par}
\vspace{0.25em}

% Same fixed height as the left panel.
\begin{minipage}[t]{\linewidth}
\centering

% Outcome legend, shifted to the right.
\begin{minipage}[c][0.58cm][c]{\linewidth}
\makebox[\linewidth][l]{%
\hspace{0.30\linewidth}%
\fontsize{5.4}{5.9}\selectfont
\begin{tabular}{@{}c@{\hspace{0.7em}}c@{\hspace{0.7em}}c@{}}
\tikz[baseline=-0.5ex]{\draw[draw=white,fill=outcomesuccess] (0,0) rectangle (0.22,0.11);}
~Success
&
\tikz[baseline=-0.5ex]{\draw[draw=white,fill=outcomereasoning] (0,0) rectangle (0.22,0.11);}
~Reasoning Failure
&
\tikz[baseline=-0.5ex]{\draw[draw=white,fill=outcomeplanning] (0,0) rectangle (0.22,0.11);}
~Planning Failure
\end{tabular}%
}
\end{minipage}

\vspace{-0.02em}

% Stateful outcomes
\begin{tikzpicture}
\begin{axis}[
    outcomeaxishalf,
    symbolic y coords={
        DeepSeek-v4-Pro,
        Gemini 3.1 Pro,
        Gemini 3.8 Flash
    },
    xlabel={Percentage of task instances (Stateful)},
]

\addplot+[draw=white,line width=0.35pt,fill=outcomesuccess]
coordinates {
    (83.750,{DeepSeek-v4-Pro})
    (81.875,{Gemini 3.1 Pro})
    (65.625,{Gemini 3.8 Flash})
};

\addplot+[draw=white,line width=0.35pt,fill=outcomereasoning]
coordinates {
    (5.625,{DeepSeek-v4-Pro})
    (13.750,{Gemini 3.1 Pro})
    (25.625,{Gemini 3.8 Flash})
};

\addplot+[draw=white,line width=0.35pt,fill=outcomeplanning]
coordinates {
    (10.625,{DeepSeek-v4-Pro})
    (4.375,{Gemini 3.1 Pro})
    (8.750,{Gemini 3.8 Flash})
};
\end{axis}
\end{tikzpicture}

\vspace{0.52em}

% Stateless outcomes
\begin{tikzpicture}
\begin{axis}[
    outcomeaxishalf,
    symbolic y coords={
        Gemini 3.1 Pro,
        Gemini 3.8 Flash,
        DeepSeek-v4-Pro
    },
    xlabel={Percentage of task instances (Stateless)},
]

\addplot+[draw=white,line width=0.35pt,fill=outcomesuccess]
coordinates {
    (70.625,{Gemini 3.1 Pro})
    (70.000,{Gemini 3.8 Flash})
    (55.625,{DeepSeek-v4-Pro})
};

\addplot+[draw=white,line width=0.35pt,fill=outcomereasoning]
coordinates {
    (20.000,{Gemini 3.1 Pro})
    (16.875,{Gemini 3.8 Flash})
    (20.625,{DeepSeek-v4-Pro})
};

\addplot+[draw=white,line width=0.35pt,fill=outcomeplanning]
coordinates {
    (9.375,{Gemini 3.1 Pro})
    (13.125,{Gemini 3.8 Flash})
    (23.750,{DeepSeek-v4-Pro})
};
\end{axis}
\end{tikzpicture}

\end{minipage}
\end{minipage}

\vspace{-0.15em}
\caption{
\textbf{Error analysis at two levels: query level on the left and task-instance level on the right.}
The left panel shows the percentage of queries at each interaction step that are non-informative across all task instances, for each model and scaffold.
Curves are smoothed using a 1D Gaussian filter ($\sigma = 2$) for visualization.
We report values up to step 100, as only a small number of task instances continue beyond this point.
The right panel shows the percentage of task instances classified as successful runs, reasoning failures, or planning failures.
}
\label{fig:error_analysis_combined}
\end{figure*}

%% file: latex/figures/budget_factor_two_subplots.tex
% Requires in the main document:
% \usepackage{tikz}
% \usepackage{pgfplots}
% \pgfplotsset{compat=1.18}

\definecolor{aaibudgetorange}{HTML}{FF7F0E}
\definecolor{aaibudgetblue}{HTML}{1F77B4}
\definecolor{aaibudgetgold}{HTML}{FFD700}

\begin{figure*}[t!]
\centering

% ------------------------------------------------------------------
% Shared model-color legend
% ------------------------------------------------------------------
\makebox[\textwidth][c]{%
\begin{tabular}{@{}c@{\hspace{2.0em}}c@{\hspace{2.0em}}c@{}}
\tikz[baseline=-0.5ex]{
    \draw[aaibudgetorange,line width=1.4pt] (0,0.065)--(0.32,0.065);
}
~\tiny Gemini 3.1 Pro Preview (high) -- Stateless
&
\tikz[baseline=-0.5ex]{
    \draw[aaibudgetblue,line width=1.4pt] (0,0.065)--(0.32,0.065);
}
~\tiny Gemini 3.8 Flash (high) -- Stateless
&
\tikz[baseline=-0.5ex]{
    \draw[aaibudgetgold!85!black,line width=1.4pt] (0,0.065)--(0.32,0.065);
}
~\tiny DeepSeek-v4-Pro (high) -- Stateless
\end{tabular}%
}

\vspace{0.45em}

% ==================================================================
% (a) Left: average cost per task instance -- Stateless only
% ==================================================================
\begin{minipage}[t]{0.49\textwidth}
\centering
\begin{tikzpicture}
\begin{axis}[
    width=\linewidth,
    height=4.2cm,
    ymin=0,
    ymax=10.6,
    xmin=0.92,
    xmax=2.08,
    ylabel={Average cost per task instance (\$)},
    xlabel={Query budget (\%)},
    xtick={1,1.25,1.5,1.75,2},
    xticklabels={100\%,125\%,150\%,175\%,200\%},
    ytick={0,2,4,6,8,10},
    grid=major,
    grid style={gray!18},
    tick label style={font=\scriptsize},
    label style={font=\scriptsize},
]

% Gemini 3.1 Pro Preview -- Stateless
\addplot+[
    color=aaibudgetorange,
    line width=1.6pt,
    solid,
    mark=*,
    mark size=2.0pt,
    mark options={draw=aaibudgetorange,fill=aaibudgetorange}
]
coordinates {
    (1,3.530846)
    (1.25,5.256742)
    (1.5,6.943779)
    (1.75,8.442708)
    (2,10.021334)
};

% Gemini 3.8 Flash -- Stateless
\addplot+[
    color=aaibudgetblue,
    line width=1.6pt,
    solid,
    mark=*,
    mark size=2.0pt,
    mark options={draw=aaibudgetblue,fill=aaibudgetblue}
]
coordinates {
    (1,0.686583)
    (1.25,1.021722)
    (1.5,1.366033)
    (1.75,1.726696)
    (2,2.081235)
};

% DeepSeek-v4-Pro -- Stateless
\addplot+[
    color=aaibudgetgold!85!black,
    line width=1.6pt,
    solid,
    mark=*,
    mark size=2.0pt,
    mark options={draw=aaibudgetgold!85!black,fill=aaibudgetgold!85!black}
]
coordinates {
    (1,0.266624)
    (1.25,0.432168)
    (1.5,0.618029)
    (1.75,0.813226)
    (2,1.014015)
};

\end{axis}
\end{tikzpicture}

\vspace{0.2em}
\hspace*{0.9cm}{\small (a) Average cost per instance.}
\end{minipage}
\hfill
% ==================================================================
% (b) Right: success rate -- Stateless only
% ==================================================================
\begin{minipage}[t]{0.49\textwidth}
\centering
\begin{tikzpicture}
\begin{axis}[
    width=\linewidth,
    height=4.2cm,
    ymin=0,
    ymax=100,
    xmin=0.92,
    xmax=2.08,
    ylabel={Success rate (\%)},
    xlabel={Query budget (\%)},
    xtick={1,1.25,1.5,1.75,2},
    xticklabels={100\%,125\%,150\%,175\%,200\%},
    ytick={0,20,40,60,80,100},
    grid=major,
    grid style={gray!18},
    tick label style={font=\scriptsize},
    label style={font=\scriptsize},
]

% Gemini 3.1 Pro Preview -- Stateless
\addplot+[
    color=aaibudgetorange,
    line width=1.6pt,
    solid,
    mark=*,
    mark size=2.0pt,
    mark options={draw=aaibudgetorange,fill=aaibudgetorange}
]
coordinates {
    (1,23.750)
    (1.25,54.375)
    (1.5,65.625)
    (1.75,68.750)
    (2,70.625)
};

% Gemini 3.8 Flash -- Stateless
\addplot+[
    color=aaibudgetblue,
    line width=1.6pt,
    solid,
    mark=*,
    mark size=2.0pt,
    mark options={draw=aaibudgetblue,fill=aaibudgetblue}
]
coordinates {
    (1,6.250)
    (1.25,16.875)
    (1.5,31.875)
    (1.75,48.750)
    (2,70.000)
};

% DeepSeek-v4-Pro -- Stateless
\addplot+[
    color=aaibudgetgold!85!black,
    line width=1.6pt,
    solid,
    mark=*,
    mark size=2.0pt,
    mark options={draw=aaibudgetgold!85!black,fill=aaibudgetgold!85!black}
]
coordinates {
    (1,3.125)
    (1.25,12.500)
    (1.5,21.250)
    (1.75,38.125)
    (2,55.625)
};

\end{axis}
\end{tikzpicture}

\vspace{0.2em}
\hspace*{0.9cm}{\small (b) Success rate.}
\end{minipage}

\caption{
\textbf{Effect of the query budget on evaluation cost and success rate.} The budget ranges from 100\% to 200\% of the number of queries required by the more efficient classic learner between $L^\star$ and TTT on the corresponding task instance.
}
\label{fig:budget_factor_two_subplots}

\end{figure*}

%% file: latex/sections/07_related_work.tex
\section{Related Work}

Recent benchmarks test whether LLM agents can reconstruct hidden mechanisms from documentation and interactive black-box access~\citep{yang2026programbench,revengebench_2026,lehmann2026tycho}, but fixed query budgets need not make every instance identifiable, and real-world complexity obscures sources of failure. AQA-Bench tests whether LLMs can maintain state and execute a predefined algorithm across successive interactions~\citep{DBLP:journals/tmlr/YangZX25}, while work on in-context learning of hidden Markov models and probabilistic automata evaluates non-interactive behavioral approximation rather than exact structural recovery~\citep{akyurek2024context,dai2026pre}. Other work uses classic algorithms such as L* to extract DFAs from trained neural models serving as oracles~\citep{weiss2018extracting,chen2024llms,vazquez2025lstarlm}. Our framework instead makes the LLM the learner: it must recognize and retrieve an appropriate algorithm, adaptively query an external oracle, and execute the algorithm to reconstruct the hidden DFA exactly. Deriving query budgets from classic learning algorithms ensures every instance is solvable within its budget and supports controlled, fine-grained analysis of information gathering, planning, and reasoning.

%% file: latex/sections/08_conclusion.tex
\section{Conclusion}
We introduce \emph{agentic automata learning}, a controlled framework that models environments as hidden DFAs and uses membership and equivalence queries to evaluate both whether language models recover hidden structure and how they do so. Even strong models struggle as complexity increases, with failures in accuracy, planning, reasoning, tool use, and managing accumulated information. These results motivate evaluations of how agents search, update beliefs, and use feedback over time. Future work could extend the framework to nondeterministic or stochastic environments and use the framework for training agents that better interact with their environment.

% We introduce \emph{agentic automata learning}, a controlled framework for evaluating whether language models can infer hidden structure through interaction. By modeling environments as hidden DFAs and allowing membership and equivalence queries, the framework measures both final success and the discovery process. Our results show that even strong models struggle as complexity increases, with failures in accuracy, planning, reasoning, tool use, and management of accumulated information. These findings motivate evaluations that examine how agents search, update beliefs, and use feedback over time. 

% Future work can extend this framework beyond DFAs to environments with non-determinism or stochasticity. It could also relax the assumption of exact oracle feedback by introducing noisy, partial, delayed, or incorrect signals, or by replacing equivalence queries with weaker feedback. Finally, the framework could serve not only for inference-time evaluation, but also as a training environment, for example through reinforcement learning.

\section*{Reproducibility Statement}
We provide detailed information to facilitate reproduction of our experiments. The task formulation, procedural DFA generation, evaluation metrics, and query-budget construction are described in the main text. Additional experimental details are provided in Appendix~\ref{app:experimental_setup}, including the complete prompts for both the Stateless and Stateful scaffolds, an example interaction, exact model versions and inference settings, evaluated task instances, and the deterministic counterexample-sampling procedure. Our evaluation uses seeds 1--20 for each minimal DFA size from 2 to 9, and the query budget for each instance is set to twice the number of queries required by the better of $L^*$ and TTT. To support reproducibility, we publicly release our code as open source.

\section*{AI Use Statement}
Generative AI tools were used during the preparation of this work to assist with translation, language editing, software implementation and code editing, LaTeX formatting, and the presentation of experimental results. Generative AI was not used to generate datasets, formulate mathematical claims or proofs, propose research hypotheses, design the research methodology or experiments, or interpret the experimental results. All AI-assisted text and code were reviewed by the authors, and AI-assisted code was tested for correctness. All experimental results and scientific claims were independently verified by the authors. The authors take full responsibility for the final content of this work, including any text, code, or other artifacts produced with the assistance of generative AI.

%% file: latex/sections/Acknowledgments.tex
\section{Acknowledgments}
This research was supported in part by the Google Cloud Research Credits Program. We would like to thank Prof. Orna Kupferman for suggesting the automata-learning setting that inspired this work, and Prof. Dana Fisman and Gal Meirom for their valuable advice and insights.

%% file: latex/iclr_ai_use_statement.tex
% ICLR 2027 AI-use statement.
% This file is intentionally empty so no disclosure is fabricated for you.
% Before submission, add the required statement here, for example using an
% unnumbered section heading as requested by the ICLR 2027 instructions:
%
% \subsection*{AI use statement}
% <your accurate disclosure text>

%% file: latex/shared/appendix.tex
% Appendix shared by both formats.
% \clearpage
\section{Experimental Setup}
This appendix provides additional details on the experimental setup used throughout our evaluation. We first present the initial prompt templates used in the stateless and stateful settings, followed by an example interaction illustrating the expected interaction format. We then introduce a more general evaluation metric that goes beyond a binary notion of success or failure, and describe the procedure used to sample counterexamples. Finally, we analyze invalid tool calls, provide the model configurations used in our experiments, and examine the context-window size throughout the interaction.
\label{app:experimental_setup}

\input{latex/sections/appendix/01_prompt_template.tex}
\input{latex/sections/appendix/Example_interaction_format.tex}
\label{latex/appendix:symmetric_difference_similarity}
\input{latex/sections/appendix/02_best_hypothesis_similarity.tex}

\input{latex/sections/appendix/07_computational_cost.tex}
\input{latex/sections/appendix/03_counterexample_sampling.tex}

\input{latex/sections/appendix/invalid-tool-calls.tex}

\input{latex/sections/appendix/demo.tex}
\input{latex/sections/appendix/model_configuration_table.tex}
\input{latex/figures/context_window_analysis_figure.tex}

\clearpage
\input{latex/figures/cost_runtime_by_states_figure}
\clearpage
\input{latex/sections/appendix/query_budget_factor_overleaf.tex}

%% file: latex/sections/appendix/01_prompt_template.tex
\subsection{Initial Prompt Templates}
\label{sec:prompt}

\DefineVerbatimEnvironment{DFAPromptVerbatim}{Verbatim}{
  breaklines=true,
  fontsize=\scriptsize,
  baselinestretch=0.95,
  frame=single,
  framesep=2mm,
  commandchars=+@~
}

As described in the main text, we evaluate the agents under two scaffolds: \emph{stateless}, where the model receives the full interaction history at each step, and \emph{stateful}, which additionally introduces an explicit state-tracking mechanism. Below, we provide the complete prompt templates used for both scaffolds.
% =========================================================
\subsubsection{Stateless Scaffold}
\label{sec:prompt_stateless}

\begin{DFAPromptVerbatim}
You are a language model tasked with reconstructing an unknown Deterministic Finite Automaton (DFA).
A fixed target DFA exists, but its structure is hidden from you.

You can learn about the target DFA only through:
- External tools available to you.
- Your own internal reasoning.

Game rules:
- You have a total budget of {MAX_CALLS} budgeted DFA-learning tool calls.
- If you exceed the allowed number of budgeted DFA-learning calls, your final score is 0.
- On your final remaining budgeted DFA-learning call, you must invoke the evaluation tool.

Scoring policy:
- Using fewer tool calls while still reconstructing the correct DFA yields a higher score. Therefore, tools should be invoked efficiently and purposefully to obtain novel and informative results.
- If you believe you have already identified the correct DFA structure, you should invoke immediately the evaluation tool,
  even if you still have remaining calls available.

STRICT OUTPUT POLICY (MANDATORY):
- You MUST NOT output any explanations, reasoning, comments, or natural language text.
- You MUST output ONLY TOOL_ACTION blocks (Do NOT output any text) and exactly ONE TOOL_ACTION block.
- Each TOOL_ACTION block must contain a single JSON object with exactly:
  {"tool_name": "<tool_name>", "input": { ... }}
- Never include call_count (it is added externally).

# IMPORTANT: The <TOOL_ACTION> block must include exactly one tool call. Each block represents a single tool invocation, so multiple tool calls--either to the same tool or to different tools--are not allowed within the same block.

Tools:

[is_word_in_language]
Input JSON: { tool_name: 'is_word_in_language', call_count: int, input: { word: str } }
Output JSON: { tool_name: 'is_word_in_language', call_count: int, error: str|null, output: { word: str, accepted: bool }|null }
Checks whether the given word is accepted by the game's DFA. If the input word is the empty string "", the returned output.word will be +textquotedbl+ensuremath@+epsilon~+textquotedbl (epsilon).

[evaluate_dfa_candidate]
Input JSON: { tool_name: 'evaluate_dfa_candidate', call_count: int, input: { candidate_dfa: { states: [str], alphabet: [str], start_state: str, accept_states: [str], transitions: [[str, str, str]] } } }
Output JSON: { tool_name: 'evaluate_dfa_candidate', call_count: int, error: str|null, output: { score: float, optimal: bool, witness_word: str }|null }
Evaluates whether the provided candidate DFA is equivalent to the game's target DFA. If equivalent, output.optimal=true, output.score=1.0, and output.witness_word='' . If not equivalent, output.optimal=false, output.score=0.0, and output.witness_word is a counterexample word accepted by exactly one of the DFAs. If the witness word is the empty string, output.witness_word will be +textquotesingle+ensuremath@+epsilon~+textquotesingle. The counterexample also has a learning role: it shows where the candidate DFA fails and helps refine the next hypothesis.

Hints:
- DFA vocabulary: {a, b}
\end{DFAPromptVerbatim}

\clearpage

% =========================================================
\subsubsection{Stateful Scaffold}
\label{sec:prompt_stateful}

\begin{DFAPromptVerbatim}
Goal: You are a language model tasked with reconstructing an unknown Deterministic Finite Automaton (DFA).
A fixed target DFA exists, but its structure is hidden from you.

You can learn about the target DFA only through:
- External tools available to you.
- Your own internal reasoning.

Game rules:
- You have a total budget of {MAX_CALLS} budgeted DFA-learning tool calls.
- If you exceed the allowed number of budgeted DFA-learning calls, your final score is 0.
- On your final remaining budgeted DFA-learning call, you must invoke the evaluation tool.

Scoring policy:
- Using fewer tool calls while still reconstructing the correct DFA yields a higher score. Therefore, tools should be invoked efficiently and purposefully to obtain novel and informative results.
- If you believe you have already identified the correct DFA structure, you should invoke immediately the evaluation tool,
  even if you still have remaining calls available.

STRUCTURED OUTPUT POLICY (MANDATORY):
- You MUST output exactly four blocks and nothing else, in this exact order:
  1. <GOAL>...</GOAL>
  2. <WORKING_MEMORY>...</WORKING_MEMORY>
  3. <THOUGHT>...</THOUGHT>
  4. <TOOL_ACTION>...</TOOL_ACTION>
- The <GOAL> block must restate the overall task objective that must remain fixed throughout the interaction: reconstruct the hidden target DFA exactly while using the available DFA-learning tool-call budget efficiently. Keep it concise and do not place reasoning or a tool call in this block.
- The <WORKING_MEMORY> block must contain a compact, up-to-date summary of the information and intermediate state that should be preserved for future steps. Update it on every turn as a complete replacement for the previous working-memory summary.
- The <THOUGHT> block must contain the reasoning used to choose the single next action.
- The <TOOL_ACTION> block must contain a single JSON object with exactly:
  {"tool_name": "<tool_name>", "input": { ... }}
- Never include call_count; it is added externally.
- Exactly one block of each type is allowed. Do not output markdown fences, comments, headings, or text outside the four blocks.
- Tool documentation later in the prompt may show a standalone <TOOL_ACTION> snippet only to illustrate that tool's JSON arguments. Those snippets are not complete response examples. In every actual response, you must still output all four required blocks.
# IMPORTANT: The <TOOL_ACTION> block must include exactly one tool call. Each block represents a single tool invocation, so multiple tool calls--either to the same tool or to different tools--are not allowed within the same block.

Tools:

[is_word_in_language]
Input JSON: { tool_name: 'is_word_in_language', call_count: int, input: { word: str } }
Output JSON: { tool_name: 'is_word_in_language', call_count: int, error: str|null, output: { word: str, accepted: bool }|null }
Checks whether the given word is accepted by the game's DFA. If the input word is the empty string "", the returned output.word will be +textquotedbl+ensuremath@+epsilon~+textquotedbl (epsilon).

[evaluate_dfa_candidate]
Input JSON: { tool_name: 'evaluate_dfa_candidate', call_count: int, input: { candidate_dfa: { states: [str], alphabet: [str], start_state: str, accept_states: [str], transitions: [[str, str, str]] } } }
Output JSON: { tool_name: 'evaluate_dfa_candidate', call_count: int, error: str|null, output: { score: float, optimal: bool, witness_word: str }|null }
Evaluates whether the provided candidate DFA is equivalent to the game's target DFA. If equivalent, output.optimal=true, output.score=1.0, and output.witness_word='' . If not equivalent, output.optimal=false, output.score=0.0, and output.witness_word is a counterexample word accepted by exactly one of the DFAs. If the witness word is the empty string, output.witness_word will be +textquotesingle+ensuremath@+epsilon~+textquotesingle. The counterexample also has a learning role: it shows where the candidate DFA fails and helps refine the next hypothesis.

Hints:
- DFA vocabulary: {a, b}
\end{DFAPromptVerbatim}

%% file: latex/sections/appendix/Example_interaction_format.tex
\clearpage
\subsection{Example Interaction Format}
\label{sec:interaction_example}

We provide an example interaction illustrating the protocol under which the model operates during a task instance:
\medskip

\begin{mdframed}[
  linewidth=0.8pt,
  roundcorner=3pt,
  innerleftmargin=10pt,
  innerrightmargin=10pt,
  innertopmargin=10pt,
  innerbottommargin=10pt,
  skipabove=10pt,
  skipbelow=10pt
]

\medskip

\textbf{Initial Prompt:} 
The interaction begins with the task prompt described in Appendix~\ref{sec:prompt}

\medskip

{\footnotesize
\textbf{Assistant:}
\begin{quote}
\scriptsize\ttfamily
\textless TOOL\_ACTION\textgreater\\
\{ "tool\_name": "is\_word\_in\_language",\\
\ \ "input": \{ "word": "" \} \}\\
\textless/TOOL\_ACTION\textgreater
\end{quote}
}

{\footnotesize
\textbf{User:}
\begin{quote}
\scriptsize\ttfamily
\textless TOOL\_RESULT\textgreater\\
\{"tool\_outputs": [\{\\
\ \ "tool\_name": "is\_word\_in\_language",\\
\ \ "call\_count": 1,\\
\ \ "error": null,\\
\ \ "output": \{"word": "$\epsilon$", "accepted": true\}
\}]\}\\
\textless/TOOL\_RESULT\textgreater
\end{quote}
}

\medskip

\noindent\textbf{Skipped tool calls:} 2--10.

\medskip

{\footnotesize
\textbf{Assistant:}
\begin{quote}
\scriptsize\ttfamily
\textless TOOL\_ACTION\textgreater\\
\{ "tool\_name": "evaluate\_dfa\_candidate",\\
\ \ "input": \{\\
\ \ \ \ "candidate\_dfa": \{\\
\ \ \ \ \ \ "states": ["q0","q1"],\\
\ \ \ \ \ \ "alphabet": ["a","b"],\\
\ \ \ \ \ \ "start\_state": "q0",\\
\ \ \ \ \ \ "accept\_states": ["q0"],\\
\ \ \ \ \ \ "transitions": [\\
\ \ \ \ \ \ \ \ ["q0","a","q0"],\\
\ \ \ \ \ \ \ \ ["q0","b","q1"],\\
\ \ \ \ \ \ \ \ ["q1","a","q0"],\\
\ \ \ \ \ \ \ \ ["q1","b","q1"]]\}\}\}\\
\textless/TOOL\_ACTION\textgreater
\end{quote}
}

{\footnotesize
\textbf{User:}
\begin{quote}
\scriptsize\ttfamily
\textless TOOL\_RESULT\textgreater\\
\{"tool\_outputs": [\{\\
\ \ "tool\_name": "evaluate\_dfa\_candidate",\\
\ \ "call\_count": 11,\\
\ \ "error": null,\\
\ \ "output": \{"score": 1.0,"optimal": true,"witness\_word": null\}\}]\}\\
\textless/TOOL\_RESULT\textgreater
\end{quote}
}

\medskip

\textbf{Termination:} 
The interaction ends once the evaluation tool indicates that the proposed DFA is equivalent to the hidden DFA, as reflected by "score": 1.0 and "optimal": true.

\end{mdframed}

\clearpage

%% file: latex/sections/appendix/02_best_hypothesis_similarity.tex
\subsection{Best Hypothesis Similarity Metric}

\definecolor{figtaborange}{HTML}{FF7F0E}
\definecolor{figtabblue}{HTML}{1F77B4}
\definecolor{figgold}{HTML}{FFD700}

\begin{figure}[b!]
\centering

\makebox[\columnwidth][c]{%
\begin{minipage}{0.98\columnwidth}
\centering
\begin{tabular}{ll}
\tikz[baseline=-0.5ex]{\draw[figtaborange, line width=1.4pt] (0,0.065)--(0.32,0.065);}
~\tiny Gemini 3.1 Pro Preview (high) &
\tikz[baseline=-0.5ex]{\draw[figgold, line width=1.4pt] (0,0.065)--(0.32,0.065);}
~\tiny DeepSeek-v4-Pro (high) \\[-0.2em]
\multicolumn{2}{c}{
\tikz[baseline=-0.5ex]{\draw[figtabblue, line width=1.4pt] (0,0.065)--(0.32,0.065);}
~\tiny Gemini 3.8 Flash (high)
}
\end{tabular}
\end{minipage}
}

\vspace{0.35em}

\begin{tikzpicture}
\begin{axis}[
    width=\columnwidth,
    height=5.8cm,
    ymin=20,
    ymax=105,
    enlarge y limits=false,
    ylabel={Best Hypothesis Similarity (\%)},
    xlabel={Number of States in the Minimal DFA},
    xtick={1,2,3,4},
    xticklabels={2--3,4--5,6--7,8--9},
    enlarge x limits=0.05,
    grid=major,
    grid style={gray!18},
    tick label style={font=\small},
    label style={font=\small},
    clip=true,
    legend style={
        at={(0.03,0.055)},
        anchor=south west,
        draw=black!25,
        fill=white,
        fill opacity=0.94,
        text opacity=1,
        font=\tiny,
        cells={anchor=west},
        row sep=-1pt,
        inner xsep=2pt,
        inner ysep=1.5pt,
    },
    legend image post style={xscale=1.55},
]

\addlegendimage{black, line width=1.2pt, dash pattern=on 4pt off 2pt, mark=triangle*, mark size=1.5pt}
\addlegendentry{Stateful}
\addlegendimage{black, line width=1.2pt, solid, mark=*, mark size=1.5pt}
\addlegendentry{Stateless}

\addplot[draw=none, fill=figtaborange, fill opacity=0.10, forget plot] coordinates {
    (1,100.00) (2,105.75) (3,103.23) (4,103.55) (4,78.49) (3,93.85) (2,87.28) (1,100.00)
} \closedcycle;

\addplot[draw=none, fill=figtaborange, fill opacity=0.10, forget plot] coordinates {
    (1,100.00) (2,105.43) (3,104.59) (4,100.70) (4,75.47) (3,87.28) (2,86.76) (1,100.00)
} \closedcycle;

\addplot[draw=none, fill=figgold, fill opacity=0.10, forget plot] coordinates {
    (1,100.00) (2,101.14) (3,100.00) (4,101.83) (4,76.36) (3,100.00) (2,98.43) (1,100.00)
} \closedcycle;

\addplot[draw=none, fill=figgold, fill opacity=0.10, forget plot] coordinates {
    (1,106.68) (2,104.80) (3,104.63) (4,96.32) (4,69.19) (3,74.68) (2,82.49) (1,87.57)
} \closedcycle;

\addplot[draw=none, fill=figtabblue, fill opacity=0.10, forget plot] coordinates {
    (1,102.53) (2,104.02) (3,104.45) (4,103.90) (4,73.63) (3,69.54) (2,91.77) (1,95.57)
} \closedcycle;

\addplot[draw=none, fill=figtabblue, fill opacity=0.10, forget plot] coordinates {
    (1,105.37) (2,104.08) (3,106.55) (4,100.02) (4,67.10) (3,76.80) (2,91.22) (1,91.79)
} \closedcycle;

% Random baseline
\draw[gray, dashed, line width=1.3pt]
(axis cs:0.85,50) -- (axis cs:4.15,50);

\node[
    anchor=north,
    font=\scriptsize,
    text=gray
] at (axis cs:2.5,48.5)
{random language proposals};

\addplot[color=figtaborange, line width=1.45pt, dash pattern=on 5.2pt off 2.3pt, mark=triangle*, mark size=1.7pt]
coordinates {(1,100.00) (2,96.52) (3,98.54) (4,91.02)};

\addplot[color=figtaborange, line width=1.45pt, solid, mark=*, mark size=1.7pt]
coordinates {(1,100.00) (2,96.10) (3,95.93) (4,88.09)};

\addplot[color=figgold, line width=1.45pt, dash pattern=on 5.2pt off 2.3pt, mark=triangle*, mark size=1.7pt]
coordinates {(1,100.00) (2,99.79) (3,100.00) (4,89.10)};

\addplot[color=figgold, line width=1.45pt, solid, mark=*, mark size=1.7pt]
coordinates {(1,97.12) (2,93.64) (3,89.66) (4,82.75)};

\addplot[color=figtabblue, line width=1.45pt, dash pattern=on 5.2pt off 2.3pt, mark=triangle*, mark size=1.7pt]
coordinates {(1,99.05) (2,97.90) (3,86.99) (4,88.77)};

\addplot[color=figtabblue, line width=1.45pt, solid, mark=*, mark size=1.7pt]
coordinates {(1,98.58) (2,97.65) (3,91.68) (4,83.56)};

\end{axis}
\end{tikzpicture}

\caption{Best Hypothesis Similarity as the number of states in the minimal DFA increases. The shaded regions represent one standard deviation, and the dashed gray line represents a random hypothesis baseline. Stateful runs are shown with dashed curves and Stateless runs with solid curves.}
\label{fig:best_hypothesis_similarity_appendix}
\end{figure}

To complement the binary success metric, we define a continuous similarity measure that captures how close the best hypothesis proposed during interaction is to the hidden DFA.

For each task instance, we compute the similarity between every intermediate hypothesis proposed by the model and the hidden DFA, and report the highest similarity value achieved throughout the interaction, referred to as \textit{Best Hypothesis Similarity}. This extends the binary success metric by providing a finer-grained measure of progress toward the hidden DFA.

Formally, for a hypothesis DFA \(h\) and a hidden DFA \(x\), we compute a symmetric-difference-based similarity approximation over all words up to length \(k=200\):

\[
\mathrm{Sim}_{\triangle,200}(h,x)
=
1-
\frac{|L(h)_{\leq 200}\triangle L(x)_{\leq 200}|}
{|\Sigma_{\leq 200}|}
\]

Here, \(L(h)_{\leq 200}\) and \(L(x)_{\leq 200}\) denote the sets of words of length at most 200 accepted by the hypothesis and hidden DFAs, respectively, \(\triangle\) denotes the symmetric difference between the two languages, and \(\Sigma_{\leq 200}\) denotes the set of all words over the alphabet whose length is at most 200.

This formulation measures the fraction of words on which the two DFAs agree, producing a similarity score in \([0,1]\), where higher values indicate greater behavioral agreement between the hypothesis and hidden languages. We use symmetric-difference similarity rather than Jaccard similarity because it is better aligned with the structure of our sampled DFA distribution. Since our DFA generation process produces languages in which, in expectation, a word has approximately equal probability of being accepted or rejected, the resulting language space is relatively balanced. In this setting, normalizing by the full input space \( |\Sigma_{\leq 200}| \) provides a stable and directly interpretable measure of behavioral agreement across the entire observable space. In contrast, Jaccard similarity normalizes by the union of accepted words, making it more sensitive to variation in language density and less suitable for our objective of measuring agreement over the full input space. Although the number of words of length at most 200 is exponential, approximately \(2^{200}\) for a binary alphabet, this metric can be computed efficiently using dynamic programming over the product DFA of the two DFAs. This enables counting the number of words in the symmetric difference without explicitly enumerating all possible strings.

Beyond aggregate evaluation, this metric enables tracking intermediate hypothesis similarity throughout interaction, as shown in Figure~\ref{fig:eq_similarity_example}. This reveals whether a model gradually approaches the target DFA or instead oscillates between hypotheses without consistent and meaningful improvement over time.

\begin{figure}[t!]
\centering
\begin{tikzpicture}
\begin{axis}[
    width=\columnwidth,
    height=6.5cm,
    xlabel={Step / call number},
    ylabel={Similarity},
    xmin=0,
    xmax=86,
    ymin=0,
    ymax=1.05,
    grid=major,
    grid style={gray!18},
    tick label style={font=\scriptsize},
    label style={font=\small},
    legend style={
        at={(0.98,0.02)},
        anchor=south east,
        font=\scriptsize
    }
]

\addplot[color=blue, mark=*, line width=1pt] coordinates {
(4,0.50) (11,0.50) (18,0.48) (37,0.50) (39,0.66)
(42,0.64) (43,0.50) (46,0.48) (48,0.52) (49,0.50)
(50,0.72) (53,0.77) (55,0.94) (57,0.77) (60,0.72)
(61,0.69) (62,0.70) (64,0.53) (65,0.42) (66,0.53)
(67,0.49) (68,0.66) (70,0.53) (71,0.50) (72,0.53)
(73,0.76) (74,0.66) (75,0.57) (76,0.63) (77,0.64)
(78,0.71) (79,0.65) (81,1.00)
};
\addlegendentry{LLM}

\addplot[color=orange, mark=*, line width=1pt] coordinates {
(6,0.50) (13,0.50) (30,0.46) (63,0.49) (84,1.00)
};
\addlegendentry{$L^\star$}

\addplot[color=green!60!black, mark=*, line width=1pt] coordinates {
(2,0.53) (6,0.50) (26,0.52) (38,0.65) (44,0.85)
(52,0.95) (59,1.00)
};
\addlegendentry{TTT}

\end{axis}
\end{tikzpicture}

\caption{
Example interaction trajectory for Gemini 3.1 Pro on a single task instance. The graph tracks the symmetric-difference similarity between the hidden DFA language and the hypotheses proposed throughout interaction.
}
\label{fig:eq_similarity_example}
\end{figure}

%% file: latex/sections/appendix/03_counterexample_sampling.tex
\subsection{How Counterexamples Are Sampled}
\label{sec:counterexample}

In each equivalence query between two DFAs, we construct the symmetric difference automaton, which accepts exactly the words on which the two automata disagree. A counterexample is a word accepted by this automaton.

We first compute the minimal length \(k_{\min}\) of a word accepted by the symmetric difference automaton. Instead of selecting a single word of this minimal length, we define a set of candidate words consisting of short words whose lengths lie in the range \(k_{\min}\) to \(k_{\min} + 3\).

The selection from this candidate set is performed in a pseudo-random yet deterministic manner. Specifically, for each given pair of automata (the target DFA and the hypothesis), we compute a fixed seed using a stable hash function over their representations. This seed is then used to select a single word from the candidate set.

This approach ensures stability by maintaining identical conditions across runs, while at the same time avoiding providing the model with additional information that could simplify the problem.

%% file: latex/sections/appendix/invalid-tool-calls.tex
\subsection{Invalid Tool-Call Analysis}
\label{app:invalid-tool-calls}

LLM agents express their queries and candidate DFAs through a JSON-based tool-calling protocol, unlike \(L^*\) and TTT, which operate directly on formal objects. We identify three broad types of invalid calls: \textit{formatting/JSON errors}, \textit{malformed DFA specifications}, and \textit{invalid tool or input errors}. Table~\ref{tab:invalid-tool-calls} reports the overall percentage of invalid tool calls for each model and run type.

\vspace{0em}

\begin{center}
\scriptsize
\begin{tabular*}{\columnwidth}{@{\extracolsep{\fill}}llr@{}}
\toprule
\textbf{Model} & \textbf{Run} & \textbf{Invalid Rate} \\
\midrule
Gemini 3.8 Flash (high) & Stateful & 8.7\% \\
DeepSeek-v4-Pro (high) & Stateless & 7.3\% \\
DeepSeek-v4-Pro (high) & Stateful & 4.6\% \\
Gemini 3.1 Pro Preview (high) & Stateful & 4.6\% \\
Gemini 3.1 Pro Preview (high) & Stateless & 1.3\% \\
Gemini 3.8 Flash (high) & Stateless & 0.2\% \\
\bottomrule
\end{tabular*}

\captionof{table}{Aggregate invalid tool-call rates by model and run type. Invalid calls include formatting or JSON errors, malformed DFA specifications, and invalid tools or input arguments.}
\label{tab:invalid-tool-calls}
\end{center}

%% file: latex/sections/appendix/demo.tex
% \section{Example Interaction in the Live Demo}
% \label{sec:demo}
% {
% \color{blue}
% [Reef:]
% Figure \ref{fig:interaction_example} shows an example interaction from our web interface, which allows users to run the agentic automata learning task with any regular expression as the target automaton. Alongside the agent-oracle dialogue, the interface provides real-time analysis of each query: whether it is informative, whether passive algorithms can already recover the hidden DFA from the observations collected so far, and, for equivalence queries, the similarity between the proposed hypothesis and the target language.
% }
% \begin{figure}[b!]
%     \centering
%     \caption{Interaction between a Gemini 3.1 Flash Lite agent and the oracle. The hidden target DFA is generated from the regular expression \texttt{((a|b)*a)}.}

%     \label{fig:interaction_example}
% \end{figure}

%% file: latex/sections/appendix/model_configuration_table.tex
% Model configuration appendix table for the main evaluation.
% Full-width, non-floating version so the appendix heading and table
% always remain on the same page.
% Required packages in the main document:
% \usepackage{array}
% \usepackage{booktabs}
% \usepackage{graphicx}
% \usepackage{hyperref}
% \usepackage{caption}

\onecolumn

\subsection{Model Configurations}
\label{sec:model_configuration_details}

% Center table contents both horizontally and vertically.
\newcolumntype{C}[1]{>{\centering\arraybackslash}m{#1}}

\begin{center}
\setlength{\tabcolsep}{3.5pt}
\renewcommand{\arraystretch}{1.74}
\scriptsize

\resizebox{\textwidth}{!}{%
\begin{tabular}{@{}
C{1.35cm}
C{3.75cm}
C{2.45cm}
C{3.20cm}
C{3.20cm}
C{1.75cm}
C{1.75cm}
C{4.15cm}
@{}}
\toprule
\textbf{Provider} &
\shortstack{\textbf{Exact model ID}\\\textbf{(click for documentation)}} &
\textbf{Run dates} &
\textbf{Reasoning parameters} &
\shortstack{\textbf{Other parameters}\\\textbf{(e.g., temperature)}} &
\shortstack{\textbf{Context}\\\textbf{token limit}} &
\shortstack{\textbf{Output}\\\textbf{token limit}} &
\textbf{Evaluated task instances} \\
\midrule

DeepSeek Platform &
\href{https://api-docs.deepseek.com/news/news260424}{\texttt{deepseek-v4-pro}} &
\shortstack{May 4--7, 2026\\July 11--12, 2026\\September 09--10, 2026} & 
\shortstack{\texttt{thinking.type=enabled}\\\texttt{reasoning\_effort=high}} &
vendor-default hyperparameters & 
1{,}000{,}000 &
384{,}000 &
\shortstack{Seeds 1--20 for each minimal DFA size 2--9\\2 distinct scaffolds\\(320 task instances total)} \\
\midrule

Google Cloud &
\href{https://storage.googleapis.com/deepmind-media/Model-Cards/Gemini-3-8-Flash-Model-Card.pdf}{\texttt{gemini-3.8-flash}} &
\shortstack{September 09--10, 2026} &
\texttt{thinking\_level=high} &
vendor-default hyperparameters &
1{,}048{,}576 &
65{,}536 &
\shortstack{Seeds 1--20 for each minimal DFA size 2--9\\2 distinct scaffolds\\(320 task instances total)} \\
\midrule

Google Cloud &
\href{https://storage.googleapis.com/deepmind-media/Model-Cards/Gemini-3-1-Pro-Model-Card.pdf}{\texttt{gemini-3.1-pro-preview}} &
\shortstack{May 4--7, 2026\\July 11--12, 2026\\September 09--10, 2026} &
\texttt{thinking\_level=high} &
vendor-default hyperparameters &
1{,}048{,}576 &
65{,}536 &
\shortstack{Seeds 1--20 for each minimal DFA size 2--9\\2 distinct scaffolds\\(320 task instances total)} \\

\bottomrule
\end{tabular}%
}

\captionof{table}{\textbf{Model configurations used in the evaluation.}
Model IDs are clickable and link to the corresponding vendor documentation.}
\label{tab:model_configurations}
\end{center}

%% file: latex/figures/context_window_analysis_figure.tex
% Requires:
% \usepackage{graphicx}
% \usepackage{subcaption}

\subsection{Context Window Analysis}
\label{sec:context_window}

\begin{figure}[!ht]
\centering

% -------------------- StateAct --------------------
{\small\bfseries Stateful\par}
\vspace{0.25em}

\begin{subfigure}[t]{0.32\textwidth}
    \centering
    \includegraphics[
        width=\linewidth,
        keepaspectratio
    ]{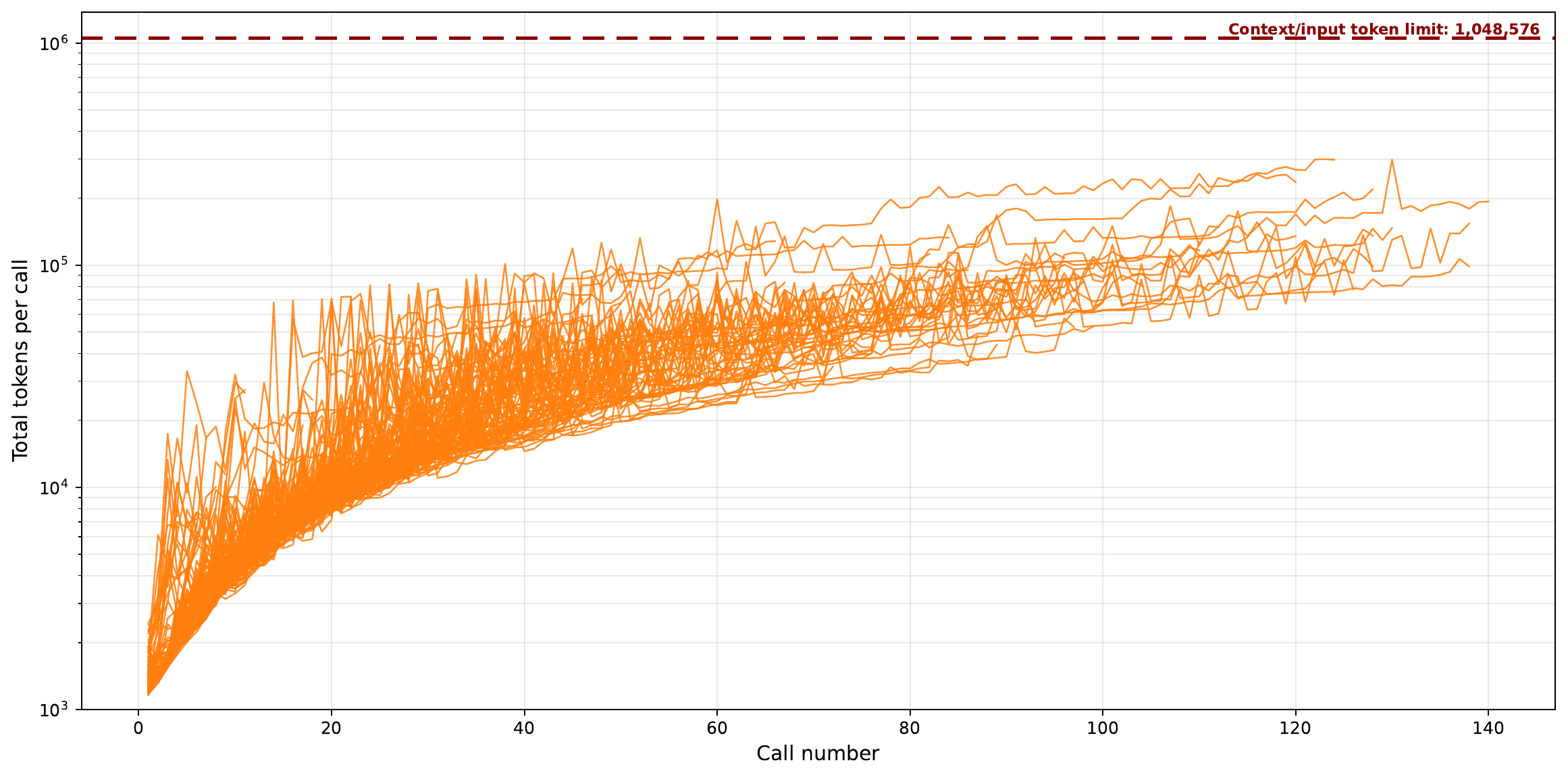}
    \caption{Gemini 3.1 Pro Preview (high)}
\end{subfigure}
\hfill
\begin{subfigure}[t]{0.32\textwidth}
    \centering
    \includegraphics[
        width=\linewidth,
        keepaspectratio
    ]{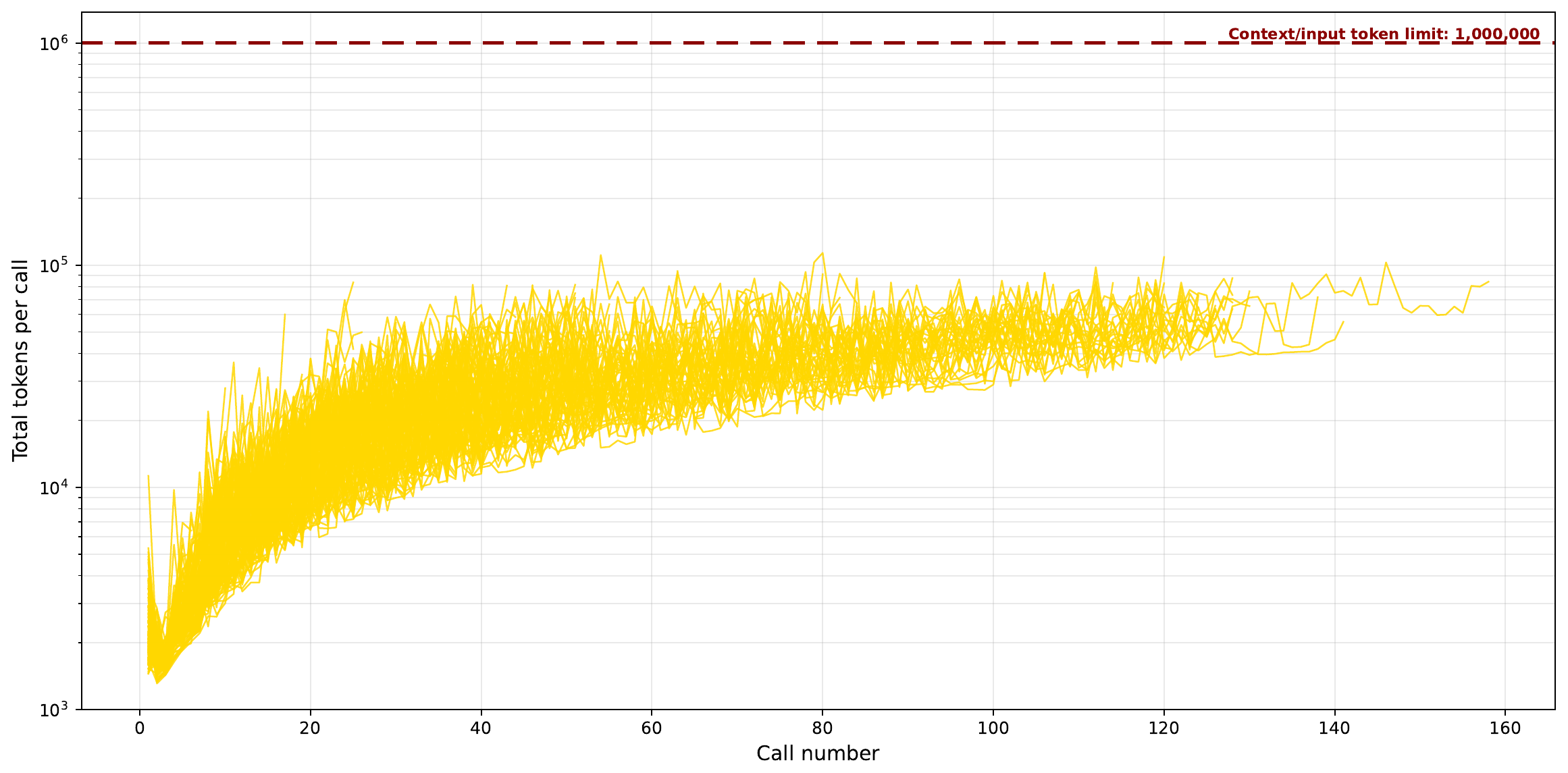}
    \caption{DeepSeek-v4-Pro (high)}
\end{subfigure}
\hfill
\begin{subfigure}[t]{0.32\textwidth}
    \centering
    \includegraphics[
        width=\linewidth,
        keepaspectratio
    ]{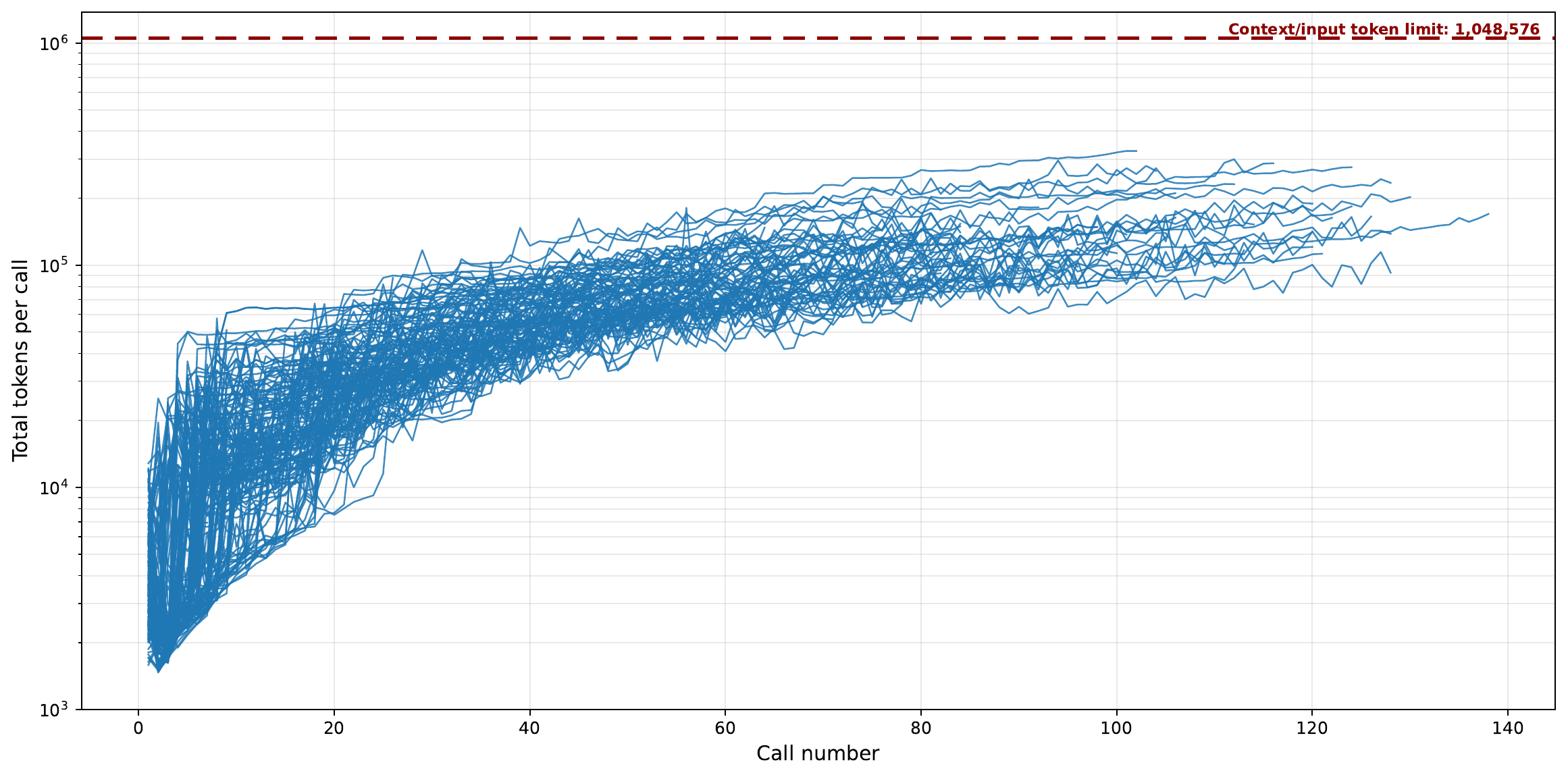}
    \caption{Gemini 3.8 Flash (high)}
\end{subfigure}

\vspace{0.75em}

% -------------------- Multi-turn --------------------
{\small\bfseries Stateless\par}
\vspace{0.25em}

\begin{subfigure}[t]{0.32\textwidth}
    \centering
    \includegraphics[
        width=\linewidth,
        keepaspectratio
    ]{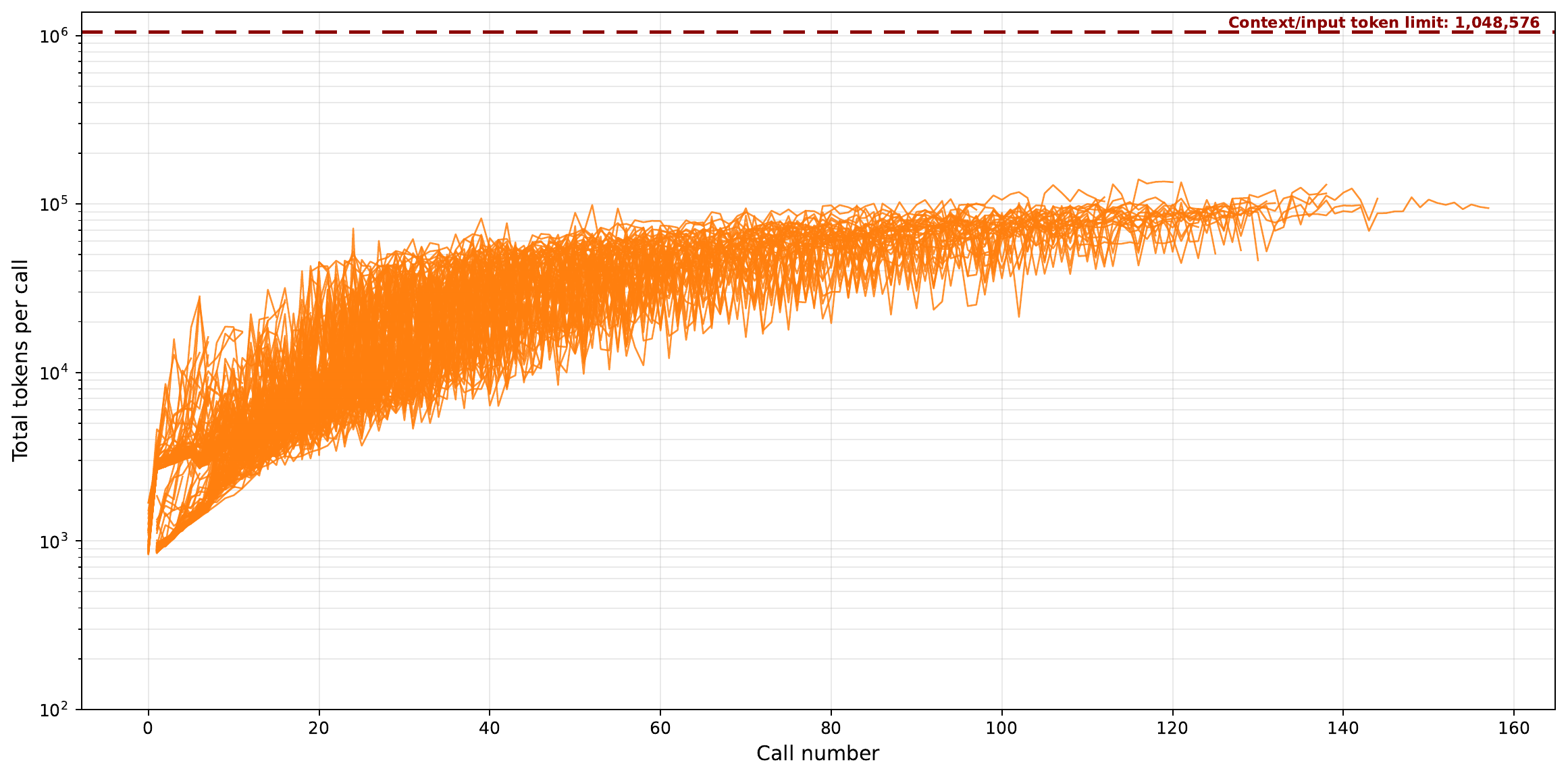}
    \caption{Gemini 3.1 Pro Preview (high)}
\end{subfigure}
\hfill
\begin{subfigure}[t]{0.32\textwidth}
    \centering
    \includegraphics[
        width=\linewidth,
        keepaspectratio
    ]{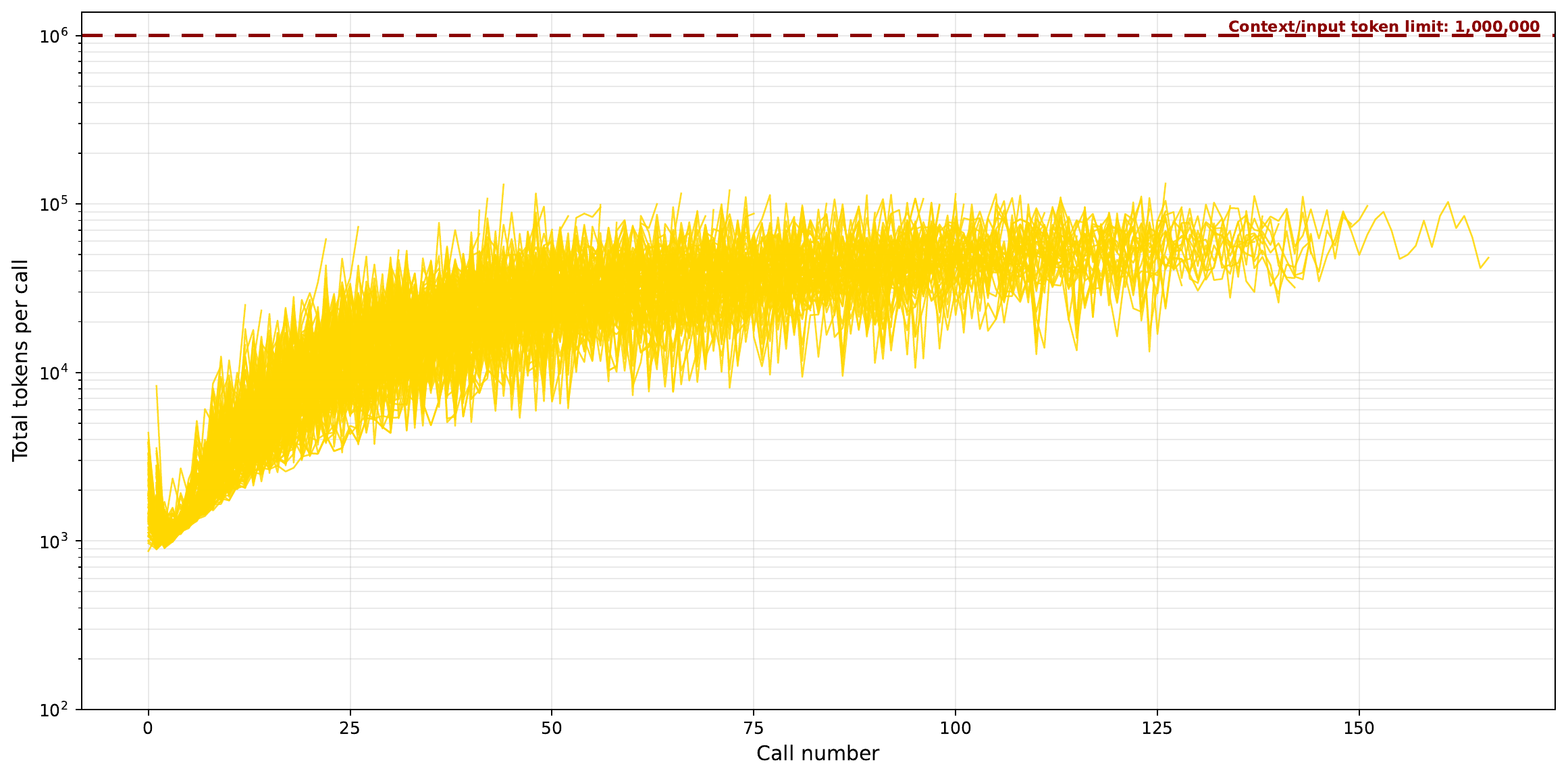}
    \caption{DeepSeek-v4-Pro (high)}
\end{subfigure}
\hfill
\begin{subfigure}[t]{0.32\textwidth}
    \centering
    \includegraphics[
        width=\linewidth,
        keepaspectratio
    ]{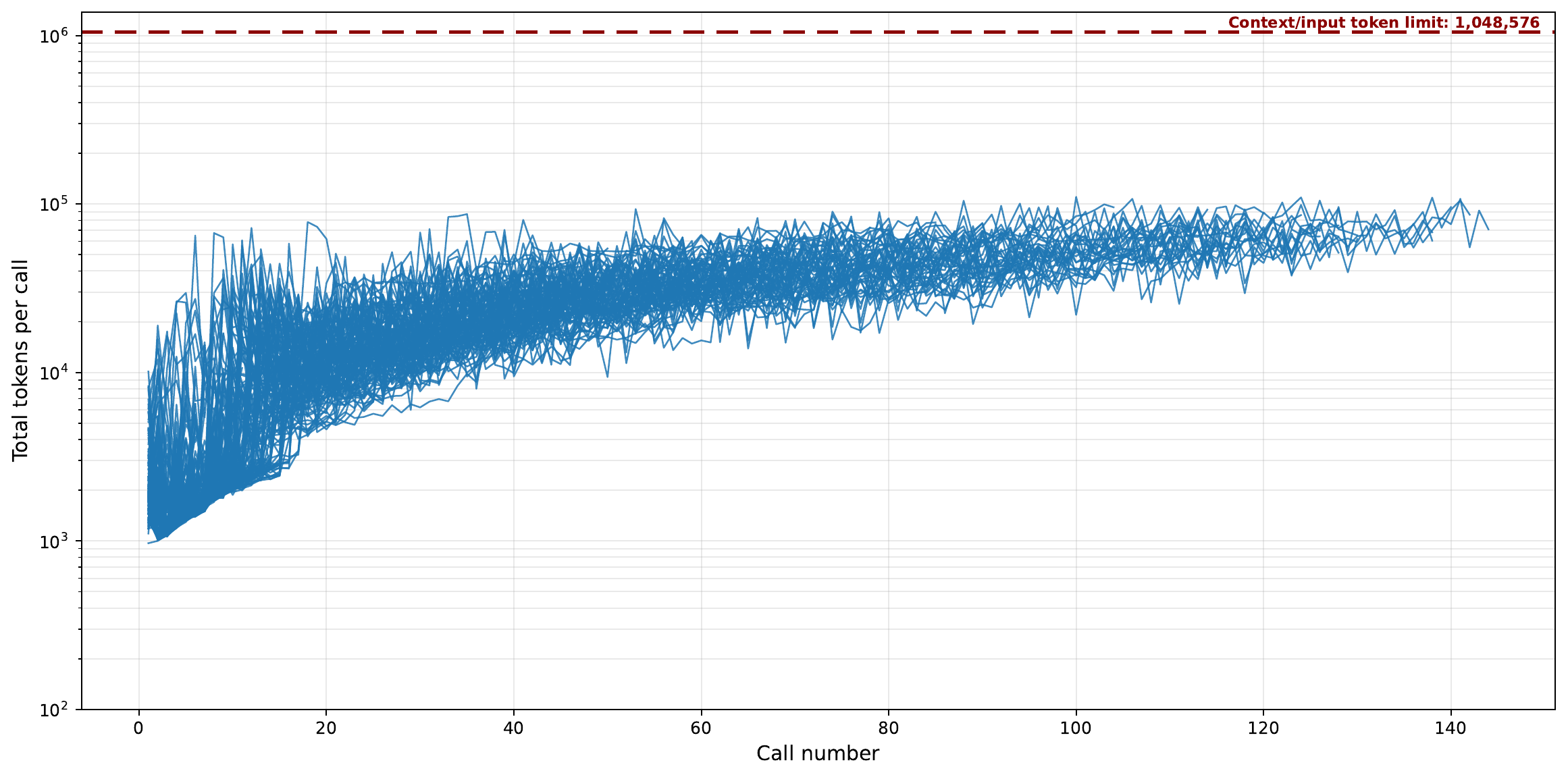}
    \caption{Gemini 3.8 Flash (high)}
\end{subfigure}

\vspace{0.3em}

\caption{
Each line represents a single run of one task instance. The x-axis shows the tool-call number, while the y-axis shows the total number of tokens in the context window at each call on a base-10 logarithmic scale. Context-window usage throughout the interaction is shown separately for the Stateful and Stateless runs of each of the three models. The red dashed line marks the model-specific context-window limit. Since no trajectory crosses this line, the observed failures cannot be attributed to exceeding the available context window.
}
\label{fig:context_window_analysis}
\end{figure}

%% file: latex/figures/cost_runtime_by_states_figure.tex
\clearpage
\section{Computational Cost, Runtime, and Token Usage}
\label{app:cost_runtime}

% Auto-generated from results.csv and the corresponding vector graph output.
% This file draws the three paired-state scatter plots directly with PGFPlots.
% It does NOT include pre-rendered graph PDFs.
%
% Required packages in the parent document:
%   \usepackage{graphicx}
%   \usepackage{tikz}
%   \usepackage{pgfplots}
%   \pgfplotsset{compat=1.18}
%   \usepackage{subcaption}
%
% Data represented here: 3 models x 2 interaction modes x 160 runs = 960 runs per panel.
% Paired-state groups: 2--3, 4--5, 6--7, 8--9.

\definecolor{figtaborange}{HTML}{FF7F0E}
\definecolor{figtabblue}{HTML}{1F77B4}
\definecolor{figgold}{HTML}{FFD700}

\begin{center}
\centering
\resizebox{0.99\textwidth}{!}{%
\begin{minipage}{\textwidth}
\centering

\makebox[\textwidth][c]{%
\begin{minipage}{0.99\textwidth}
\centering
\begin{tabular}{ccc}
\tikz[baseline=-0.5ex]{\draw[draw=black, fill=figtaborange] (0,0) rectangle (0.24,0.16);}~\scriptsize Gemini 3.1 Pro Preview (high) &
\hspace{1.1em}\tikz[baseline=-0.5ex]{\draw[draw=black, fill=figgold] (0,0) rectangle (0.24,0.16);}~\scriptsize DeepSeek-v4-Pro (high) &
\hspace{1.1em}\tikz[baseline=-0.5ex]{\draw[draw=black, fill=figtabblue] (0,0) rectangle (0.24,0.16);}~\scriptsize Gemini 3.8 Flash (high)
\end{tabular}
\end{minipage}%
}

\vspace{0.55em}
\begin{minipage}[t]{0.98\textwidth}
\centering
\begin{tikzpicture}
\begin{axis}[
    width=\linewidth,
    height=5.25cm,
    xmin=-1,
    xmax=55,
    ymin=0.55,
    ymax=4.45,
    y dir=reverse,
    xlabel={Game cost (USD)},
    ylabel={Minimal DFA States},
    ytick={1,2,3,4},
    yticklabels={2--3,4--5,6--7,8--9},
    xtick={0,10,20,30,40,50},
    xticklabel={\$\pgfmathprintnumber[fixed,precision=2]{\tick}},
    grid=major,
    grid style={gray!18},
    tick label style={font=\scriptsize},
    label style={font=\small},
    clip=true,
    legend style={
        at={(0.97,0.97)},
        anchor=north east,
        draw=black!25,
        fill=white,
        fill opacity=0.90,
        text opacity=1,
        font=\scriptsize,
        cells={anchor=west},
        row sep=-1pt,
    },
]

\addlegendimage{only marks, mark=triangle*, mark size=1.5pt, mark options={draw=black,fill=black}}
\addlegendentry{Stateful}
\addlegendimage{only marks, mark=*, mark size=1.5pt, mark options={draw=black,fill=black}}
\addlegendentry{Stateless}

\addplot+[only marks, mark=triangle*, mark size=1.35pt, opacity=0.90,
    mark options={draw=black, fill=figtaborange}] coordinates {
(0.308327,0.775) (0.229331,0.775) (0.218963,0.775) (0.294009,0.775) (0.148114,0.775) (0.260682,0.775) (0.250807,0.775) (0.320423,0.775)
(0.290800,0.775) (0.332272,0.775) (0.329803,0.775) (0.348565,0.775) (0.433978,0.775) (0.561110,0.775) (0.529265,0.775) (0.185884,0.775)
(0.047149,0.775) (0.674418,0.775) (0.107629,0.775) (0.046162,0.775) (0.049371,0.775) (0.481868,0.775) (0.265125,0.775) (0.172553,0.775)
(0.306105,0.775) (0.276729,0.775) (0.278704,0.775) (0.215754,0.775) (0.264632,0.775) (0.456195,0.775) (0.289318,0.775) (0.637636,0.775)
(0.653682,0.775) (0.547286,0.775) (1.170357,0.775) (0.250561,0.775) (0.956578,0.775) (1.167148,0.775) (1.087166,0.775) (0.250067,0.775)
(0.392752,1.775) (0.550248,1.775) (0.800810,1.775) (0.553210,1.775) (0.999778,1.775) (0.972130,1.775) (0.833642,1.775) (0.663309,1.775)
(0.744032,1.775) (1.550273,1.775) (1.279222,1.775) (3.023032,1.775) (3.142018,1.775) (3.312844,1.775) (3.509344,1.775) (3.846799,1.775)
(5.893011,1.775) (6.649633,1.775) (3.597719,1.775) (14.396554,1.775) (0.281666,1.775) (0.415216,1.775) (0.418426,1.775) (0.489768,1.775)
(0.373250,1.775) (0.569997,1.775) (0.518897,1.775) (0.684292,1.775) (0.578637,1.775) (0.770693,1.775) (0.728233,1.775) (0.991631,1.775)
(1.343899,1.775) (1.388827,1.775) (2.168654,1.775) (1.506579,1.775) (2.989459,1.775) (3.106964,1.775) (4.439014,1.775) (6.174184,1.775)
(0.969167,2.775) (0.934113,2.775) (1.506579,2.775) (1.597917,2.775) (0.981510,2.775) (1.326372,2.775) (2.441680,2.775) (1.279469,2.775)
(1.636920,2.775) (2.877138,2.775) (3.344689,2.775) (4.250167,2.775) (0.658125,2.775) (0.756375,2.775) (0.801797,2.775) (0.699598,2.775)
(0.935101,2.775) (0.682811,2.775) (0.831914,2.775) (0.871164,2.775) (0.709719,2.775) (5.477548,2.775) (1.444370,2.775) (1.569034,2.775)
(0.872399,2.775) (5.643190,2.775) (5.081834,2.775) (1.983510,2.775) (2.202227,2.775) (2.167666,2.775) (8.141154,2.775) (9.145869,2.775)
(4.946555,2.775) (8.972080,2.775) (11.626552,2.775) (6.554099,2.775) (10.700832,2.775) (5.311659,2.775) (7.211484,2.775) (10.680096,2.775)
(1.162458,3.775) (1.602607,3.775) (1.801822,3.775) (1.479424,3.775) (1.434496,3.775) (1.565084,3.775) (1.563109,3.775) (1.406354,3.775)
(2.349354,3.775) (1.907231,3.775) (2.213582,3.775) (2.854427,3.775) (3.526624,3.775) (3.614259,3.775) (3.284455,3.775) (3.461453,3.775)
(4.506900,3.775) (2.101015,3.775) (4.987534,3.775) (4.872991,3.775) (7.058185,3.775) (6.157891,3.775) (4.973710,3.775) (8.134735,3.775)
(8.153250,3.775) (8.213730,3.775) (9.082920,3.775) (7.762226,3.775) (9.186107,3.775) (7.652127,3.775) (10.283887,3.775) (13.480708,3.775)
(9.822755,3.775) (14.134143,3.775) (14.761164,3.775) (17.264558,3.775) (18.626972,3.775) (19.447283,3.775) (20.831173,3.775) (11.312795,3.775)
};

\addplot+[only marks, mark=triangle*, mark size=1.35pt, opacity=0.90,
    mark options={draw=black, fill=figgold}] coordinates {
(0.017773,0.865) (0.026166,0.865) (0.035547,0.865) (0.028388,0.865) (0.053074,0.865) (0.053814,0.865) (0.063442,0.865) (0.058505,0.865)
(0.059986,0.865) (0.089115,0.865) (0.094793,0.865) (0.025672,0.865) (0.032338,0.865) (0.107876,0.865) (0.118244,0.865) (0.088621,0.865)
(0.134043,0.865) (0.250067,0.865) (0.008145,0.865) (0.012589,0.865) (0.011108,0.865) (0.008145,0.865) (0.009380,0.865) (0.012342,0.865)
(0.015057,0.865) (0.015304,0.865) (0.035053,0.865) (0.006417,0.865) (0.007405,0.865) (0.010861,0.865) (0.029128,0.865) (0.007652,0.865)
(0.008145,0.865) (0.006417,0.865) (0.012342,0.865) (0.023451,0.865) (0.037522,0.865) (0.058505,0.865) (0.275740,0.865) (0.252042,0.865)
(0.093559,1.865) (0.176504,1.865) (0.177738,1.865) (0.203904,1.865) (0.243155,1.865) (0.265125,1.865) (0.250807,1.865) (0.283888,1.865)
(0.293515,1.865) (0.299686,1.865) (0.308573,1.865) (0.373497,1.865) (0.035548,1.865) (0.376460,1.865) (0.073069,1.865) (0.417685,1.865)
(0.093806,1.865) (0.113555,1.865) (0.114789,1.865) (0.184896,1.865) (0.542842,1.865) (0.193043,1.865) (0.551236,1.865) (0.226616,1.865)
(0.131822,1.865) (0.215508,1.865) (0.235750,1.865) (0.268828,1.865) (0.270063,1.865) (0.619369,1.865) (0.187612,1.865) (0.294009,1.865)
(0.676146,1.865) (0.361401,1.865) (0.556173,1.865) (0.506060,1.865) (0.632205,1.865) (0.241922,1.865) (0.557160,1.865) (0.618875,1.865)
(0.181934,2.865) (0.228097,2.865) (0.223407,2.865) (0.261422,2.865) (0.261176,2.865) (0.339924,2.865) (0.332519,2.865) (0.430028,2.865)
(0.451011,2.865) (0.478412,2.865) (0.513219,2.865) (0.214767,2.865) (0.240687,2.865) (0.583574,2.865) (0.274012,2.865) (0.595670,2.865)
(0.333506,2.865) (0.307586,2.865) (0.657385,2.865) (0.394480,2.865) (0.395961,2.865) (0.395715,2.865) (0.425338,2.865) (0.465082,2.865)
(0.478906,2.865) (0.495445,2.865) (0.501370,2.865) (0.894863,2.865) (0.599126,2.865) (0.641092,2.865) (0.680343,2.865) (1.164926,2.865)
(0.812659,2.865) (1.242440,2.865) (1.164433,2.865) (0.922264,2.865) (0.971636,2.865) (1.770718,2.865) (1.903528,2.865) (1.758622,2.865)
(0.166877,3.865) (0.573700,3.865) (0.732677,3.865) (0.835864,3.865) (1.168135,3.865) (1.202943,3.865) (1.229850,3.865) (1.357970,3.865)
(1.517687,3.865) (1.440421,3.865) (0.530006,3.865) (0.545311,3.865) (0.616900,3.865) (0.988669,3.865) (1.103459,3.865) (1.084450,3.865)
(1.085438,3.865) (1.370807,3.865) (1.355748,3.865) (1.355995,3.865) (0.319188,3.865) (0.688242,3.865) (0.765509,3.865) (0.953862,3.865)
(1.114567,3.865) (1.447333,3.865) (1.473747,3.865) (1.381175,3.865) (1.691723,3.865) (1.595201,3.865) (0.494458,3.865) (0.393246,3.865)
(0.691945,3.865) (0.758844,3.865) (1.204671,3.865) (1.044953,3.865) (1.215779,3.865) (1.341924,3.865) (1.222691,3.865) (1.435977,3.865)
};

\addplot+[only marks, mark=triangle*, mark size=1.35pt, opacity=0.90,
    mark options={draw=black, fill=figtabblue}] coordinates {
(0.138734,0.955) (0.186871,0.955) (0.306105,0.955) (0.130834,0.955) (0.116516,0.955) (0.085166,0.955) (0.149596,0.955) (0.084919,0.955)
(0.177491,0.955) (0.353008,0.955) (0.339924,0.955) (0.477178,0.955) (0.097509,0.955) (0.545311,0.955) (0.155768,0.955) (0.172061,0.955)
(0.642820,0.955) (0.608260,0.955) (0.258954,0.955) (0.275740,0.955) (0.395221,0.955) (0.239700,0.955) (0.261916,0.955) (0.343134,0.955)
(0.145647,0.955) (0.429781,0.955) (0.184156,0.955) (0.267841,0.955) (0.190574,0.955) (0.244636,0.955) (0.765262,0.955) (0.711447,0.955)
(0.556913,0.955) (0.514207,0.955) (0.716384,0.955) (0.184156,0.955) (0.421882,0.955) (0.443111,0.955) (0.462613,0.955) (0.647264,0.955)
(0.684045,1.955) (0.540621,1.955) (0.756869,1.955) (0.817349,1.955) (0.714162,1.955) (0.882026,1.955) (1.376978,1.955) (1.242687,1.955)
(1.341183,1.955) (1.490286,1.955) (1.063961,1.955) (1.171345,1.955) (1.658150,1.955) (1.969686,1.955) (2.059789,1.955) (1.848478,1.955)
(1.854156,1.955) (2.071145,1.955) (2.360957,1.955) (0.264385,1.955) (0.752919,1.955) (0.685033,1.955) (0.649979,1.955) (0.743538,1.955)
(0.923498,1.955) (0.319682,1.955) (0.361895,1.955) (1.246143,1.955) (0.597892,1.955) (1.019773,1.955) (0.602335,1.955) (0.773408,1.955)
(0.596658,1.955) (0.708731,1.955) (0.894863,1.955) (0.774396,1.955) (0.877583,1.955) (1.107161,1.955) (1.279222,1.955) (1.677405,1.955)
(0.890173,2.955) (1.246143,2.955) (1.623590,2.955) (1.691970,2.955) (1.952159,2.955) (2.085216,2.955) (1.830211,2.955) (2.026216,2.955)
(1.797625,2.955) (1.989928,2.955) (2.442914,2.955) (2.351082,2.955) (3.667086,2.955) (3.608581,2.955) (3.289639,2.955) (3.370609,2.955)
(3.561184,2.955) (5.336839,2.955) (1.586808,2.955) (2.834679,2.955) (0.581352,2.955) (0.811671,2.955) (1.083463,2.955) (1.215779,2.955)
(1.058037,2.955) (2.015108,2.955) (2.190624,2.955) (1.825273,2.955) (2.333802,2.955) (2.015108,2.955) (2.398726,2.955) (2.154583,2.955)
(2.417734,2.955) (2.754696,2.955) (2.420203,2.955) (2.984522,2.955) (0.867708,2.955) (3.827791,2.955) (3.955417,2.955) (3.177565,2.955)
(1.296255,3.955) (1.952653,3.955) (2.054605,3.955) (1.936854,3.955) (2.081513,3.955) (2.383174,3.955) (2.622380,3.955) (2.520921,3.955)
(2.648547,3.955) (2.312079,3.955) (2.763090,3.955) (2.534992,3.955) (2.788516,3.955) (3.436027,3.955) (3.607840,3.955) (2.844306,3.955)
(3.203979,3.955) (3.558715,3.955) (3.881113,3.955) (3.259276,3.955) (3.844084,3.955) (3.944062,3.955) (4.267940,3.955) (3.328643,3.955)
(3.971216,3.955) (4.532079,3.955) (4.590091,3.955) (3.259276,3.955) (3.549088,3.955) (4.930756,3.955) (4.649831,3.955) (5.855489,3.955)
(4.877681,3.955) (5.432126,3.955) (6.356118,3.955) (5.444469,3.955) (3.020810,3.955) (3.446148,3.955) (3.116098,3.955) (5.209213,3.955)
};

\addplot+[only marks, mark=*, mark size=1.35pt, opacity=0.90,
    mark options={draw=black, fill=figtaborange}] coordinates {
(0.042707,1.045) (0.048137,1.045) (0.044188,1.045) (0.166630,1.045) (0.080229,1.045) (0.049865,1.045) (0.248339,1.045) (0.242168,1.045)
(0.242908,1.045) (0.365104,1.045) (0.353502,1.045) (0.319435,1.045) (0.412995,1.045) (0.461379,1.045) (0.453479,1.045) (0.656644,1.045)
(0.834136,1.045) (1.005702,1.045) (0.649979,1.045) (1.231578,1.045) (0.098497,1.045) (0.400405,1.045) (0.220445,1.045) (0.230566,1.045)
(0.038510,1.045) (0.556420,1.045) (0.294256,1.045) (0.371276,1.045) (0.446567,1.045) (1.041991,1.045) (0.230072,1.045) (0.362389,1.045)
(0.738108,1.045) (1.468809,1.045) (0.232294,1.045) (0.636402,1.045) (0.695895,1.045) (0.354983,1.045) (0.705522,1.045) (0.294749,1.045)
(0.107877,2.045) (0.466563,2.045) (0.680590,2.045) (0.638377,2.045) (0.766003,2.045) (0.743538,2.045) (0.730702,2.045) (0.995334,2.045)
(1.543114,2.045) (1.612481,2.045) (2.505863,2.045) (2.576218,2.045) (3.798909,2.045) (4.349898,2.045) (7.787652,2.045) (8.680787,2.045)
(10.167370,2.045) (12.527340,2.045) (17.006838,2.045) (16.072478,2.045) (1.267373,2.045) (0.739342,2.045) (1.325138,2.045) (1.031129,2.045)
(0.507542,2.045) (1.986719,2.045) (1.035325,2.045) (1.075563,2.045) (1.791701,2.045) (3.540942,2.045) (3.383693,2.045) (11.248611,2.045)
(19.466785,2.045) (11.881310,2.045) (0.981263,2.045) (4.662668,2.045) (2.052877,2.045) (1.891185,2.045) (14.184255,2.045) (4.750796,2.045)
(1.241206,3.045) (1.761584,3.045) (2.005233,3.045) (3.620183,3.045) (4.180552,3.045) (7.257646,3.045) (1.630749,3.045) (2.850231,3.045)
(3.174109,3.045) (3.072651,3.045) (4.072675,3.045) (5.041102,3.045) (7.495618,3.045) (7.362068,3.045) (20.368560,3.045) (2.982300,3.045)
(3.286430,3.045) (5.697006,3.045) (21.678146,3.045) (17.860722,3.045) (3.262485,3.045) (3.917648,3.045) (7.195932,3.045) (2.780370,3.045)
(7.949591,3.045) (8.344319,3.045) (2.307142,3.045) (13.580686,3.045) (2.347133,3.045) (20.002715,3.045) (25.411143,3.045) (27.377620,3.045)
(5.640228,3.045) (20.696142,3.045) (8.471944,3.045) (20.680343,3.045) (1.255770,3.045) (6.817004,3.045) (18.366289,3.045) (7.809129,3.045)
(4.308178,4.045) (4.452344,4.045) (8.406774,4.045) (11.122960,4.045) (11.683823,4.045) (15.144289,4.045) (16.645684,4.045) (21.481399,4.045)
(20.656644,4.045) (27.965390,4.045) (28.528228,4.045) (34.447900,4.045) (31.646794,4.045) (32.612506,4.045) (33.460959,4.045) (33.930731,4.045)
(34.641684,4.045) (41.153571,4.045) (36.520526,4.045) (41.625811,4.045) (26.370683,4.045) (28.346787,4.045) (33.877904,4.045) (41.030882,4.045)
(14.419759,4.045) (4.713274,4.045) (26.934261,4.045) (30.203658,4.045) (7.221852,4.045) (33.452072,4.045) (50.163173,4.045) (38.814337,4.045)
(30.772421,4.045) (30.304870,4.045) (30.987188,4.045) (24.229430,4.045) (34.680194,4.045) (19.630946,4.045) (36.484485,4.045) (32.988718,4.045)
};

\addplot+[only marks, mark=*, mark size=1.35pt, opacity=0.90,
    mark options={draw=black, fill=figgold}] coordinates {
(0.002715,1.135) (0.003949,1.135) (0.010367,1.135) (0.003455,1.135) (0.007652,1.135) (0.006911,1.135) (0.011601,1.135) (0.005924,1.135)
(0.010614,1.135) (0.019502,1.135) (0.020983,1.135) (0.040732,1.135) (0.027648,1.135) (0.039991,1.135) (0.038263,1.135) (0.045175,1.135)
(0.069860,1.135) (0.073069,1.135) (0.169592,1.135) (0.177245,1.135) (0.003949,1.135) (0.017774,1.135) (0.004443,1.135) (0.004443,1.135)
(0.006664,1.135) (0.011355,1.135) (0.009133,1.135) (0.006664,1.135) (0.019008,1.135) (0.023205,1.135) (0.028636,1.135) (0.053814,1.135)
(0.051100,1.135) (0.137746,1.135) (0.061467,1.135) (0.042953,1.135) (0.206868,1.135) (0.059245,1.135) (0.094053,1.135) (0.063689,1.135)
(0.047644,2.135) (0.116024,2.135) (0.170579,2.135) (0.182921,2.135) (0.223654,2.135) (0.303142,2.135) (0.326347,2.135) (0.408304,2.135)
(0.475697,2.135) (0.540374,2.135) (0.584808,2.135) (0.464588,2.135) (0.591227,2.135) (0.666765,2.135) (0.738848,2.135) (0.788467,2.135)
(0.851416,2.135) (0.900541,2.135) (0.889926,2.135) (1.114567,2.135) (0.332272,2.135) (0.417685,2.135) (0.286850,2.135) (0.394480,2.135)
(0.227851,2.135) (0.651213,2.135) (0.189587,2.135) (0.446074,2.135) (0.215261,2.135) (0.269816,2.135) (0.777852,2.135) (0.938804,2.135)
(0.815128,2.135) (0.713175,2.135) (0.412501,2.135) (1.099262,2.135) (0.280678,2.135) (0.556173,2.135) (0.881779,2.135) (0.514454,2.135)
(0.332519,3.135) (0.419660,3.135) (0.540867,3.135) (1.014096,3.135) (1.009652,3.135) (1.021995,3.135) (1.309586,3.135) (1.058037,3.135)
(1.375003,3.135) (1.425609,3.135) (0.811671,3.135) (0.783776,3.135) (1.287862,3.135) (1.273297,3.135) (1.439927,3.135) (1.450295,3.135)
(1.565331,3.135) (1.616678,3.135) (1.825767,3.135) (1.875879,3.135) (1.073835,3.135) (0.933620,3.135) (0.933126,3.135) (0.401145,3.135)
(0.990397,3.135) (1.611247,3.135) (1.880817,3.135) (1.020514,3.135) (1.122960,3.135) (1.114320,3.135) (2.149399,3.135) (0.982004,3.135)
(1.845022,3.135) (1.771211,3.135) (1.786023,3.135) (2.152114,3.135) (0.598386,3.135) (1.905996,3.135) (0.449036,3.135) (3.619196,3.135)
(1.180972,4.135) (1.805772,4.135) (1.639142,4.135) (1.706041,4.135) (1.982522,4.135) (2.263448,4.135) (2.128416,4.135) (2.365894,4.135)
(2.306648,4.135) (2.493520,4.135) (2.327631,4.135) (2.480930,4.135) (2.814930,4.135) (2.886272,4.135) (2.918117,4.135) (2.925523,4.135)
(2.958108,4.135) (3.016367,4.135) (2.719396,4.135) (3.154361,4.135) (3.104742,4.135) (1.891432,4.135) (2.542398,4.135) (2.349601,4.135)
(2.115826,4.135) (1.619887,4.135) (1.707769,4.135) (1.998321,4.135) (2.295539,4.135) (3.035128,4.135) (4.006270,4.135) (2.395764,4.135)
(2.395764,4.135) (2.697425,4.135) (3.467378,4.135) (4.295835,4.135) (3.466884,4.135) (1.806018,4.135) (3.057839,4.135) (2.703597,4.135)
};

\addplot+[only marks, mark=*, mark size=1.35pt, opacity=0.90,
    mark options={draw=black, fill=figtabblue}] coordinates {
(0.101953,1.225) (0.058258,1.225) (0.098497,1.225) (0.179466,1.225) (0.050359,1.225) (0.215755,1.225) (0.095041,1.225) (0.044188,1.225)
(0.262657,1.225) (0.074797,1.225) (0.041225,1.225) (0.084179,1.225) (0.211558,1.225) (0.342146,1.225) (0.067885,1.225) (0.107384,1.225)
(0.269569,1.225) (0.187365,1.225) (0.454714,1.225) (0.476437,1.225) (0.435706,1.225) (0.247845,1.225) (0.413488,1.225) (0.493471,1.225)
(0.559876,1.225) (0.441383,1.225) (0.416451,1.225) (0.028389,1.225) (0.037029,1.225) (0.591474,1.225) (0.023698,1.225) (0.048631,1.225)
(0.487793,1.225) (0.072576,1.225) (0.376706,1.225) (0.056036,1.225) (0.105902,1.225) (0.142684,1.225) (0.344862,1.225) (0.742551,1.225)
(0.404848,2.225) (0.715890,2.225) (1.015330,2.225) (0.527290,2.225) (1.175294,2.225) (1.132588,2.225) (1.277988,2.225) (0.531487,2.225)
(1.298477,2.225) (1.391049,2.225) (1.387346,2.225) (1.573477,2.225) (1.189612,2.225) (1.853415,2.225) (1.879829,2.225) (1.816386,2.225)
(1.427584,2.225) (1.986472,2.225) (1.528302,2.225) (2.079785,2.225) (0.104914,2.225) (0.293268,2.225) (0.431262,2.225) (0.659853,2.225)
(0.537658,2.225) (0.562838,2.225) (0.595670,2.225) (0.665778,2.225) (0.686267,2.225) (0.741070,2.225) (0.873139,2.225) (0.940532,2.225)
(1.203683,2.225) (1.174307,2.225) (1.347355,2.225) (1.585327,2.225) (1.436718,2.225) (1.429559,2.225) (1.592239,2.225) (1.478684,2.225)
(1.108396,3.225) (1.384631,3.225) (1.615443,3.225) (1.586561,3.225) (1.634452,3.225) (2.368856,3.225) (2.415266,3.225) (2.396504,3.225)
(2.768274,3.225) (2.683601,3.225) (2.719149,3.225) (2.884544,3.225) (0.809697,3.225) (3.187440,3.225) (3.194105,3.225) (3.222741,3.225)
(3.142265,3.225) (1.011874,3.225) (1.067664,3.225) (3.390605,3.225) (3.688810,3.225) (1.417956,3.225) (1.483868,3.225) (3.821126,3.225)
(1.616431,3.225) (1.723321,3.225) (1.769236,3.225) (2.380706,3.225) (2.295539,3.225) (2.223950,3.225) (2.303439,3.225) (2.385643,3.225)
(2.541411,3.225) (2.616949,3.225) (2.658422,3.225) (2.917130,3.225) (2.600163,3.225) (5.039374,3.225) (3.022785,3.225) (2.966995,3.225)
(2.831963,4.225) (3.505888,4.225) (4.306944,4.225) (4.445926,4.225) (4.524180,4.225) (3.932953,4.225) (4.815967,4.225) (4.713027,4.225)
(4.670814,4.225) (4.596263,4.225) (4.565158,4.225) (5.139105,4.225) (4.770545,4.225) (5.801674,4.225) (5.376336,4.225) (5.779703,4.225)
(6.023106,4.225) (6.438322,4.225) (6.350194,4.225) (2.464884,4.225) (2.169641,4.225) (2.346145,4.225) (7.983905,4.225) (2.462169,4.225)
(2.991681,4.225) (3.099558,4.225) (3.049939,4.225) (3.538720,4.225) (3.480461,4.225) (3.655237,4.225) (3.685848,4.225) (3.994174,4.225)
(4.350144,4.225) (4.630576,4.225) (4.505172,4.225) (5.017650,4.225) (5.580735,4.225) (6.430176,4.225) (6.883655,4.225) (7.983905,4.225)
};

\end{axis}
\end{tikzpicture}
\end{minipage}

\par\vspace{0.8em}

\begin{minipage}[t]{0.98\textwidth}
\centering
\begin{tikzpicture}
\begin{axis}[
    width=\linewidth,
    height=5.25cm,
    xmin=-0.5,
    xmax=20.5,
    ymin=0.55,
    ymax=4.45,
    y dir=reverse,
    xlabel={Total runtime (hours)},
    ylabel={Minimal DFA States},
    ytick={1,2,3,4},
    yticklabels={2--3,4--5,6--7,8--9},
    xtick={0,4,8,12,16,20},
    grid=major,
    grid style={gray!18},
    tick label style={font=\scriptsize},
    label style={font=\small},
    clip=true,
    legend style={
        at={(0.97,0.97)},
        anchor=north east,
        draw=black!25,
        fill=white,
        fill opacity=0.90,
        text opacity=1,
        font=\scriptsize,
        cells={anchor=west},
        row sep=-1pt,
    },
]

\addlegendimage{only marks, mark=triangle*, mark size=1.5pt, mark options={draw=black,fill=black}}
\addlegendentry{Stateful}
\addlegendimage{only marks, mark=*, mark size=1.5pt, mark options={draw=black,fill=black}}
\addlegendentry{Stateless}

\addplot+[only marks, mark=triangle*, mark size=1.35pt, opacity=0.90,
    mark options={draw=black, fill=figtaborange}] coordinates {
(0.065530,0.775) (0.043600,0.775) (0.041360,0.775) (0.054040,0.775) (0.030986,0.775) (0.051049,0.775) (0.052635,0.775) (0.061916,0.775)
(0.056601,0.775) (0.067375,0.775) (0.064984,0.775) (0.067180,0.775) (0.086690,0.775) (0.109024,0.775) (0.099496,0.775) (0.038143,0.775)
(0.012081,0.775) (0.138210,0.775) (0.023172,0.775) (0.010976,0.775) (0.011801,0.775) (0.149430,0.775) (0.054573,0.775) (0.035301,0.775)
(0.059931,0.775) (0.057372,0.775) (0.055612,0.775) (0.045455,0.775) (0.056484,0.775) (0.090380,0.775) (0.061085,0.775) (0.123623,0.775)
(0.219876,0.775) (0.099892,0.775) (0.273403,0.775) (0.168460,0.775) (0.184050,0.775) (0.214059,0.775) (0.202296,0.775) (0.054548,0.775)
(0.086995,1.775) (0.124483,1.775) (0.151097,1.775) (0.176879,1.775) (0.274462,1.775) (0.279223,1.775) (0.287217,1.775) (0.267848,1.775)
(0.324026,1.775) (0.347064,1.775) (0.364864,1.775) (0.731945,1.775) (0.771956,1.775) (0.874560,1.775) (1.072935,1.775) (1.227979,1.775)
(1.689261,1.775) (1.680258,1.775) (1.909736,1.775) (3.373407,1.775) (0.056908,1.775) (0.080477,1.775) (0.083031,1.775) (0.098552,1.775)
(0.075555,1.775) (0.110101,1.775) (0.103257,1.775) (0.133249,1.775) (0.115489,1.775) (0.146914,1.775) (0.162963,1.775) (0.205897,1.775)
(0.222235,1.775) (0.263374,1.775) (0.331948,1.775) (0.337591,1.775) (0.645106,1.775) (0.680453,1.775) (0.889643,1.775) (1.247723,1.775)
(0.228677,2.775) (0.210188,2.775) (0.307922,2.775) (0.348221,2.775) (0.418104,2.775) (0.482657,2.775) (0.505824,2.775) (0.564531,2.775)
(0.602970,2.775) (0.676968,2.775) (0.794601,2.775) (0.894183,2.775) (0.144188,2.775) (0.161334,2.775) (0.181889,2.775) (0.179677,2.775)
(0.184770,2.775) (0.166633,2.775) (0.175103,2.775) (0.219795,2.775) (0.203770,2.775) (1.129377,2.775) (0.304625,2.775) (0.317174,2.775)
(0.276762,2.775) (1.305694,2.775) (1.341494,2.775) (0.523510,2.775) (0.527417,2.775) (0.546951,2.775) (1.747809,2.775) (1.842449,2.775)
(0.959552,2.775) (1.844744,2.775) (2.132637,2.775) (1.403656,2.775) (2.261738,2.775) (1.463271,2.775) (1.662473,2.775) (2.200091,2.775)
(0.233670,3.775) (0.291169,3.775) (0.338287,3.775) (0.345423,3.775) (0.286898,3.775) (0.310000,3.775) (0.295239,3.775) (0.292979,3.775)
(0.425936,3.775) (0.411348,3.775) (0.454650,3.775) (0.538103,3.775) (0.814296,3.775) (0.770734,3.775) (0.644493,3.775) (0.838044,3.775)
(0.943069,3.775) (0.870377,3.775) (1.064493,3.775) (1.085827,3.775) (1.354865,3.775) (1.283398,3.775) (1.377804,3.775) (1.519294,3.775)
(1.603497,3.775) (1.852631,3.775) (1.904816,3.775) (1.948768,3.775) (1.890693,3.775) (1.934725,3.775) (2.234522,3.775) (2.382680,3.775)
(2.391482,3.775) (2.616840,3.775) (3.494929,3.775) (3.563441,3.775) (3.716494,3.775) (4.085836,3.775) (5.151878,3.775) (3.362344,3.775)
};

\addplot+[only marks, mark=triangle*, mark size=1.35pt, opacity=0.90,
    mark options={draw=black, fill=figgold}] coordinates {
(0.054090,0.865) (0.078188,0.865) (0.103966,0.865) (0.082610,0.865) (0.161232,0.865) (0.154771,0.865) (0.177963,0.865) (0.163092,0.865)
(0.166582,0.865) (0.254870,0.865) (0.266198,0.865) (0.088929,0.865) (0.092486,0.865) (0.308393,0.865) (0.362322,0.865) (0.268076,0.865)
(0.391832,0.865) (0.657180,0.865) (0.026484,0.865) (0.039054,0.865) (0.034851,0.865) (0.027448,0.865) (0.029828,0.865) (0.040107,0.865)
(0.056731,0.865) (0.048802,0.865) (0.113109,0.865) (0.021536,0.865) (0.024414,0.865) (0.032789,0.865) (0.084739,0.865) (0.026210,0.865)
(0.026577,0.865) (0.022460,0.865) (0.040197,0.865) (0.074517,0.865) (0.118415,0.865) (0.194867,0.865) (0.964532,0.865) (0.715967,0.865)
(0.272011,1.865) (0.519936,1.865) (0.517127,1.865) (0.599828,1.865) (0.664048,1.865) (0.751986,1.865) (0.814011,1.865) (0.816558,1.865)
(0.866332,1.865) (0.835322,1.865) (0.840062,1.865) (1.053961,1.865) (0.098239,1.865) (1.064716,1.865) (0.182202,1.865) (1.161867,1.865)
(0.249281,1.865) (0.289299,1.865) (0.303500,1.865) (0.475619,1.865) (1.445441,1.865) (0.522023,1.865) (1.475559,1.865) (0.603246,1.865)
(0.346744,1.865) (0.588462,1.865) (0.613607,1.865) (0.700101,1.865) (0.711829,1.865) (1.686572,1.865) (0.491419,1.865) (0.761644,1.865)
(1.808905,1.865) (0.976828,1.865) (1.565669,1.865) (1.368994,1.865) (1.680066,1.865) (0.718381,1.865) (1.514132,1.865) (1.883081,1.865)
(0.539668,2.865) (0.669459,2.865) (0.717345,2.865) (0.710760,2.865) (0.746967,2.865) (1.012250,2.865) (1.049335,2.865) (1.290708,2.865)
(1.319382,2.865) (1.427840,2.865) (1.547169,2.865) (0.609368,2.865) (0.706892,2.865) (1.700070,2.865) (0.847648,2.865) (1.817281,2.865)
(0.967509,2.865) (0.927683,2.865) (1.897450,2.865) (1.000667,2.865) (1.113332,2.865) (1.148419,2.865) (1.151053,2.865) (1.258080,2.865)
(1.320121,2.865) (1.305150,2.865) (1.399200,2.865) (2.522753,2.865) (1.709771,2.865) (1.823038,2.865) (1.933645,2.865) (3.317375,2.865)
(2.436844,2.865) (3.401396,2.865) (3.412347,2.865) (2.631595,2.865) (2.704424,2.865) (4.825728,2.865) (5.132100,2.865) (4.866806,2.865)
(0.430364,3.865) (1.606978,3.865) (1.810752,3.865) (2.177932,3.865) (2.901926,3.865) (3.013065,3.865) (3.051526,3.865) (3.241302,3.865)
(3.736677,3.865) (6.042457,3.865) (1.255614,3.865) (1.339183,3.865) (1.751117,3.865) (2.334948,3.865) (2.577903,3.865) (2.662395,3.865)
(2.968273,3.865) (3.209925,3.865) (3.260174,3.865) (3.520399,3.865) (0.814783,3.865) (1.808024,3.865) (2.093910,3.865) (2.406665,3.865)
(3.148794,3.865) (3.487728,3.865) (3.635857,3.865) (3.922025,3.865) (4.194634,3.865) (4.377925,3.865) (1.155358,3.865) (1.229910,3.865)
(1.705088,3.865) (2.039772,3.865) (2.895581,3.865) (2.900029,3.865) (3.058238,3.865) (3.082238,3.865) (3.439209,3.865) (3.714396,3.865)
};

\addplot+[only marks, mark=triangle*, mark size=1.35pt, opacity=0.90,
    mark options={draw=black, fill=figtabblue}] coordinates {
(0.060412,0.955) (0.078354,0.955) (0.107682,0.955) (0.046736,0.955) (0.049466,0.955) (0.026254,0.955) (0.053793,0.955) (0.028935,0.955)
(0.065179,0.955) (0.137761,0.955) (0.145682,0.955) (0.158888,0.955) (0.054762,0.955) (0.169336,0.955) (0.073131,0.955) (0.068541,0.955)
(0.187841,0.955) (0.183695,0.955) (0.125335,0.955) (0.102139,0.955) (0.134718,0.955) (0.095401,0.955) (0.084452,0.955) (0.125033,0.955)
(0.042507,0.955) (0.168312,0.955) (0.066304,0.955) (0.075001,0.955) (0.080479,0.955) (0.071068,0.955) (0.243795,0.955) (0.255647,0.955)
(0.168091,0.955) (0.144414,0.955) (0.279782,0.955) (0.110500,0.955) (0.120790,0.955) (0.236483,0.955) (0.129007,0.955) (0.202953,0.955)
(0.213466,1.955) (0.180708,1.955) (0.247819,1.955) (0.259613,1.955) (0.283162,1.955) (0.337425,1.955) (0.417056,1.955) (0.400257,1.955)
(0.412092,1.955) (0.465327,1.955) (0.506279,1.955) (0.491015,1.955) (0.520201,1.955) (0.573903,1.955) (0.553009,1.955) (0.629724,1.955)
(0.599369,1.955) (0.670067,1.955) (0.653825,1.955) (0.124995,1.955) (0.231728,1.955) (0.225713,1.955) (0.265862,1.955) (0.259948,1.955)
(0.304105,1.955) (0.121831,1.955) (0.142665,1.955) (0.364155,1.955) (0.192790,1.955) (0.396850,1.955) (0.215519,1.955) (0.243509,1.955)
(0.236436,1.955) (0.249557,1.955) (0.292028,1.955) (0.288109,1.955) (0.327789,1.955) (0.389335,1.955) (0.403634,1.955) (0.516934,1.955)
(0.287669,2.955) (0.450665,2.955) (0.460815,2.955) (0.469447,2.955) (0.551993,2.955) (0.564347,2.955) (0.575045,2.955) (0.603925,2.955)
(0.580243,2.955) (0.582868,2.955) (0.647510,2.955) (0.744272,2.955) (0.870231,2.955) (0.929919,2.955) (1.012465,2.955) (1.045868,2.955)
(1.146064,2.955) (1.327275,2.955) (0.451660,2.955) (0.865527,2.955) (0.215531,2.955) (0.286892,2.955) (0.338761,2.955) (0.382477,2.955)
(0.461802,2.955) (0.553982,2.955) (0.556100,2.955) (0.547524,2.955) (0.605202,2.955) (0.553982,2.955) (0.622354,2.955) (0.635446,2.955)
(0.669262,2.955) (0.720039,2.955) (0.749047,2.955) (0.907577,2.955) (0.954219,2.955) (1.039825,2.955) (1.075956,2.955) (1.111119,2.955)
(0.380524,3.955) (0.489805,3.955) (0.516815,3.955) (0.481192,3.955) (0.561724,3.955) (0.624994,3.955) (0.652645,3.955) (0.678066,3.955)
(0.674806,3.955) (0.687014,3.955) (0.692569,3.955) (0.748640,3.955) (0.799961,3.955) (0.860914,3.955) (0.843977,3.955) (0.844885,3.955)
(0.848327,3.955) (0.903712,3.955) (0.886065,3.955) (0.914780,3.955) (0.963702,3.955) (0.966155,3.955) (1.049732,3.955) (1.125319,3.955)
(1.131018,3.955) (1.156704,3.955) (1.170268,3.955) (0.914780,3.955) (1.170863,3.955) (1.235521,3.955) (1.251379,3.955) (1.310552,3.955)
(1.397367,3.955) (1.553388,3.955) (1.515945,3.955) (3.678038,3.955) (0.745962,3.955) (0.797168,3.955) (0.884764,3.955) (1.319240,3.955)
};

\addplot+[only marks, mark=*, mark size=1.35pt, opacity=0.90,
    mark options={draw=black, fill=figtaborange}] coordinates {
(0.010116,1.045) (0.012058,1.045) (0.011127,1.045) (0.034308,1.045) (0.018477,1.045) (0.012342,1.045) (0.046346,1.045) (0.045007,1.045)
(0.049612,1.045) (0.066214,1.045) (0.066287,1.045) (0.060040,1.045) (0.075981,1.045) (0.087313,1.045) (0.090770,1.045) (0.116122,1.045)
(0.138769,1.045) (0.174108,1.045) (0.214576,1.045) (0.216088,1.045) (0.022190,1.045) (0.084719,1.045) (0.049530,1.045) (0.052702,1.045)
(0.008507,1.045) (0.109805,1.045) (0.064213,1.045) (0.080825,1.045) (0.092365,1.045) (0.208698,1.045) (0.047363,1.045) (0.074582,1.045)
(0.276806,1.045) (0.253969,1.045) (0.050565,1.045) (0.123810,1.045) (0.186362,1.045) (0.076546,1.045) (0.127179,1.045) (0.062469,1.045)
(0.023882,2.045) (0.090024,2.045) (0.135450,2.045) (0.130853,2.045) (0.147021,2.045) (0.140328,2.045) (0.183200,2.045) (0.185619,2.045)
(0.282453,2.045) (0.372636,2.045) (0.462951,2.045) (0.452402,2.045) (0.742168,2.045) (1.029617,2.045) (1.366368,2.045) (1.557233,2.045)
(1.747760,2.045) (2.140717,2.045) (2.714721,2.045) (2.743245,2.045) (0.250355,2.045) (0.161846,2.045) (0.273121,2.045) (0.213045,2.045)
(0.113256,2.045) (0.411761,2.045) (0.208297,2.045) (0.230202,2.045) (0.348936,2.045) (0.809269,2.045) (0.709937,2.045) (2.081497,2.045)
(3.556276,2.045) (2.214815,2.045) (0.231118,2.045) (0.950556,2.045) (0.428225,2.045) (0.405691,2.045) (2.849165,2.045) (0.832038,2.045)
(0.237572,3.045) (0.310556,3.045) (0.365221,3.045) (0.651948,3.045) (0.744204,3.045) (1.426224,3.045) (0.503651,3.045) (0.585938,3.045)
(0.643328,3.045) (0.643801,3.045) (0.664663,3.045) (0.857152,3.045) (1.301798,3.045) (1.322188,3.045) (3.270554,3.045) (0.545018,3.045)
(0.574828,3.045) (0.954136,3.045) (2.691025,3.045) (2.826569,3.045) (0.683405,3.045) (0.770290,3.045) (1.390251,3.045) (0.693336,3.045)
(1.645637,3.045) (1.564132,3.045) (0.430083,3.045) (2.320360,3.045) (0.592108,3.045) (3.440054,3.045) (4.396380,3.045) (4.812470,3.045)
(1.367913,3.045) (3.809145,3.045) (1.566064,3.045) (3.171064,3.045) (0.400329,3.045) (1.280991,3.045) (3.194299,3.045) (1.610546,3.045)
(0.860593,4.045) (0.863902,4.045) (1.380412,4.045) (1.795170,4.045) (1.879176,4.045) (2.501719,4.045) (2.757044,4.045) (3.271864,4.045)
(3.356626,4.045) (4.277503,4.045) (4.541033,4.045) (4.882857,4.045) (4.970107,4.045) (5.202054,4.045) (5.406981,4.045) (5.423373,4.045)
(5.709881,4.045) (6.294474,4.045) (6.336152,4.045) (6.925018,4.045) (4.690548,4.045) (4.923014,4.045) (5.717967,4.045) (7.049075,4.045)
(2.577992,4.045) (1.022579,4.045) (4.365720,4.045) (5.108338,4.045) (1.633138,4.045) (5.711581,4.045) (8.844457,4.045) (7.213571,4.045)
(5.661197,4.045) (5.420662,4.045) (5.794572,4.045) (4.685798,4.045) (6.615101,4.045) (4.107212,4.045) (6.881116,4.045) (6.217662,4.045)
};

\addplot+[only marks, mark=*, mark size=1.35pt, opacity=0.90,
    mark options={draw=black, fill=figgold}] coordinates {
(0.009689,1.135) (0.013707,1.135) (0.032680,1.135) (0.012291,1.135) (0.023513,1.135) (0.021893,1.135) (0.035035,1.135) (0.020594,1.135)
(0.036559,1.135) (0.055769,1.135) (0.066573,1.135) (0.127592,1.135) (0.091647,1.135) (0.132256,1.135) (0.125206,1.135) (0.149608,1.135)
(0.210037,1.135) (0.217054,1.135) (0.425698,1.135) (0.498087,1.135) (0.020948,1.135) (0.080658,1.135) (0.024109,1.135) (0.024629,1.135)
(0.035942,1.135) (0.056381,1.135) (0.046986,1.135) (0.033841,1.135) (0.086720,1.135) (0.116608,1.135) (0.135010,1.135) (0.252513,1.135)
(0.246011,1.135) (0.662142,1.135) (0.300349,1.135) (0.209646,1.135) (0.944572,1.135) (0.277824,1.135) (0.450925,1.135) (0.313362,1.135)
(0.133439,2.135) (0.304619,2.135) (0.463258,2.135) (0.467770,2.135) (0.684444,2.135) (0.742718,2.135) (0.994785,2.135) (1.024214,2.135)
(1.136121,2.135) (1.296429,2.135) (1.370663,2.135) (1.368058,2.135) (1.555564,2.135) (1.615120,2.135) (1.711546,2.135) (1.938508,2.135)
(2.030264,2.135) (2.200530,2.135) (2.249382,2.135) (2.549863,2.135) (1.605386,2.135) (1.936648,2.135) (1.434225,2.135) (2.071058,2.135)
(1.105122,2.135) (3.155540,2.135) (0.918351,2.135) (2.083967,2.135) (1.036496,2.135) (1.289967,2.135) (3.419473,2.135) (4.219323,2.135)
(3.625547,2.135) (3.202684,2.135) (1.882915,2.135) (5.054376,2.135) (1.220379,2.135) (2.536878,2.135) (4.049521,2.135) (2.329732,2.135)
(0.758471,3.135) (0.944485,3.135) (1.278006,3.135) (2.344274,3.135) (2.422001,3.135) (2.553643,3.135) (2.925925,3.135) (2.934085,3.135)
(3.221090,3.135) (3.308509,3.135) (1.847040,3.135) (2.161216,3.135) (2.780911,3.135) (2.879547,3.135) (3.203943,3.135) (3.377406,3.135)
(3.495304,3.135) (3.841523,3.135) (4.109432,3.135) (4.324881,3.135) (4.854865,3.135) (4.060043,3.135) (4.202851,3.135) (1.766510,3.135)
(4.429477,3.135) (7.289521,3.135) (8.300872,3.135) (4.587801,3.135) (4.975802,3.135) (5.119958,3.135) (9.773174,3.135) (4.359272,3.135)
(8.564747,3.135) (8.201757,3.135) (8.228128,3.135) (9.845291,3.135) (2.604156,3.135) (8.583354,3.135) (1.950803,3.135) (16.823991,3.135)
(2.612779,4.135) (3.722252,4.135) (3.763650,4.135) (3.763091,4.135) (4.150253,4.135) (4.917210,4.135) (4.962196,4.135) (5.136314,4.135)
(5.312572,4.135) (5.692973,4.135) (5.826784,4.135) (5.879990,4.135) (6.540710,4.135) (6.560103,4.135) (6.698084,4.135) (6.770190,4.135)
(6.836627,4.135) (6.827861,4.135) (6.967840,4.135) (7.007261,4.135) (14.451992,4.135) (8.626209,4.135) (11.510809,4.135) (10.694176,4.135)
(9.682498,4.135) (7.469224,4.135) (7.780116,4.135) (9.280539,4.135) (10.547624,4.135) (13.923123,4.135) (18.782055,4.135) (11.361560,4.135)
(14.118884,4.135) (12.327310,4.135) (15.675890,4.135) (19.552144,4.135) (15.680166,4.135) (7.878028,4.135) (13.779740,4.135) (12.288433,4.135)
};

\addplot+[only marks, mark=*, mark size=1.35pt, opacity=0.90,
    mark options={draw=black, fill=figtabblue}] coordinates {
(0.037735,1.225) (0.022407,1.225) (0.036907,1.225) (0.069299,1.225) (0.018225,1.225) (0.062603,1.225) (0.041583,1.225) (0.018289,1.225)
(0.087522,1.225) (0.026759,1.225) (0.016987,1.225) (0.029522,1.225) (0.071396,1.225) (0.109707,1.225) (0.022679,1.225) (0.035453,1.225)
(0.094507,1.225) (0.059808,1.225) (0.139793,1.225) (0.137119,1.225) (0.138225,1.225) (0.080148,1.225) (0.134477,1.225) (0.154733,1.225)
(0.164501,1.225) (0.148882,1.225) (0.127473,1.225) (0.013707,1.225) (0.015097,1.225) (0.175896,1.225) (0.014694,1.225) (0.016760,1.225)
(0.160019,1.225) (0.024687,1.225) (0.105591,1.225) (0.020992,1.225) (0.032545,1.225) (0.050002,1.225) (0.098851,1.225) (0.194049,1.225)
(0.140569,2.225) (0.223503,2.225) (0.308357,2.225) (0.181034,2.225) (0.339078,2.225) (0.330864,2.225) (0.400172,2.225) (0.171355,2.225)
(0.366279,2.225) (0.394215,2.225) (0.398430,2.225) (0.438759,2.225) (0.341615,2.225) (0.493941,2.225) (0.530615,2.225) (0.496168,2.225)
(0.407355,2.225) (0.558840,2.225) (0.464880,2.225) (0.540660,2.225) (0.038954,2.225) (0.110555,2.225) (0.137291,2.225) (0.182075,2.225)
(0.175905,2.225) (0.178993,2.225) (0.170216,2.225) (0.202608,2.225) (0.203665,2.225) (0.216795,2.225) (0.253961,2.225) (0.288908,2.225)
(0.325618,2.225) (0.331233,2.225) (0.354934,2.225) (0.410598,2.225) (0.385028,2.225) (0.404168,2.225) (0.457785,2.225) (0.464863,2.225)
(0.339243,3.225) (0.413319,3.225) (0.478794,3.225) (0.502743,3.225) (0.505883,3.225) (0.649940,3.225) (0.683102,3.225) (0.739351,3.225)
(0.746587,3.225) (0.731946,3.225) (0.775442,3.225) (0.783139,3.225) (0.239262,3.225) (0.860688,3.225) (0.871936,3.225) (0.891044,3.225)
(0.918151,3.225) (0.313198,3.225) (0.324180,3.225) (0.956809,3.225) (0.977770,3.225) (0.404877,3.225) (0.423281,3.225) (1.052372,3.225)
(0.460535,3.225) (0.485813,3.225) (0.541815,3.225) (0.625130,3.225) (0.629370,3.225) (0.667555,3.225) (0.660679,3.225) (0.641413,3.225)
(0.687371,3.225) (0.716073,3.225) (0.780213,3.225) (0.795623,3.225) (0.781252,3.225) (1.417781,3.225) (0.834367,3.225) (0.876106,3.225)
(0.725330,4.225) (0.887489,4.225) (1.101670,4.225) (1.145973,4.225) (1.155042,4.225) (1.139061,4.225) (1.181437,4.225) (1.194482,4.225)
(1.241700,4.225) (1.232152,4.225) (1.244976,4.225) (1.262269,4.225) (1.283831,4.225) (1.375006,4.225) (1.372729,4.225) (1.410594,4.225)
(1.552576,4.225) (1.600956,4.225) (1.604327,4.225) (0.664565,4.225) (0.672030,4.225) (0.697615,4.225) (1.958714,4.225) (0.737825,4.225)
(0.810189,4.225) (0.837890,4.225) (0.853439,4.225) (0.923192,4.225) (0.964605,4.225) (0.967092,4.225) (1.016973,4.225) (1.074484,4.225)
(1.110760,4.225) (1.219239,4.225) (1.246491,4.225) (1.299327,4.225) (1.488998,4.225) (1.628077,4.225) (1.766737,4.225) (1.938958,4.225)
};

\end{axis}
\end{tikzpicture}
\end{minipage}

\par\vspace{0.8em}

\begin{minipage}[t]{0.98\textwidth}
\centering
\begin{tikzpicture}
\begin{axis}[
    width=\linewidth,
    height=5.25cm,
    xmin=-0.5,
    xmax=20.5,
    ymin=0.55,
    ymax=4.45,
    y dir=reverse,
    xlabel={Total tokens (millions)},
    ylabel={Minimal DFA States},
    ytick={1,2,3,4},
    yticklabels={2--3,4--5,6--7,8--9},
    xtick={0,4,8,12,16,20},
    grid=major,
    grid style={gray!18},
    tick label style={font=\scriptsize},
    label style={font=\small},
    clip=true,
    legend style={
        at={(0.97,0.97)},
        anchor=north east,
        draw=black!25,
        fill=white,
        fill opacity=0.90,
        text opacity=1,
        font=\scriptsize,
        cells={anchor=west},
        row sep=-1pt,
    },
]

\addlegendimage{only marks, mark=triangle*, mark size=1.5pt, mark options={draw=black,fill=black}}
\addlegendentry{Stateful}
\addlegendimage{only marks, mark=*, mark size=1.5pt, mark options={draw=black,fill=black}}
\addlegendentry{Stateless}

\addplot+[only marks, mark=triangle*, mark size=1.35pt, opacity=0.90,
    mark options={draw=black, fill=figtaborange}] coordinates {
(0.060818,0.775) (0.044557,0.775) (0.058746,0.775) (0.063428,0.775) (0.024000,0.775) (0.057246,0.775) (0.057234,0.775) (0.068777,0.775)
(0.055732,0.775) (0.074409,0.775) (0.071851,0.775) (0.089578,0.775) (0.114328,0.775) (0.096389,0.775) (0.096238,0.775) (0.028081,0.775)
(0.009748,0.775) (0.141948,0.775) (0.017246,0.775) (0.008705,0.775) (0.010094,0.775) (0.103391,0.775) (0.031753,0.775) (0.019486,0.775)
(0.045379,0.775) (0.033613,0.775) (0.043828,0.775) (0.027348,0.775) (0.035834,0.775) (0.055910,0.775) (0.029772,0.775) (0.070094,0.775)
(0.105361,0.775) (0.070191,0.775) (0.213400,0.775) (0.031137,0.775) (0.094327,0.775) (0.127214,0.775) (0.118934,0.775) (0.043259,0.775)
(0.137418,1.775) (0.159627,1.775) (0.470622,1.775) (0.228923,1.775) (0.256157,1.775) (0.261139,1.775) (0.253223,1.775) (0.124658,1.775)
(0.204506,1.775) (0.319587,1.775) (0.281578,1.775) (0.904121,1.775) (1.728502,1.775) (1.140671,1.775) (0.804077,1.775) (1.524950,1.775)
(1.024985,1.775) (1.918193,1.775) (0.815347,1.775) (3.081228,1.775) (0.086925,1.775) (0.128070,1.775) (0.128635,1.775) (0.117952,1.775)
(0.116202,1.775) (0.130711,1.775) (0.105512,1.775) (0.165654,1.775) (0.158390,1.775) (0.220725,1.775) (0.194128,1.775) (0.339501,1.775)
(0.206839,1.775) (0.240678,1.775) (0.305995,1.775) (0.812092,1.775) (0.480999,1.775) (0.695970,1.775) (0.745910,1.775) (1.156953,1.775)
(0.630143,2.775) (0.462561,2.775) (0.399225,2.775) (0.615090,2.775) (0.475495,2.775) (0.390865,2.775) (0.797162,2.775) (0.427415,2.775)
(0.619820,2.775) (1.182715,2.775) (1.370557,2.775) (1.027165,2.775) (0.254110,2.775) (0.412368,2.775) (0.224169,2.775) (0.339425,2.775)
(0.417391,2.775) (0.313685,2.775) (0.447474,2.775) (0.305290,2.775) (0.349465,2.775) (5.292465,2.775) (0.450330,2.775) (0.703230,2.775)
(0.264927,2.775) (2.060455,2.775) (1.516800,2.775) (0.796297,2.775) (1.292071,2.775) (0.595320,2.775) (3.720842,2.775) (2.997706,2.775)
(0.697796,2.775) (3.845303,2.775) (10.078725,2.775) (1.845861,2.775) (3.285858,2.775) (1.832359,2.775) (1.990263,2.775) (3.054975,2.775)
(0.953505,3.775) (1.412886,3.775) (1.719742,3.775) (0.514931,3.775) (1.030008,3.775) (1.092810,3.775) (1.175478,3.775) (0.449642,3.775)
(2.514179,3.775) (0.967467,3.775) (0.934929,3.775) (1.641637,3.775) (1.136840,3.775) (1.703641,3.775) (2.140080,3.775) (1.251087,3.775)
(1.636525,3.775) (0.713620,3.775) (1.168579,3.775) (2.739230,3.775) (7.691704,3.775) (2.197473,3.775) (5.211391,3.775) (3.384382,3.775)
(4.499232,3.775) (6.377812,3.775) (6.142991,3.775) (2.469387,3.775) (4.593497,3.775) (4.930907,3.775) (9.536649,3.775) (14.679010,3.775)
(3.525250,3.775) (5.782092,3.775) (5.216345,3.775) (6.394388,3.775) (7.713155,3.775) (7.643434,3.775) (13.383786,3.775) (2.894016,3.775)
};

\addplot+[only marks, mark=triangle*, mark size=1.35pt, opacity=0.90,
    mark options={draw=black, fill=figgold}] coordinates {
(0.030246,0.865) (0.050873,0.865) (0.040271,0.865) (0.053925,0.865) (0.080062,0.865) (0.087131,0.865) (0.117289,0.865) (0.078689,0.865)
(0.091894,0.865) (0.097424,0.865) (0.151311,0.865) (0.061481,0.865) (0.055983,0.865) (0.130917,0.865) (0.152652,0.865) (0.105697,0.865)
(0.138021,0.865) (0.269234,0.865) (0.016485,0.865) (0.021100,0.865) (0.023333,0.865) (0.016124,0.865) (0.019289,0.865) (0.016303,0.865)
(0.023666,0.865) (0.025800,0.865) (0.038774,0.865) (0.013217,0.865) (0.015397,0.865) (0.020814,0.865) (0.035637,0.865) (0.011722,0.865)
(0.016307,0.865) (0.015262,0.865) (0.025370,0.865) (0.037374,0.865) (0.042055,0.865) (0.062692,0.865) (0.257033,0.865) (0.255449,0.865)
(0.193171,1.865) (0.274969,1.865) (0.293363,1.865) (0.336712,1.865) (0.303642,1.865) (0.329295,1.865) (0.418704,1.865) (0.442092,1.865)
(0.418980,1.865) (0.417341,1.865) (0.508054,1.865) (0.472063,1.865) (0.061348,1.865) (0.451304,1.865) (0.131036,1.865) (0.461073,1.865)
(0.157347,1.865) (0.179728,1.865) (0.146681,1.865) (0.217975,1.865) (0.791563,1.865) (0.254178,1.865) (0.780485,1.865) (0.287069,1.865)
(0.181904,1.865) (0.271984,1.865) (0.296950,1.865) (0.444024,1.865) (0.301852,1.865) (0.759574,1.865) (0.245626,1.865) (0.343814,1.865)
(0.838802,1.865) (0.440606,1.865) (0.794525,1.865) (0.584557,1.865) (0.775860,1.865) (0.386144,1.865) (0.648015,1.865) (0.676350,1.865)
(0.392381,2.865) (0.508433,2.865) (0.368202,2.865) (0.779913,2.865) (0.543086,2.865) (0.759722,2.865) (0.501872,2.865) (0.799534,2.865)
(0.742423,2.865) (0.651243,2.865) (0.760653,2.865) (0.332136,2.865) (0.428313,2.865) (1.036767,2.865) (0.366882,2.865) (0.894911,2.865)
(0.513533,2.865) (0.545329,2.865) (0.989964,2.865) (0.611356,2.865) (0.668880,2.865) (0.598943,2.865) (0.511996,2.865) (0.709004,2.865)
(0.566450,2.865) (0.719118,2.865) (0.716086,2.865) (1.669974,2.865) (0.792160,2.865) (0.673176,2.865) (0.882241,2.865) (1.741843,2.865)
(0.948862,2.865) (1.888763,2.865) (1.530772,2.865) (1.249123,2.865) (1.202964,2.865) (2.055750,2.865) (3.533286,2.865) (2.098779,2.865)
(0.439965,3.865) (1.601347,3.865) (1.490861,3.865) (2.048356,3.865) (1.975420,3.865) (3.068055,3.865) (2.921498,3.865) (3.191933,3.865)
(5.283590,3.865) (3.272125,3.865) (0.887038,3.865) (1.019772,3.865) (1.609110,3.865) (3.401653,3.865) (3.296749,3.865) (2.701564,3.865)
(2.469555,3.865) (2.564412,3.865) (3.127056,3.865) (2.594502,3.865) (1.104612,3.865) (3.194949,3.865) (3.531979,3.865) (2.323066,3.865)
(3.377331,3.865) (2.962824,3.865) (3.600484,3.865) (3.196580,3.865) (3.387544,3.865) (3.480705,3.865) (1.074060,3.865) (1.072033,3.865)
(2.295648,3.865) (1.582601,3.865) (3.412625,3.865) (2.600083,3.865) (2.924552,3.865) (3.346182,3.865) (2.382187,3.865) (2.155496,3.865)
};

\addplot+[only marks, mark=triangle*, mark size=1.35pt, opacity=0.90,
    mark options={draw=black, fill=figtabblue}] coordinates {
(0.115647,0.955) (0.183883,0.955) (0.519562,0.955) (0.075560,0.955) (0.078174,0.955) (0.030573,0.955) (0.075887,0.955) (0.031789,0.955)
(0.102489,0.955) (0.270212,0.955) (0.541743,0.955) (0.275681,0.955) (0.042281,0.955) (0.322847,0.955) (0.069870,0.955) (0.078772,0.955)
(0.389503,0.955) (0.369334,0.955) (0.238883,0.955) (0.119001,0.955) (0.315990,0.955) (0.095247,0.955) (0.106818,0.955) (0.117549,0.955)
(0.049497,0.955) (0.336704,0.955) (0.087889,0.955) (0.085378,0.955) (0.076481,0.955) (0.094086,0.955) (0.498784,0.955) (0.526794,0.955)
(0.187723,0.955) (0.163114,0.955) (0.589065,0.955) (0.096730,0.955) (0.131566,0.955) (1.135813,0.955) (0.173998,0.955) (0.224814,0.955)
(0.557685,1.955) (1.132992,1.955) (1.804266,1.955) (1.168871,1.955) (0.847802,1.955) (1.991393,1.955) (1.660932,1.955) (2.187858,1.955)
(0.934689,1.955) (1.885873,1.955) (2.621903,1.955) (2.449805,1.955) (2.560383,1.955) (2.473722,1.955) (2.294939,1.955) (4.744230,1.955)
(2.599310,1.955) (2.754100,1.955) (2.281050,1.955) (0.196772,1.955) (0.429980,1.955) (1.144323,1.955) (0.620965,1.955) (0.679773,1.955)
(0.869303,1.955) (0.609299,1.955) (0.269018,1.955) (0.983319,1.955) (1.096483,1.955) (1.210344,1.955) (0.484728,1.955) (0.543789,1.955)
(0.745129,1.955) (0.783261,1.955) (1.789244,1.955) (0.926739,1.955) (1.210364,1.955) (1.824285,1.955) (0.983910,1.955) (2.346632,1.955)
(0.867131,2.955) (2.188275,2.955) (2.594133,2.955) (2.506844,2.955) (1.847235,2.955) (7.236404,2.955) (3.526400,2.955) (6.857617,2.955)
(3.988000,2.955) (3.601907,2.955) (9.047298,2.955) (3.730753,2.955) (4.046457,2.955) (4.341614,2.955) (8.127461,2.955) (8.594451,2.955)
(16.558550,2.955) (5.559981,2.955) (1.803155,2.955) (4.522464,2.955) (0.794872,2.955) (0.958967,2.955) (1.122679,2.955) (2.009795,2.955)
(1.548838,2.955) (4.952749,2.955) (4.327657,2.955) (1.935277,2.955) (2.307359,2.955) (4.952749,2.955) (6.134053,2.955) (2.586308,2.955)
(2.731781,2.955) (3.576713,2.955) (3.962578,2.955) (4.951484,2.955) (2.042512,2.955) (6.655220,2.955) (5.220230,2.955) (4.856014,2.955)
(1.857253,3.955) (2.329426,3.955) (2.745483,3.955) (4.570816,3.955) (4.753035,3.955) (6.443986,3.955) (4.314421,3.955) (7.068689,3.955)
(2.467016,3.955) (4.523545,3.955) (5.833595,3.955) (4.446217,3.955) (7.667803,3.955) (7.161389,3.955) (4.780150,3.955) (13.780824,3.955)
(4.860227,3.955) (5.437797,3.955) (7.936651,3.955) (8.220543,3.955) (11.521598,3.955) (6.624193,3.955) (7.208327,3.955) (13.609873,3.955)
(7.028985,3.955) (13.078214,3.955) (8.590716,3.955) (8.220543,3.955) (10.530994,3.955) (10.133609,3.955) (7.001734,3.955) (6.748333,3.955)
(16.998839,3.955) (17.209049,3.955) (10.932469,3.955) (11.396068,3.955) (11.457085,3.955) (5.976888,3.955) (10.245483,3.955) (15.338558,3.955)
};

\addplot+[only marks, mark=*, mark size=1.35pt, opacity=0.90,
    mark options={draw=black, fill=figtaborange}] coordinates {
(0.007195,1.045) (0.009524,1.045) (0.008196,1.045) (0.022129,1.045) (0.017632,1.045) (0.013216,1.045) (0.030904,1.045) (0.030397,1.045)
(0.036679,1.045) (0.040604,1.045) (0.038471,1.045) (0.051019,1.045) (0.053591,1.045) (0.064410,1.045) (0.069583,1.045) (0.063295,1.045)
(0.083090,1.045) (0.113909,1.045) (0.085688,1.045) (0.131837,1.045) (0.021970,1.045) (0.053010,1.045) (0.035857,1.045) (0.038665,1.045)
(0.008249,1.045) (0.063424,1.045) (0.044687,1.045) (0.045911,1.045) (0.054620,1.045) (0.127130,1.045) (0.051734,1.045) (0.070783,1.045)
(0.114823,1.045) (0.170008,1.045) (0.054269,1.045) (0.093777,1.045) (0.098757,1.045) (0.078110,1.045) (0.101912,1.045) (0.055435,1.045)
(0.020941,2.045) (0.068229,2.045) (0.118857,2.045) (0.100589,2.045) (0.117495,2.045) (0.093473,2.045) (0.109824,2.045) (0.131919,2.045)
(0.231332,2.045) (0.209052,2.045) (0.290271,2.045) (0.291922,2.045) (0.486039,2.045) (0.568725,2.045) (0.975273,2.045) (1.042858,2.045)
(1.206809,2.045) (1.543649,2.045) (2.255554,2.045) (2.103434,2.045) (0.170132,2.045) (0.129260,2.045) (0.180216,2.045) (0.155739,2.045)
(0.100862,2.045) (0.254957,2.045) (0.151598,2.045) (0.172266,2.045) (0.245841,2.045) (0.453476,2.045) (0.450172,2.045) (1.552213,2.045)
(3.048719,2.045) (1.466244,2.045) (0.192214,2.045) (0.622913,2.045) (0.300068,2.045) (0.244247,2.045) (1.780472,2.045) (0.536931,2.045)
(0.209004,3.045) (0.233476,3.045) (0.269559,3.045) (0.451407,3.045) (0.568965,3.045) (0.923080,3.045) (0.235835,3.045) (0.422673,3.045)
(0.403715,3.045) (0.379393,3.045) (0.558782,3.045) (0.649175,3.045) (0.991995,3.045) (0.880183,3.045) (3.171833,3.045) (0.388613,3.045)
(0.446243,3.045) (0.693120,3.045) (3.986797,3.045) (2.654176,3.045) (0.452967,3.045) (0.497082,3.045) (1.047022,3.045) (0.405587,3.045)
(1.061115,3.045) (1.328406,3.045) (0.333215,3.045) (1.805030,3.045) (0.347707,3.045) (2.903736,3.045) (3.525966,3.045) (4.348039,3.045)
(0.790081,3.045) (3.003616,3.045) (1.086698,3.045) (3.131402,3.045) (0.230240,3.045) (0.901745,3.045) (3.005681,3.045) (1.015868,3.045)
(0.573955,4.045) (0.593258,4.045) (1.058743,4.045) (1.399816,4.045) (1.716288,4.045) (2.280368,4.045) (3.089615,4.045) (3.133727,4.045)
(3.680844,4.045) (6.783244,4.045) (4.488664,4.045) (5.695701,4.045) (5.171207,4.045) (5.283606,4.045) (5.043877,4.045) (6.084965,4.045)
(6.997376,4.045) (6.490774,4.045) (7.291923,4.045) (6.778568,4.045) (4.484302,4.045) (4.689533,4.045) (5.851290,4.045) (6.697887,4.045)
(1.991316,4.045) (0.673082,4.045) (3.848886,4.045) (5.176293,4.045) (1.041582,4.045) (6.189568,4.045) (8.367380,4.045) (5.722474,4.045)
(4.931336,4.045) (5.530660,4.045) (5.733975,4.045) (3.142896,4.045) (5.994585,4.045) (3.230196,4.045) (5.719849,4.045) (5.727695,4.045)
};

\addplot+[only marks, mark=*, mark size=1.35pt, opacity=0.90,
    mark options={draw=black, fill=figgold}] coordinates {
(0.007429,1.135) (0.008039,1.135) (0.024780,1.135) (0.009187,1.135) (0.019210,1.135) (0.016929,1.135) (0.016739,1.135) (0.011916,1.135)
(0.019715,1.135) (0.032620,1.135) (0.037822,1.135) (0.050508,1.135) (0.031079,1.135) (0.044030,1.135) (0.037534,1.135) (0.054934,1.135)
(0.067879,1.135) (0.077399,1.135) (0.157397,1.135) (0.154463,1.135) (0.009273,1.135) (0.021546,1.135) (0.012400,1.135) (0.011081,1.135)
(0.015401,1.135) (0.017952,1.135) (0.018452,1.135) (0.015239,1.135) (0.022517,1.135) (0.028378,1.135) (0.038955,1.135) (0.058507,1.135)
(0.062795,1.135) (0.120207,1.135) (0.084342,1.135) (0.061484,1.135) (0.190981,1.135) (0.061055,1.135) (0.101721,1.135) (0.073496,1.135)
(0.060161,2.135) (0.138900,2.135) (0.171590,2.135) (0.191961,2.135) (0.248641,2.135) (0.264437,2.135) (0.319663,2.135) (0.349265,2.135)
(0.403603,2.135) (0.488695,2.135) (0.504075,2.135) (0.392835,2.135) (0.522253,2.135) (0.556830,2.135) (0.625182,2.135) (0.669347,2.135)
(0.715435,2.135) (0.787576,2.135) (0.721395,2.135) (0.929698,2.135) (0.343606,2.135) (0.349828,2.135) (0.282279,2.135) (0.332787,2.135)
(0.231982,2.135) (0.538280,2.135) (0.179906,2.135) (0.380330,2.135) (0.216900,2.135) (0.248561,2.135) (0.689862,2.135) (0.785846,2.135)
(0.752542,2.135) (0.604013,2.135) (0.367788,2.135) (0.928524,2.135) (0.260166,2.135) (0.495665,2.135) (0.748644,2.135) (0.465483,2.135)
(0.309340,3.135) (0.396584,3.135) (0.470625,3.135) (0.888370,3.135) (0.861753,3.135) (0.926838,3.135) (1.070917,3.135) (1.030222,3.135)
(1.246490,3.135) (1.246542,3.135) (0.715290,3.135) (0.739548,3.135) (1.190012,3.135) (1.210756,3.135) (1.345422,3.135) (1.345562,3.135)
(1.464577,3.135) (1.333311,3.135) (1.698964,3.135) (1.642935,3.135) (0.940003,3.135) (0.880708,3.135) (0.856876,3.135) (0.377095,3.135)
(0.878008,3.135) (1.495336,3.135) (2.847348,3.135) (0.963463,3.135) (0.979480,3.135) (0.968099,3.135) (2.018124,3.135) (0.791557,3.135)
(1.614758,3.135) (1.529744,3.135) (1.886818,3.135) (1.916330,3.135) (0.501077,3.135) (1.762971,3.135) (0.419229,3.135) (3.324188,3.135)
(1.083683,4.135) (1.584844,4.135) (1.548258,4.135) (1.615862,4.135) (1.802612,4.135) (2.199441,4.135) (1.980263,4.135) (2.141075,4.135)
(2.440053,4.135) (2.276524,4.135) (2.014705,4.135) (2.279218,4.135) (2.611600,4.135) (2.563836,4.135) (2.945342,4.135) (2.717953,4.135)
(2.914641,4.135) (2.885080,4.135) (2.677977,4.135) (2.775277,4.135) (2.843624,4.135) (1.647521,4.135) (2.323516,4.135) (2.064662,4.135)
(2.175086,4.135) (1.439154,4.135) (1.598005,4.135) (1.783587,4.135) (2.125023,4.135) (2.964028,4.135) (4.106792,4.135) (2.392992,4.135)
(4.060699,4.135) (2.572755,4.135) (3.264591,4.135) (3.714569,4.135) (3.339458,4.135) (1.597764,4.135) (2.987297,4.135) (2.477085,4.135)
};

\addplot+[only marks, mark=*, mark size=1.35pt, opacity=0.90,
    mark options={draw=black, fill=figtabblue}] coordinates {
(0.047551,1.225) (0.026537,1.225) (0.039679,1.225) (0.091638,1.225) (0.021044,1.225) (0.085787,1.225) (0.046684,1.225) (0.017979,1.225)
(0.120027,1.225) (0.028667,1.225) (0.017201,1.225) (0.028853,1.225) (0.098499,1.225) (0.154023,1.225) (0.023136,1.225) (0.036162,1.225)
(0.131399,1.225) (0.060458,1.225) (0.194700,1.225) (0.182838,1.225) (0.223736,1.225) (0.083119,1.225) (0.200582,1.225) (0.219146,1.225)
(0.211986,1.225) (0.251753,1.225) (0.148075,1.225) (0.012638,1.225) (0.014904,1.225) (0.254629,1.225) (0.012485,1.225) (0.018033,1.225)
(0.203601,1.225) (0.024424,1.225) (0.111937,1.225) (0.022478,1.225) (0.035862,1.225) (0.048806,1.225) (0.103409,1.225) (0.224109,1.225)
(0.279463,2.225) (0.387974,2.225) (0.548918,2.225) (0.290779,2.225) (0.661385,2.225) (0.688135,2.225) (1.013956,2.225) (0.461673,2.225)
(0.638778,2.225) (0.816661,2.225) (0.766544,2.225) (0.936790,2.225) (0.586506,2.225) (0.787401,2.225) (1.016359,2.225) (1.013209,2.225)
(0.989282,2.225) (1.327651,2.225) (1.562486,2.225) (0.858210,2.225) (0.049644,2.225) (0.187560,2.225) (0.238055,2.225) (0.256880,2.225)
(0.287852,2.225) (0.335975,2.225) (0.300225,2.225) (0.382709,2.225) (0.326944,2.225) (0.323821,2.225) (0.457634,2.225) (0.566731,2.225)
(0.651393,2.225) (0.553818,2.225) (0.513478,2.225) (0.740036,2.225) (0.658504,2.225) (0.703694,2.225) (0.936446,2.225) (0.999218,2.225)
(0.618215,3.225) (0.878422,3.225) (1.093469,3.225) (1.447596,3.225) (1.069916,3.225) (1.679854,3.225) (1.666286,3.225) (2.245636,3.225)
(2.359067,3.225) (1.967936,3.225) (1.949391,3.225) (1.753195,3.225) (0.399350,3.225) (2.222055,3.225) (2.686680,3.225) (1.711698,3.225)
(3.125071,3.225) (0.583742,3.225) (0.653029,3.225) (2.515647,3.225) (1.851097,3.225) (0.724177,3.225) (0.858225,3.225) (2.700789,3.225)
(0.798123,3.225) (0.952516,3.225) (1.337714,3.225) (1.753323,3.225) (1.639069,3.225) (2.073818,3.225) (1.362703,3.225) (1.391178,3.225)
(1.293403,3.225) (2.299717,3.225) (1.822089,3.225) (2.060596,3.225) (1.950795,3.225) (5.115777,3.225) (2.102565,3.225) (1.912284,3.225)
(1.682168,4.225) (2.100073,4.225) (2.491914,4.225) (3.390935,4.225) (4.396203,4.225) (3.790786,4.225) (3.290680,4.225) (3.766385,4.225)
(4.956839,4.225) (4.853680,4.225) (5.617976,4.225) (4.147246,4.225) (4.405465,4.225) (5.014185,4.225) (2.753180,4.225) (6.618329,4.225)
(4.430793,4.225) (4.454924,4.225) (4.627510,4.225) (1.474257,4.225) (1.788412,4.225) (3.767542,4.225) (6.794675,4.225) (1.982820,4.225)
(1.789944,4.225) (1.674720,4.225) (1.702386,4.225) (3.245991,4.225) (3.125555,4.225) (2.583274,4.225) (2.318250,4.225) (2.778431,4.225)
(3.251916,4.225) (3.353095,4.225) (4.588783,4.225) (5.337869,4.225) (4.315948,4.225) (3.400825,4.225) (5.042284,4.225) (6.085999,4.225)
};

\end{axis}
\end{tikzpicture}
\end{minipage}

\end{minipage}%
}
\captionof{figure}{\textbf{Computational cost, runtime, and token usage by paired number of minimal DFA states.} Each point is a run. The y-axis groups runs into state pairs (2--3, 4--5, 6--7, and 8--9); color indicates the model, while triangles denote Stateful and circles denote Stateless interaction.}
\label{fig:cost_runtime_tokens_by_paired_states}
\end{center}

%% file: latex/sections/appendix/query_budget_factor_overleaf.tex
% Intentionally empty appendix fragment; retained for source compatibility.